\documentclass[journal]{IEEEtran}
\usepackage{graphicx}
\usepackage{graphicx}
\usepackage{amssymb}
\usepackage{subfigure}
\usepackage{makecell}
\usepackage{color,soul}
\usepackage{multirow}
\usepackage{rotating}
\usepackage{enumitem}
\usepackage[normalem]{ulem}
\usepackage{colortbl}
\usepackage{adjustbox}
\usepackage{tabularx}
\usepackage{tikz,pgfplots,siunitx,pgfplotstable}
\usepackage{filecontents}

\ifCLASSINFOpdf
\else

\fi
\usepackage{xcolor}
\definecolor{bluegreen}{RGB}{0,153,153}
\definecolor{bluepurple}{RGB}{102,0,204}
\definecolor{bluered}{RGB}{204,0,102}

\usepackage{subfigure}
\usepackage{amsmath}
\usepackage{mathtools,nccmath}

\usepackage[linesnumbered,boxed,ruled,commentsnumbered]{algorithm2e}

\SetCommentSty{mycommfont}
\newcommand{\MODELNAME}{MTGN}

\begin{document}

\title{Who Should I Engage with At What Time? A Missing Event Aware Temporal Graph Neural Network} 
\author{\IEEEauthorblockN{Mingyi Liu,
Zhiying Tu~\IEEEmembership{Member,~IEEE},
Xiaofei Xu~\IEEEmembership{Member,~IEEE}, 
Zhongjie Wang~\IEEEmembership{Member,~IEEE}}
\thanks{Mingyi Liu, Zhiying Tu, Xiaofei Xu, and Zhongjie Wang are with the Faculty of Computing, Harbin Institute of Technology, Harbin, China (e-mail: liumy@hit.edu.cn, tzy\_hit@hit.edu.cn, xiaofei@hit.edu.cn, rainy@hit.edu.cn)}

\thanks{Manuscript received XXXXXX; XXXXXXXXX. 
Corresponding author: Zhongjie Wang (email: rainy@hit.edu.cn).}}

\markboth{Journal of \LaTeX\ Class Files,~Vol.~14, No.~8, August~2015}%
{Shell \MakeLowercase{\textit{et al.}}: Bare Demo of IEEEtran.cls for IEEE Transactions on Magnetics Journals}

\IEEEtitleabstractindextext{%
\begin{abstract}
Temporal graph neural network has recently received significant attention due to its wide application scenarios, such as bioinformatics, knowledge graphs, and social networks. There are some temporal graph neural networks that achieve remarkable results. However, these works focus on future event prediction and are performed under the assumption that all historical events are observable. In real-world applications, events are not always observable, and estimating event time is as important as predicting future events. In this paper, we propose \MODELNAME, a missing event-aware temporal graph neural network, which uniformly models evolving graph structure and timing of events to support predicting what will happen in the future and when it will happen. \MODELNAME~models the dynamic of both observed and missing events as two coupled temporal point processes, thereby incorporating the effects of missing events into the network. Experimental results on several real-world temporal graphs demonstrate that \MODELNAME~significantly outperforms existing methods with up to $89\%$ and $112\%$ more accurate time and link prediction. {Code can be found on https://github.com/HIT-ICES/TNNLS-MTGN}.
\end{abstract}

}

\maketitle

\IEEEdisplaynontitleabstractindextext

\IEEEpeerreviewmaketitle

\section{Introduction}\label{sec:intro}
\IEEEPARstart{G}raph structured data has recently received significant attention due to its wide application scenarios in various domains such as social networks, knowledge graphs, and bioinformatics. Graph neural networks (GNNs) are developed to efficiently learn high-dimensional and non-Euclidean graph information from graphs. Most existing GNNs~\cite{GCN,GAT} are designed for static graphs. In the real world, Graphs tend to evolve continuously. For example, new friendships may be established between people in a social network. Incorporating dynamics into GNNs is a non-trivial problem.

\begin{figure}[h]
    \includegraphics[width=\linewidth]{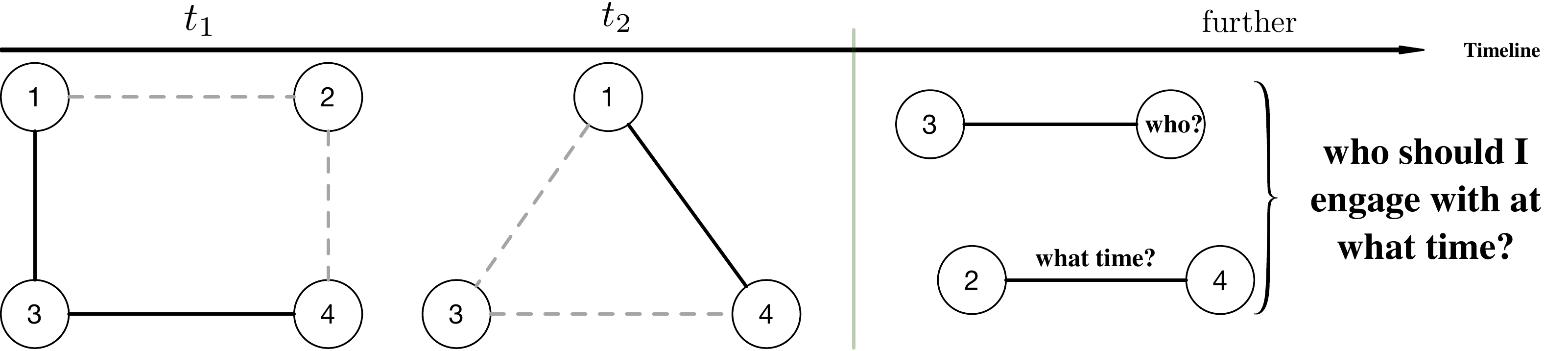}
    \caption{A toy example showing the two core tasks and missing events phenomenon in temporal graph. The solid lines are observed events and the dash lines are missing events.}
\end{figure}
Recently, a few temporal graph neural networks have been developed. These methods can be classified as discrete temporal GNNs and continuous temporal GNNs based on how they represent the temporal graph~\cite{skarding2020foundations, kazemi2020representation}. The discrete temporal GNNs~\cite{CTGCN,EvolveGCN, jiao2021temporal} treat temporal graph as a sequence of graph snapshots to simplify the model, which results in their inability to capture the fully continuous evolution as fine-grained temporal information is lost. As a result, the discrete temporal GNNs can only predict what will happen in the future but cannot predict when. The continuous temporal GNNs treat the temporal graph as a stream of events. The continuous temporal GNNs can be roughly divided into temporal point process (TPP) based methods~\cite{TREND,EvoKG,DyRep} and non-TPP based methods~\cite{APAN,StreamGNN,tang2022dynamic}. The non-TPP based methods encode the event time as an auxiliary feature of the temporal graph structure using RNNs or Transformers, with the consequence that they still fail to predict the event time. TPP based methods model the temporal graph as a temporal point process of events and parameterize this process using a neural network. TPP can naturally model the event time so that, theoretically, the TPP based methods can predict the time of the event. 

{However, most TPP-based methods still treat event time prediction as a ``second-class citizen'' (e.g., not modeled in the optimization objective) to assist in predicting whether an event will occur. We believe that predicting when an event will occur is as important as predicting whether it will occur, e.g., a task with a 2-hour deadline requires a different response than a task with a 2-day deadline. In short, we believe that a temporal GNN should be able to answer ``Who should I engage with at what tim'' rather than just ``Who should I engage with'' (as shown in Fig.~1). Predicting events and their corresponding times in one model will broaden the application scenario of temporal GNN.}

{Additionally, existing work is often working under the assumption that all events in a temporal graph are fully observed, but in many real-world scenarios, this assumption is difficult to hold. We may miss observing events for a variety of reasons (as the dash lines in Fig.~1). For example, the sensor may fail at a certain time, and events are lost during that time; in social networks, we may only observe interactions on Facebook, but interactions on Twitter or offline are missing. Missing events are part of temporal graphs and should be modeled in the temporal GNNs to improve the performance of the temporal GNNs.}

Therefore, in this work, we propose \MODELNAME, a missing event aware method that uniformly models evolving graph structure and event time to support predicting what will happen and when it will happen. \MODELNAME~is a TPP based method that models the dynamics of both observed and missing events as two coupled TPPs. Both the observation of events and the generation of missing events depend on previously observed events as well as previously generated missing events.

The main contributions are summarized as follows:
\begin{itemize}
	\item We point out the importance of modeling the timing of events and the missing events. We present a problem formulation that unifies modeling missing events, the  of events, and evolving graph structure.
	\item We propose a novel method called \MODELNAME. \MODELNAME~can effectively learn node's structure and temporal features in a uniform embedding. Additionally, \MODELNAME~model the dynamic of observed and missing events as two coupled TPPs and parameterization with log-norm mixture distribution, which makes it can effectively estimate event time.
	\item We conduct extensive experiments on five real-world datasets to demonstrate the superiority of the proposed method \MODELNAME.
\end{itemize}

The remainder of this paper is organized as follows. Section \ref{sec:realted_work}  discusses the related works. Section~\ref{sec:preliminaries} presents related preliminaries. Section~\ref{sec:model} gives a detailed interpretation of \MODELNAME. Section~\ref{sec:experiments} shows all experimental results, as well as detailed analysis and discussion. Section~\ref{sec:conclusion} offers some concluding remarks.

\section{Related Works}\label{sec:realted_work}
\subsection{Neural Temporal Point Process}\label{sec:TPP}
A temporal point process~\cite{PointProcess} is a stochastic process whose realizations consist of discrete events in time $\mathcal{T} = \{t_1, t_2, ...,t_T\}, t_i \le t_{i+1}$. There are two common ways to characterize a temporal point process: intensity-based methods and intensity-free methods.

Intensity-based methods~\cite{Du2016RecurrentMT,DBLP:conf/aaai/XiaoYYZC17,DBLP:conf/nips/MeiE17} are defined in terms of the conditional intensity function $\lambda(t)$, which specifying the dependency of next arrival time $t$ on the history $\mathcal{H}_t=\{t_i \in \mathcal{T} | t_i < t\}$. With the given conditional intensity function $\lambda(t)$, we can obtain the conditional probability density function as follows:
\begin{equation}~\label{eq:r_tpp}
	p(t)=\lambda(t)\underbrace{\exp(-\int_{t_T}^t\lambda(s)\text{d}s)}_{\text{survival function}}
\end{equation}
where the survival function of the process~\cite{Aalen2008SurvivalAE} denotes that no events happen during $[t_T, t)$. Intensity-based methods are very flexible as $\lambda(t)$ can choose different forms (e.g., Poisson, Renewal, Self-correcting, Hawkes, etc.) to capture different interests in different scenarios. However, the drawback of the intensity-based methods are also obvious, as the integral term involved in the survival function leads to there is no closed form for Eq.~\mbox{\eqref{eq:r_tpp}}, and thus requiring Monte Carlo integration.

To reduce the computationally expensive of intensity-based methods, intensity-free methods do not directly model the intensity function. For example, Omi et al.~\cite{FullyNN} introduce a fully neural network (FullyNN) to model the cumulative intensity function $\Lambda(t)=\int_{t_T}^t\lambda(s)\text{d}s$. Shchur et al.~\cite{TPPMixLogNorm} directly estimate conditional density $p(t)$ by utilizing neural density estimation~\cite{papamakarios2019neural}. Similar to \cite{TPPMixLogNorm}, Gupta et al.~\cite{IMTPP} also directly estimate conditional density, and they first introduce missing events in modeling temporal point process. Inspired by \cite{IMTPP}, we introduce the missing events into temporal graph.
 
 For readers who want to learn more about neural temporal point process, we recommend the survey conducted by Shchur et al.~\mbox{\cite{NeuralTPPSurvey}}, which provides more details about the background, models, application scenarios and limitations of using neural networks to solve temporal point process related problems.
 
\subsection{Temporal Graph Neural Networks}\label{sec:GNN}
\subsubsection{Discrete Temporal Graph Neural Networks}
Discrete temporal graph neural networks treat temporal graphs as a sequence of graph snapshots\cite{xue2022dynamic}. Most discrete dynamic embedding approaches\cite{seo2018structured,narayan2018learning,niepert2016learning,chen2019lstm,zheng2020mathnet} focus on the learning representations of entire temporal graphs rather than node representations. Some approaches\cite{sanchez2018graph,chang2016compositional,manessi2020dynamic,jin2019recurrent, cui2022dygcn} are starting to focus on the dynamic representation at node level, they encode each graph snapshot using static embedding approaches\cite{liben2007link,hisano2018semi,hamilton2017inductive} to embed each node, and then combines some time-series models (e.g. LSTM\cite{hochreiter1997long}, RNN\cite{sherstinsky2020fundamentals}) for per node to model the discrete dynamic. 

The main disadvantages of discrete temporal graph neural networks are twofold:
\begin{itemize}
	\item Graph snapshots are generated by aggregating events over a period of time, and a large amount of fine-grained temporal information is lost in the aggregation process.
	\item Discrete temporal GNNs simplify the temporal features. For example, they treat the temporal graph as a sequence of graph snapshots consisting of multiple time steps, while this ignores the fact that the time intervals between adjacent snapshots may be vary.
\end{itemize}
These two drawbacks result in discrete temporal GNNs failing to predict event times and generally performing worse than continuous temporal graph neural networks.

\subsubsection{Continuous Temporal Graph Neural Networks}
Continuous temporal graph neural networks treat temporal graphs as a stream of observed events. These methods can be roughly divided into two main categories: TPP based methods and non-TPP based methods\cite{skarding2020foundations}. 

For the non-TPP based methods, the embedding of interacting nodes is updated by an RNN/transformer based architecture according to the historical information. Representative works of this type of method are JODIE\cite{JODIE}, TGN\cite{TGN}, APAN\cite{APAN} and Streaming GNN\cite{StreamGNN}. JODIE is designed for user-item networks and uses two RNNs to maintain the embedding of each node. With one RNN for users and another one for items. Instead of keeping the embedding of nodes directly, TGN calculates the embedding of nodes at different times by introducing message and memory mechanisms. These methods encode the event time as an auxiliary feature of temporal graph structure, which helps overcome the drawbacks of discrete temporal GNNs but leads to failing to predict the event time.

Know-Evolve\cite{KnowEvolve} is the pioneer in bringing the temporal point processes to dynamic graph representation learning, which models temporal knowledge graph as multi-relational timestamped edges by parameterizing a TPP by a deep recurrent architecture. DyRep\cite{DyRep} is the successor of Know-Evolve. DyRep extends Know-Evolve using TPP to model long-term events and short-term events and introduce aggregation mechanisms. LDG\cite{knyazev2019learning} argues long-term events are often specific by humans and can be suboptimal and expensive to obtain. LDG uses Neural Relational Inference (NRI) model to infer the type of events on the graph and replaces the self-attention originally used in DyRep by generating a temporal attention matrix to better aggregate neighbor information. GHN\cite{han2020graph} is another TPP based approach, which uses an adapted continuous-time LSTM for Hawkes process. TREND\cite{TREND} is a Hawkes process based approach, which captures the individual and collective characteristics of events by integrating event and node dynamics. Theoretically, these TPP based methods can predict the time of events. However, most of them still treat event time prediction as a ``second-class citizen'' to assist in predicting events, e.g. event time is not included in optimization objective and no closed-form event time exception is provided. For a very recent work, EvoKG\cite{EvoKG}, which jointly model the evolving graph structure and timing of events have achieved state-of-the-art performance. However, EvoKG learns structural and temporal embeddings separately, which limits its performance and robustness (detailed discuss in Section~\ref{sec:exp_evokg}).

Finally, it should be noted that all the temporal GNNs mentioned above do not model missing events.

\section{Preliminaries}\label{sec:preliminaries}
\subsection{Notations}
A temporal graph $\mathcal{G}$ can be represented by a sequence of $|E|$ discrete events $\{e_i=(u_i,v_i,t_i,k_i)\}_{i=1}^{|E|}\}$, where $u_i$ and $v_i$ are two nodes involved in the event $e_i$. $t_i$ is the time of the event, and $t_i \le t_j \Leftrightarrow i<j$. $k_i \in \{0, 1\}$ and $k_i=1$ represents the event $e_i$  is observed and $k_i=0$ represents the event $e_i$ is missed\footnote{In this paper, while we focus on undirected graph without attributes, but it's easy to be generalized to directed graphs with attributes}. For convenience, we omit $k$ and use $\mathcal{O}$($\mathcal{M}$) to denotes observed (missing) temporal graph, $o=(u,v,t)$($m=(u,v,t)$) to denotes a observed(missing) event. We preserve boldface lowercase letters (e.g., $\mathbf{u}$) for vectors and boldface capitals for matrices (e.g., $\mathbf{W}$). Frequently-used notations has been summarized in Appendix~\ref{tab:notations}.

\subsection{Problem Definition}
Given a temporal graph $\mathcal{G}=\mathcal{O}\cup\mathcal{M}$ denoted as a sequence of  events, our goal is to model the probability distribution $p(\mathcal{G})$. We assume that both observed and missing events at time $t$ depend on the history of previously generated missing events and observed events. As in practice, only the observed events $\mathcal{O}$ can be evaluated for validation, and missing events are  intractable, we can model the probability distribution as:
\begin{equation}\label{eq:pg}
\begin{aligned}
p_{\theta}(\mathcal{O}_T) & = \prod_{t=1}^T \int_{\mathcal{M}_{t}} p_{\theta}(\mathcal{O}_t|\mathcal{G}^*_{\bar{t}}, \mathcal{M}_{t})p_{\theta}( \mathcal{M}_{t} )\mathrm{d} \mathcal{M}_{t}\\
&= \mathbb{E}_{q_{\phi}} \prod_{t=1}^T \frac{ p_{\theta}(\mathcal{O}_t|\mathcal{G}^*_{\bar{t}}, \mathcal{M}_{t})p_{\theta}(\mathcal{M}_{t} |\mathcal{G}^*_{\bar{t}})}{q_{\phi}(\mathcal{M}_{t} | \mathcal{O}_t, \mathcal{G}^*_{\bar{t}})} \\
\end{aligned}
\end{equation}
where $\bar{t}$ is timestamp of last observed event, and we assume events happen at same time interval $(\bar{t}, t]$ are independent of each other, then we have:
\begin{align}
	p_{\theta}(\mathcal{O}_t|\mathcal{G}^*_{\bar{t}}, \mathcal{M}_{t}) &= \prod\limits_{(u,v,t)\in\mathcal{O}_t}p_{\theta}(u,v,t|\mathcal{G}^*_{\bar{t}}, \mathcal{M}_{t}) \label{eq:pg_o}\\
	p_{\theta}(\mathcal{M}_{t} |\mathcal{G}^*_{\bar{t}}) &= \prod\limits_{(u,v,t)\in\mathcal{M}_t}p_{\theta}(u,v,t|\mathcal{G}^*_{\bar{t}}) \label{eq:pg_pri_m}\\
	q_{\phi}(\mathcal{M}_{t} | \mathcal{O}_t, \mathcal{G}^*_{\bar{t}}) &= \prod\limits_{(u,v,t)\in\mathcal{M}_t}q_{\phi}(u,v,t | \mathcal{O}_t, \mathcal{G}^*_{\bar{t}})\label{eq:pg_pos_m}
\end{align}
where $\mathcal{G}^*_{\bar{t}}$ denotes all observed and generated missing events until time $\bar{t}$; $\mathcal{O}_t$ and $\mathcal{M}_t$ are observed and generated missing events at time interval $(\bar{t}, t]$, and for  convince, we let the time of the missing event with the same subscript lag slightly behind that of the observed event; $q_{\phi}(\cdot)$ is a variational approximation posterior distribution, which aims to generate missing events $\mathcal{M}_{t}$ within the interval $(\bar{t}, t)$, based on the historical events as well as the further observed events $\mathcal{O}_t$.

Then our goal is to maximize the marginal log-likelihood of observed events, which can be achieved by maximizing the following evidence lower bound (ELBO) of the log-likelihood of observed events:
\begin{equation}\label{eq:elbo}
	\begin{aligned}
	&\mathcal{L}(\theta,\phi;\mathcal{O}_{T}) = \mathbb{E}_{q_{\phi}}\sum\limits_{t=1}^T \underbrace{\sum\limits_{(u,v,t)\in\mathcal{O}_t}\log p_{\theta}(u,v,t|\mathcal{G}^*_{\bar{t}}, \mathcal{M}_{t})}_{\mathcal{L}_O} \\
& -\sum\limits_{t=1}^T \underbrace{\sum\limits_{(u,v,t)\in\mathcal{M}_t}\text{KL}(q_{\phi}(u,v,t | \mathcal{O}_t, \mathcal{G}^*_{\bar{t}}) || p_{\theta}(u,v,t|\mathcal{G}^*_{\bar{t}}))}_{\mathcal{L}_M}
	\end{aligned}
\end{equation}

where $\text{KL}(\cdot)$ is the Kullback-Leibler divergence. Similar to \cite{EvoKG}, we further decompose the event joint conditional probability in Eq.~\eqref{eq:elbo} as follows:
\begin{align}
p_{\theta}(u,v,t|\mathcal{G}^*_{\bar{t}}, \mathcal{M}_{t}) &= p_{\theta}(u,v | \mathcal{G}^*_{\bar{t}}, \mathcal{M}_{t})p_{\theta}(t|u, v,\mathcal{G}^*_{\bar{t}}, \mathcal{M}_{t}) \label{eq:pg_event_st_d_o}\\
p_{\theta}(u,v,t|\mathcal{G}^*_{\bar{t}}) &= p_{\theta}(u,v | \mathcal{G}^*_{\bar{t}})p_{\theta}(t|u, v,\mathcal{G}^*_{\bar{t}}) \label{eq:pg_event_st_d_pri_m}\\
q_{\phi}(u,v,t | \mathcal{O}_t, \mathcal{G}^*_{\bar{t}}) &= q_{\phi}(u,v | \mathcal{O}_t, \mathcal{G}^*_{\bar{t}}) q_{\phi}(t | u,v,\mathcal{O}_t, \mathcal{G}^*_{\bar{t}}) \label{eq:pg_event_st_d_pos_m}
\end{align}

Then, by chain rule for relative entropy, the KL divergence term in Eq.~\eqref{eq:elbo} can be denoted as:
\begin{equation}\label{eq:KL_all}
\begin{split}
	&\text{KL}(q_{\phi}(u,v,t | \mathcal{O}_t, \mathcal{G}^*_{\bar{t}}) || p_{\theta}(u,v,t|\mathcal{G}^*_{\bar{t}})) = \\
&\qquad \qquad \text{KL}(q_{\phi}(u,v| \mathcal{O}_t, \mathcal{G}^*_{\bar{t}}) || p_{\theta}(u,v|\mathcal{G}^*_{\bar{t}})) \\
&\qquad \qquad +\text{KL}(q_{\phi}(t | u,v,\mathcal{O}_t, \mathcal{G}^*_{\bar{t}}) || p_{\theta}(t|u,v,\mathcal{G}^*_{\bar{t}}))
\end{split}	
\end{equation}

Above equations suggest the components of our model from two perspectives corresponding to the challenges mentioned in Introduction. From the event perspective, the ELBO in Eq.~\eqref{eq:elbo} suggests our model consist of the three components: one TPP for observed events ($p_{\theta}(u,v,t|\mathcal{G}^*_{\bar{t}}, \mathcal{M}_{t})$), one prior TPP for missing events ($p_{\theta}(u,v,t|\mathcal{G}^*_{\bar{t}})$) and one posterior TPP for missing events ($q_{\phi}(u,v | \mathcal{O}_t, \mathcal{G}^*_{\bar{t}})$). From the temporal graph perspective, Eq.(6) - Eq.(8) suggest our model consist of two components: graph structure modeling components ($p_{\theta}(u,v | \mathcal{G}^*_{\bar{t}}$), $p_{\theta}(u,v | \mathcal{G}^*_{\bar{t}})$ and $q_{\phi}(u,v | \mathcal{O}_t, \mathcal{G}^*_{\bar{t}})$) and event time modeling components ($p_{\theta}(t| u,v, \mathcal{G}^*_{\bar{t}}$), $p_{\theta}(t | u,v, \mathcal{G}^*_{\bar{t}})$ and $q_{\phi}(t |u,v, \mathcal{O}_t, \mathcal{G}^*_{\bar{t}})$).

\section{Modeling A Temporal Graph}\label{sec:model}
Modeling a temporal graph is equivalent to modeling each observed and missing event in the temporal graph, i.e., Eq.(7)-Eq. (9). To model these events, we learn the observed (missing) embeddings of nodes. Unlike EvoKG\cite{EvoKG}, which learns the structural embeddings and temporal node embeddings separately, we learn the time-evolving structural dynamics and temporal characteristics of nodes in unified embeddings. The unified embeddings make the model treat events as a whole, allowing the model to have higher consistency in predicting link and predicting time, i.e., the model can obtain optimal link prediction performance and time prediction performance with the same set of parameters.

We utilize message passing framework to learn node observed (missing) embeddings. Given concurrent observed events $\mathcal{O}_t$, we summarise node $u$'s observed embedding as follows:
\begin{equation}\label{eq:massege_passing_o}
	\mathbf{o}^{l+1,t}_u = \mathbf{W}^{l}_s\mathbf{o}^{l,t}_u + \frac{1}{|\mathcal{N}^{\mathcal{O}_t}_u|}\sum\limits_{(u,v,t)\in \mathcal{O}_t} \mathbf{W}^{l}_n \mathbf{o}^{l,t}_v+\mathbf{W}^l_t(t-\bar{t}^o_{u,v}) 
\end{equation}
where $\mathbf{o}^{l+1,t}_u$ is the observed embeddings of node $u$ learned by $l$-th layer of GNN and $\mathbf{o}^{0,t}_u$ is set to static embedding $\mathbf{o}_u$; $\mathcal{N}^{\mathcal{O}_t}_u$ is node $u$'s neighbors in $\mathcal{O}_t$ and $\bar{t}^o_{u,v} = \max(\bar{t}^o_u,\bar{t}^o_v)$ is the last time of node $u$ or node $v$ involved in an observed event. $\mathbf{W}^{l}_{\boldsymbol{\bullet}}$ are learnable weights in the $l$-th layer GNN. Specially, $\mathbf{W}^{\boldsymbol{\bullet}}_s$ models the self-evolution, which indicates a node evolves with respect to its previous state; $\mathbf{W}^{\bullet}_n$ models the structural message passing from neighborhoods; $\mathbf{W}^{\bullet}_t$ models the event time impact. So that Eq.~\eqref{eq:massege_passing_o} encodes the structural and temporal information of the observed events in a uniform embedding.

For the generated missing events $\mathcal{M}_t$ (the generation algorithm is described in Section~\ref{sec:generation}), we summarise node $u$'s missing embedding as follows:
\begin{equation}\label{eq:massege_passing_m}
	\mathbf{m}^{l+1,t}_u {=} \mathbf{U}^{l}_s\mathbf{m}^{l,t}_u {+} \frac{1}{|\mathcal{N}^{\mathcal{M}_t}_u|}\quad\sum\limits_{\mathclap{(u,v,t')\in \mathcal{M}_t}} \mathbf{U}^{l}_n \mathbf{m}^{l,t}_v{+}\mathbf{U}^l_t(t'-\bar{t}^m_{u,v}) 
\end{equation}
Eq.~\eqref{eq:massege_passing_m} is similar to Eq.~\eqref{eq:massege_passing_o}, except that the generated missing event $m_1$ and $m_2$ may have different time $t_1'$ and $t_2'$ ($\bar{t} < t_1' \le t$, $\bar{t} < t_2' \le t$ ). $\bar{t}^m_{u,v}=\max(\bar{t}^m_u,\bar{t}^m_v)$ is the last time of node $u$ or node $v$ involved in an generated missing event.


%
$\mathbf{o}^{L,t}_u$ ($\mathbf{m}^{L,t}_u$ )only summarizes the interaction patterns in concurrent events $\mathcal{O}_t$($\mathcal{M}_t$). To further model the dynamic of temporal updates, we utilize GRU~\cite{GRU} to capture the temporal dependency among node embeddings:
\begin{align}
	\mathbf{o}^{*,t}_u &= \text{GRU}(\mathbf{o}^{L,t}_u,\mathbf{o}^{*,\bar{t}}_u) \label{eq:o_star_t}\\
	\mathbf{m}^{*,t}_u &= \text{GRU}(\mathbf{m}^{L,t}_u,\mathbf{m}^{*,\bar{t}}_u) \label{eq:m_star_t}\\
	\mathbf{g}^{*,t}_u &= [\mathbf{o}^{*,t}_u; \mathbf{m}^{*,t}_u] \label{eq:g_star_t}
\end{align}
where $\mathbf{o}^{*,t}_u$ and $\mathbf{m}^{*,t}_u$ are observed evolving-embedding and missing evolving-embedding of node $u$. $\mathbf{g}^{*,t}_u$ is the node $u$'s evolving-embedding. $[;]$ is a concatenation operation.

To enhance the structural information in the node evolve-embeddings, we concatenate the static embeddings and evolving-embeddings as follows:
\begin{equation}\label{eq:g_s}
	\mathbf{\bar{o}}^{t}_u = [\mathbf{o}_u;\mathbf{o}^{*,t}_u] 	\quad 
	\mathbf{\bar{m}}^{t}_u = [\mathbf{m}_u;\mathbf{m}^{*,t}_u] \quad 
	\mathbf{\bar{g}}^{t}_u = [\mathbf{\bar{o}}^{t}_u;\mathbf{\bar{m}}^{t}_u] 
\end{equation}
	
Graph-level embeddings on observed events, missing events and all events are obtained by using element-wise max pooling:
\begin{align}
	\mathbf{\bar{o}}^{t} &= \max(\{\mathbf{\bar{o}}^{t}_u | u \in nodes(\mathcal{O}^*_{\bar{t}})\}) \label{eq:o_t_bar}\\
	\mathbf{\bar{m}}^{t} &= \max(\{\mathbf{\bar{m}}^{t}_u | u \in nodes(\mathcal{M}^*_{\bar{t}})\}) \label{eq:m_t_bar}\\
	\mathbf{\bar{g}}^{t} &= \max(\{\mathbf{\bar{g}}^{t}_u | u \in nodes(\mathcal{G}^*_{\bar{t}})\}) \label{eq:g_t_bar}
\end{align}

Based on the above learned embeddings, we give detail description about parameterization of three temporal point processes in Section~\ref{sec:param_tpp_o} - Section~\ref{sec:param_tpp_m_pos}.

\subsection{Parameterization of TPP for observed events}\label{sec:param_tpp_o}
Based on Eq.~\eqref{eq:pg_event_st_d_o}, we first parameterize the graph structure and then parameterize the event time. 

For the graph structure part, it is not practical to direct parameterize the event $(u,v)$, we decompose $p_\theta(u,v|\mathcal{G}_{\bar{t}}^*, \mathcal{M}_t)$ as follows:
\begin{equation}
	p_\theta(u,v|\mathcal{G}_{\bar{t}}^*, \mathcal{M}_t)	 {=} p_\theta(u|\mathcal{G}_{\bar{t}}^*, \mathcal{M}_t)	p_\theta(v|u,\mathcal{G}_{\bar{t}}^*, \mathcal{M}_t)	
\end{equation}
and parameterize each component separately: 
\begin{align}
	p_\theta(u|\mathcal{G}_{\bar{t}}^*, \mathcal{M}_t)	&= \text{softmax}(\text{MLP}([\mathbf{\bar{g}}^{t}])) \label{o_u_p}\\
	p_\theta(v|u,\mathcal{G}_{\bar{t}}^*, \mathcal{M}_t) &= \text{softmax}(\text{MLP}([\mathbf{\bar{g}}^{t}_u;\mathbf{\bar{g}}^{t}])) \label{o_v_p}
\end{align}
where MLP is the abbreviation of multilayer perceptron.

For the event time part, follow~\cite{TPPMixLogNorm,EvoKG}, we directly model event time's probability conditional function (PDF) $p(t)$ rather than the conditional intensity function $\lambda(t)$, which makes \MODELNAME~  more flex and have a closed-form likelihood and expectation. We first conduct the context vector for the observed event as follows:
\begin{equation}\label{eq:context_o}
	\mathbf{c} = [\mathbf{g}^{*,t}_u; \mathbf{g}^{*,t}_v]
\end{equation}
and then obtain the {$K$-dimension ($K$ mixture component)} params of the log-norm mixture distribution as follows: 
\begin{equation}\label{eq:tpp_params_o}
	\boldsymbol{\omega}{=}\text{softmax}(\text{MLP}(\mathbf{c}))\quad \boldsymbol{\mu}{=}\text{MLP}(\mathbf{c})\quad \boldsymbol{\sigma}{=}\exp(\text{MLP}(\mathbf{c})) 
\end{equation}
With the above-mentioned params, the PDF of the time $\tau=t-\bar{t}^o_{u,v}$ as follows:  
\begin{equation}\label{eq:o_t_p}
	\begin{aligned}
		p_{\theta}&(\tau|u, v,\mathcal{G}^*_{\bar{t}}, \mathcal{M}_{t}) = p_{\theta}(\tau|\boldsymbol{\omega},\boldsymbol{\mu},\boldsymbol{\sigma}) \\
		&= \sum\limits_{k=0}^K\frac{\omega_k}{\tau\sigma_k\sqrt{2\pi}}\exp(-\frac{(\log\tau-\mu_k)^2}{2\sigma_k})
	\end{aligned}
\end{equation}
It is reasonable to model $\tau$ rather than directly model $t$, as $t$ can be very large thus making the neural network confused.

\subsection{Parameterization of prior TPP for missing events}
For the graph structure part, same as observed events, we decompose $p_\theta(u,v|\mathcal{G}_{\bar{t}}^*)$ as follows:
\begin{equation}
	p_\theta(u,v|\mathcal{G}_{\bar{t}}^*)	 = p_\theta(u|\mathcal{G}_{\bar{t}}^*)	p_\theta(v|u,\mathcal{G}_{\bar{t}}^*)	
\end{equation}
and parameterize each component separately:
\begin{align}
	p_\theta(u|\mathcal{G}_{\bar{t}}^*)	&= \text{softmax}(\text{MLP}([\mathbf{\bar{g}}^{\bar{t}}])) \label{eq:m_u_pri}\\
	p_\theta(v|u,\mathcal{G}_{\bar{t}}^*) &= \text{softmax}(\text{MLP}([\mathbf{\bar{g}}^{\bar{t}}_u;\mathbf{\bar{g}}^{\bar{t}}])) \label{eq:m_v_pri}
\end{align}

For the event time part, we also use log-norm mixture distribution to model missing event time's PDF. We first conduct the context vector $\mathbf{c}^{\theta}=[\mathbf{g}^{*,\bar{t}}_u; \mathbf{g}^{*,\bar{t}}_v]$ and obtain the params of the log-norm mixture params $\boldsymbol{\omega}^{\theta},\boldsymbol{\mu}^{\theta},\boldsymbol{\sigma}^{\theta}$ same as Eq.~\eqref{eq:tpp_params_o} with $\mathbf{c}^{\theta}$ as input. Then the PDF of time $\Delta = t' - \bar{t}$ as follows:
\begin{equation}\label{eq:m_t_pri}
	\begin{aligned}
		p_{\theta}(\Delta|&u, v,\mathcal{G}^*_{\bar{t}}) = p_{\theta}(\Delta|\boldsymbol{\omega}^{\theta},\boldsymbol{\mu}^{\theta},\boldsymbol{\sigma}^{\theta}) \\
		&= \sum\limits_{k=0}^K\frac{\omega^\theta_k}{\Delta\sigma^\theta_k\sqrt{2\pi}}\exp(-\frac{(\log\Delta-\mu^\theta_k)^2}{2\sigma^\theta_k})
	\end{aligned}
\end{equation}
Model time $\Delta$ makes true the missing event time is large than the latest observed time $\bar{t}$.

\subsection{Parameterization of posterior TPP for missing events}\label{sec:param_tpp_m_pos}
For the graph structure part, we decompose $q_\phi(u,v|\mathcal{G}_{\bar{t}}^*, \mathcal{O}_t)$ as follows:
\begin{equation}
	q_\phi(u,v|\mathcal{G}_{\bar{t}}^*, \mathcal{O}_t)	 {=} q_\phi(u|\mathcal{G}_{\bar{t}}^*, \mathcal{O}_t)	q_\phi(v|u,\mathcal{G}_{\bar{t}}^*, \mathcal{O}_t)	
\end{equation}
and then parameterize each component separately:
\begin{align}
	q_\phi(u|\mathcal{G}_{\bar{t}}^*, \mathcal{O}_t)	&= \text{softmax}(\text{MLP}([\mathbf{\bar{g}}^{\bar{t}};\mathbf{\bar{o}}^{t}])) \label{eq:m_u_pos}\\
	q_\phi(v|u,\mathcal{G}_{\bar{t}}^*, \mathcal{M}_t) &{=} \text{softmax}(\text{MLP}([\mathbf{\bar{g}}^{\bar{t}}_u;\mathbf{\bar{g}}^{\bar{t}};\mathbf{\bar{o}}_u^{t};\mathbf{\bar{o}}^{t}])) \label{eq:m_v_pos}
\end{align}

For the event time part, we conduct the context vector $\mathbf{c}^{\phi}=[\mathbf{g}^{*,\bar{t}}_u; \mathbf{g}^{*,\bar{t}}_v;\mathbf{o}^{*,t}_u; \mathbf{o}^{*,t}_v]$ and obtain the params of the log-norm mixture params $\boldsymbol{\omega}^{\phi},\boldsymbol{\mu}^{\phi},\boldsymbol{\sigma}^{\phi}$ same as Eq.~\eqref{eq:tpp_params_o} with $\mathbf{c}^{\phi}$ as input. Then we have the density of the inter-arrival time $\Delta$ as:
\begin{equation}\label{eq:m_t_pos}
	\begin{aligned}
		q_{\phi}&(\Delta|u, v,\mathcal{G}^*_{\bar{t}},\mathcal{O}^*_{t}) = q_{\phi}(\Delta|\boldsymbol{\omega}^{\phi},\boldsymbol{\mu}^{\phi},\boldsymbol{\sigma}^{\phi})\odot \mathbb{I}(\Delta {<} t{-}\bar{t}) \\
		&= \mathbb{I}(\Delta {<} t{-}\bar{t}){\odot}\sum\limits_{k=0}^K\frac{\omega^\phi_k}{\Delta\sigma^\phi_k\sqrt{2\pi}}\exp({-}\frac{(\log\Delta-\mu^\phi_k)^2}{2\sigma^\phi_k})
	\end{aligned}
\end{equation}
where $\mathbb{I}(\cdot)$ is an indicator function. Different with density of observed events inter-arrival time (Eq.~\eqref{eq:o_t_p}) and prior missing events inter-arrival time(Eq.~\eqref{eq:m_t_pos}), the density of posterior missing events inter-arrival time is a truncated log-norm mixture distribution, which makes the missing events time $t'$ sampled from this distribution satisfy $\bar{t} < t' <= t$.

\subsection{Generating missing events}\label{sec:generation}
\begin{algorithm}[h]
\caption{Missing Events Generation}\label{alg:generation}
\SetKwInOut{Input}{Input}
\SetKwInOut{Output}{Output}
\Input{Current time $t$; Last observed time $\bar{t}$; Number of sampled missing events ratio $Q$.}

\Output{Generated missing events $\mathcal{M}_t$ and the KL divergence $\mathcal{L}_{M}$ between prior and posterior TPP for missing events.}
$\mathcal{M}_t \gets \{\}$ \\
$\mathcal{L}_{M} \gets 0$ \\
\tcc{executed in parallel}
\For{i=$1, \dots, Q|\mathcal{O}_t|$}{
	\tcc{Sample node $u$}
	Compute $p_\theta(\cdot|\mathcal{G}_{\bar{t}}^*)$ based on Eq.~\eqref{eq:m_u_pri}.\\
	Compute $q_\phi(\cdot|\mathcal{G}_{\bar{t}}^*, \mathcal{O}_t)$ based on Eq.~\eqref{eq:m_u_pos}.\\
	$\mathcal{L}_{M} \gets \mathcal{L}_{M} + \text{KL}(q_\phi(\cdot|\mathcal{G}_{\bar{t}}^*, \mathcal{O}_t) ||p_\theta(\cdot|\mathcal{G}_{\bar{t}}^*))$ \\
	$u \thicksim q_\phi(\cdot|\mathcal{G}_{\bar{t}}^*, \mathcal{O}_t)$ \\
	\tcc{Sample node $v$}
	Compute $p_\theta(\cdot|u,\mathcal{G}_{\bar{t}}^*)$ based on Eq.~\eqref{eq:m_v_pri}. \\
	Compute $q_\phi(\cdot|u,\mathcal{G}_{\bar{t}}^*, \mathcal{M}_t)$ based on Eq.~\eqref{eq:m_v_pos}. \\
	$\mathcal{L}_{M} {\gets} \mathcal{L}_{M} + \text{KL}(q_\phi(\cdot|u,\mathcal{G}_{\bar{t}}^*, \mathcal{O}_t) || p_\theta(\cdot|u,\mathcal{G}_{\bar{t}}^*))$ \\
	$v \thicksim q_\phi(\cdot|u,\mathcal{G}_{\bar{t}}^*, \mathcal{O}_t)$ \\
	\tcc{Sample time interval $\Delta$}
	Compute $p_\theta(\cdot|u,v,\mathcal{G}_{\bar{t}}^*)$ based on Eq.~\eqref{eq:m_t_pri}\\
	Compute $q_\phi(\cdot|u,v,\mathcal{G}_{\bar{t}}^*, \mathcal{O}_t))$ based on Eq.~\eqref{eq:m_t_pri}\\
	$\mathcal{L}_{M} {\gets} \mathcal{L}_{M} + \text{KL}(q_\phi(\cdot|u,v,\mathcal{G}_{\bar{t}}^*, \mathcal{O}_t) || p_\theta(\cdot|u,v,\mathcal{G}_{\bar{t}}^*))$ \\
	$\Delta \thicksim q_\phi(\cdot|u,v,\mathcal{G}_{\bar{t}}^*, \mathcal{O}_t))$ \\
	$t' \gets \bar{t} + \Delta$ \\
	$\mathcal{M}_t \gets \mathcal{M}_t \cup \{(u,v,t')\}$ \\
}

\Return $\mathcal{M}_t$, $\mathcal{L}_{M}$
\end{algorithm} 

Algorithm~\ref{alg:generation} illustrates the missing events generation steps. For each missing event $(u,v,t')$, we sample $u$, $v$ and $\Delta$ in order according to the corresponding posterior probability distribution. Unlike the recursive sampling of missing events in~\cite{IMTPP}, we directly sample $Q|\mathcal{O}_t|$ missing events in parallel to improve model efficiency.

During the generation of missing events, we also calculate the KL divergence term in ELBO. For the discrete probability distribution $p$ ($p_\theta(\cdot|\mathcal{G}_{\bar{t}}^*)$ and $p_\theta(\cdot|u,\mathcal{G}_{\bar{t}}^*)$) and $q$ ($q_\phi(\cdot|\mathcal{G}_{\bar{t}}^*, \mathcal{O}_t)$ and $q_\phi(\cdot|u,\mathcal{G}_{\bar{t}}^*, \mathcal{M}_t)$ ), the KL divergence is defined as:
\begin{equation}
	KL(q||p) = \sum_{x \in nodes(\mathcal{G})}q(x)\log\frac{q(x)}{p(x)}
\end{equation}
For $\text{KL}(q_\phi(\cdot|u,v,\mathcal{G}_{\bar{t}}^*, \mathcal{O}_t) || p_\theta(\cdot|u,v,\mathcal{G}_{\bar{t}}^*))$, there is no closed-form expression, so we estimate it using Monte Carlo sampling:
\begin{equation}
\begin{aligned}
		\text{KL}(q_\phi(\cdot|u,v,\mathcal{G}_{\bar{t}}^*, \mathcal{O}_t) || &p_\theta(\cdot|u,v,\mathcal{G}_{\bar{t}}^*)) \\
	&=\frac{1}{n}\sum_{i=1}^n\log\frac{q_\phi(\Delta_i|u,v,\mathcal{G}_{\bar{t}}^*, \mathcal{O}_t)}{p_\theta(\Delta_i|u,v,\mathcal{G}_{\bar{t}}^*)}
\end{aligned}
\end{equation}
where $\{\Delta_i\}_{i=1}^n$ are independently sampled from $q_\phi(\cdot|u,v,\mathcal{G}_{\bar{t}}^*, \mathcal{O}_t)$. And luckily, we found that in practice $n$ does not need to take a very large number, usually 10 is enough.

\subsection{Parameter Learning}
We optimize \MODELNAME~by maximizing the ELBO:
\begin{equation}
	\max\limits_{\theta,\phi}\mathcal{L}(\theta,\phi;\mathcal{O}_T)
\end{equation}
The parameter learning algorithm are show in Algorithm~\ref{alg:learning}. Directly training on the entire event sequence is inefficient for the following issues:
\begin{enumerate}
	\item Computing resource issue: maintain the entire history for each node's various embedding for backpropagation requires high computation and memory cost.
	\item Gradient related issue: \MODELNAME~contains GRUs, which will suffer from gradient explosion or gradient disappearance.
\end{enumerate}
 to address the above-mentioned issues, backpropagation through time (BPTT) training is conducted over the entire event sequence.

\begin{algorithm}[h]
\caption{Parameter Learning}\label{alg:learning}
\SetKwInOut{Input}{Input}

\Input{Observed events $\mathcal{O}$; Maximum number of epochs $max\_epochs$; number $L$ of GNN layers; Number of time steps $b$ for BPTT.
}

\For{epoch = $1, \dots,max\_epochs$ }{
	\For{$t \in$ Timestep($\mathcal{O}$)}{
		Compute $\mathbf{o}_{\bullet}^{L,t}$ and $\mathbf{o}_{\bullet}^{*,t}$ based on Eq.~\eqref{eq:massege_passing_o} and Eq.~\eqref{eq:o_star_t}. \\
		Generate missing events $\mathcal{M}_t$ and compute $\mathcal{L}_{M}$ based on Algorithm~\ref{alg:generation}. \\
		Compute $\mathbf{m}_{\bullet}^{L,t}$ and $\mathbf{m}_{\bullet}^{*,t}$ based on Eq.~\eqref{eq:massege_passing_m} and Eq.~\eqref{eq:m_star_t}. \\
		Compute $\mathcal{L}_O$ based on Eq.~\eqref{eq:elbo}, Eq.~\eqref{o_u_p}, Eq.~\eqref{o_v_p} and Eq.~\eqref{eq:o_t_p} \\
		Compute $\mathcal{L} \gets \mathcal{L} + \mathcal{L}_O + \mathcal{L}_M$ \\
		Optimize model parameters every $b$ time steps.
	}
}

\end{algorithm}

\subsection{Complexity Analysis}
In this section, we analyze the complexity of MTGN. Particularly, we use $\mathcal{O}$ to stand complexity analysis symbols instead of observed events. We use $N_t$ ($QN_t$) to denote the number of observed events (generated missing events) at time $t\in\{1, \dots, T\}$, and $|V|$ is used to denote the node numbers in the temporal graph. The time complexity of GNN used to learn node observed (missing) embedding is $\mathcal{O}(LN_{max})$ ($\mathcal{O}(LQN_{max})$), where $N_{max}$ is maximum observed events at different timestamp. The time complexity of GRU used in Eq.(13) and Eq.(14) is $\mathcal{O}(|V|)$, while we can only update the involved nodes to reduce the complexity to $\mathcal{O}(N_{max})$. The time complexity of missing events generation algorithm at each time is $\mathcal{O}(LQN_{max})$.

Overall, the time complexity of MTGN is $\mathcal{O}(T((L+2LQ+1)N_{max}))=\mathcal{O}(TN_{max})$. This indicates that MTGN is linear to the total number of observed events, and proves that MTGN can be applied to large-scale temporal graphs.
	
\section{Experiments}\label{sec:experiments}
In this section, we present a comprehensive set of experiments to demonstrate the effectiveness of \MODELNAME.

\subsection{Datasets}
We use five publicly available datasets for experiments. The datasets are described as follows:
\begin{itemize}
	\item \textbf{LSED}\footnote{{https://github.com/HIT-ICES/LSED}}. LSED is a dynamic interaction network built upon the openly accessible news aiming to provide a relatively complete and accurate evolution trace of ``Chinese Internet+'' service ecosystem. Each node represents a stakeholder or a service, and each edge denotes two nodes that were mentioned in the same news. The missing events in LSED could be unreported news. Our work is motivated while working on this dataset.
	\item \textbf{ENRON}~\cite{ENRON}. ENRON is a large set of email messages. Each node represents an employee of Enron corporation, and each edge denotes an email sent from one employee to another. The missing events in ENRON could be offline face-to-face proximity.
	\item \textbf{UCI}~\cite{UCI}. UCI is an online community of students from the University of California, Irvine. Each node represents a student, and edge denotes a message sent or received between users. The missing events in UCI could be offline face-to-face proximity or message sent or received on other social platforms.
	\item \textbf{HYPERTEXT}~\cite{HYPERTEXT}. HYPERTEXT is a human contact network of face-to-face proximity. Each node denotes a person who attended in the ACM Hypertext 2009 conference. Edges denote two people have a conversation during a certain time interval.
	\item \textbf{RT-POL}~\cite{RT-POL}. RT-POL is a network of political communication on Twitter. Nodes are Twitter users and edges denote whether the users have retweeted each other. RT-POL exhibits a highly segregated partisan structure, which is very different from the user communication networks.
\end{itemize}

\begin{table}[h]
\centering
\caption{Basic statistic of five real-world datasets. \%Inductive events indicates what percentage of events in the test events that not appear in train events. Time interval denotes the minimum duration between two observed events.}\label{tab:dataset}
\begin{adjustbox}{width=\linewidth}
\begin{tabular}{cccccc}
\hline
Dataset   & \begin{tabular}[c]{@{}c@{}}\#Train\\ Events\end{tabular} & \begin{tabular}[c]{@{}c@{}}\#Test\\ Events\end{tabular} & \begin{tabular}[c]{@{}c@{}}\%Inductive\\ Events\end{tabular} & \#Nodes & \begin{tabular}[c]{@{}c@{}}Time \\ Interval\end{tabular} \\ \hline
LSED      & 8,111                                                     & 1,266                                                    & 61.2\%                                                       & 4,301       & 1 day                                                    \\
ENRON     & 15,083                                                    & 997                                                     & 28.3\%                                                       & 151        & 1 day                                                    \\
UCI       & 27,219                                                    & 3,336                                                    & 77.1\%                                                       & 1,899       & 1 day                                                    \\
HYPERTEXT & 3,392                                                     & 714                                                     & 45.1\%                                                       & 113        & 1 hour                                                   \\
RT-POL    & 47,492                                                    & 10,629                                                   & 83.1\%                                                       & 18,470      & 1 hour                                                   \\ \hline
\end{tabular}
\end{adjustbox}
\end{table}

The basic statistics of these datasets are summarized in Table.~\ref{tab:dataset}. From Table.~\ref{tab:dataset}, it is clear that the 5 selected datasets represented differently characterized temporal graphs. So the experiments conduct on these 5 datasets are able to demonstrate the generalization capability of \MODELNAME. It should be noted that for the events involved two same nodes in the test set, we only keep the earliest event time for testing.

\subsection{Baselines}
We employ two static GNNs (GCN~\cite{GCN} and GAT~\cite{GAT}), two discrete temporal GNNs (DySAT~\cite{DySAT} and EvolveGCN~\cite{EvolveGCN}) and two continuous temporal GNNs (KnowEvolve~\cite{KnowEvolve} and EvoKG~\cite{EvoKG}) as baselines. These baselines have been introduced in Section~\ref{sec:realted_work}. Table.~\ref{tab:qualitative} comparison baselines and our model.
\begin{table*}
	\centering
	\caption{Comparison of \MODELNAME~ with state-of-the-art baselines.}\label{tab:qualitative}
	\begin{tabular}{cccccccc} 
\hline
Key Properties         & \MODELNAME & GCN                   & GAT                   & DySAT                 & EvolveGCN             & KnowEvolve            & EvoKG                  \\ 
\hline
Model graph structure   & $\surd$                  & $\surd$               & $\surd$  & $\surd$  & $\surd$  & $\surd$  & $\surd$   \\
Model event time$t$    & $\surd$                   & $\times$ & $\times$ & $\surd$  & $\surd$  & $\surd$  & $\surd$   \\
Predict event time~$t$ & $\surd$                   & $\times$ & $\times$ & $\times$ & $\times$ & $\surd$  & $\surd$   \\
Model missing event    & $\surd$                   & $\times$ & $\times$ & $\times$ & $\times$ & $\times$ & $\times$  \\
\hline
\end{tabular}
\end{table*}

For the static GNNs, we generate a cumulative graph from observed events in the training set, and edge timestamps are ignored. For the discrete temporal GNNs, we uniformly generate 10 snapshots of the same time span and set the time window to 3.

Please refer to Appendix~\ref{apd:implement} for more details on the hyperparameters of the baselines and \MODELNAME.

\subsection{Tasks and Metrics}
We study the effectiveness of \MODELNAME~by evaluating our model and baselines on the following two tasks:
\begin{enumerate}
	\item \textbf{Further link prediction}. For a given test observed event $(u,?,t)$, we replace $?$ with nodes in the graph and compute the score (Eq.~\eqref{o_v_p} for \MODELNAME). Then, we rank all possible nodes in descending order of the score and report the rank of the ground truth node. In this paper, we report HITS@\{3,5,10\}, which is the percentage of ground truth nodes ranked in the top 3, 5 and 10 predictions. 

	\item \textbf{Event time prediction.} For a given test observed event $(u,v,?)$, the task aims to predict when the event is next observed. The next time point $\hat{t}$ can be obtained by compute the expected value of Eq.~\eqref{eq:o_t_p} as follows:
	\begin{equation}
		\mathbb{E}_{p_{\theta,\tau}}[\tau] = \sum\limits_{k=1}^K\omega_k\exp(\mu_k+\sigma_k^2/2)
	\end{equation}
	In this paper, we report Mean Absolute Error (MAE), which is the average of the absolute difference between the predicted time and the ground truth.
\end{enumerate}

\subsection{Experiment Results}
\subsubsection{Overall}
\begin{table*}
\centering
\caption{The further link prediction results and event time prediction results. The best result is marked in \textbf{bold} and the second best result is marked with \uline{underline}. * indicates the best result significantly outperforms the second best results based on two-tail $t\text{-}test$($p\text{-}value < 0.05$)}\label{tab:overview}
\arrayrulecolor{black}
\begin{tabular}{ccccccccc} 
\hline
                           &         & GCN           & GAT           & DySAT         & EvolveGCN             & KnowEvolve     & EvoKG                  & \begin{tabular}[c]{@{}c@{}}MTGN\\(improv.)\end{tabular}                   \\ 
\hline
\multirow{4}{*}{LSED}      & MAE     & $\times$             & $\times$             & $\times$             & $\times$                     & 866.265$\pm$0.383 & \uline{66.284}$\pm$3.141  & \begin{tabular}[c]{@{}c@{}}\textbf{21.717}$^*$$\pm$0.866\\(\textit{+67.236\%})\end{tabular}  \\
                           & HITS@3  & 2.045$\pm$0.487  & 0.645$\pm$0.210  & 0.226$\pm$0.010  & \uline{3.620}$\pm$0.178   & 0.039$\pm$0.088   & 1.715$\pm$0.418           & \begin{tabular}[c]{@{}c@{}}\textbf{7.700}$^*$$\pm$0.254\\(\textit{+112.355\%})\end{tabular}  \\
                           & HITS@5  & 3.327$\pm$0.406  & 1.980$\pm$0.602  & 0.444$\pm$0.099  & \uline{7.344}$\pm$0.033  & 0.039$\pm$0.088   & 3.189$\pm$0.434           & \begin{tabular}[c]{@{}c@{}}\textbf{8.802}$^*$$\pm$0.482\\(\textit{+19.853\%})\end{tabular}   \\
                           & HITS@10 & 7.467$\pm$0.204  & 4.041$\pm$0.537  & 0.939$\pm$0.021  & \uline{10.990}$\pm$0.055 & 0.172$\pm$0.172   & 6.749$\pm$0.545           & \begin{tabular}[c]{@{}c@{}}\textbf{13.91}$^*$$\pm$0.214\\(\textit{+26.570\%})\end{tabular}   \\ 
\arrayrulecolor{black}\cline{1-1}\arrayrulecolor{black}\cline{2-9}
\multirow{4}{*}{ENRON}     & MAE     & $\times$             & $\times$             & $\times$             & $\times$                     & 926.480$\pm$0.002 & \textbf{11.006}$\pm$1.472 & \begin{tabular}[c]{@{}c@{}}\uline{11.450}$\pm$0.044\\(\textit{-4.034\%})\end{tabular}    \\
                           & HITS@3  & 15.800$\pm$0.543 & 14.920$\pm$0.672 & 11.950$\pm$0.056 & 12.962$\pm$0.010         & 0.981$\pm$0.000   & \uline{24.144}$\pm$0.702  & \begin{tabular}[c]{@{}c@{}}\textbf{38.794}$^*$$\pm$0.500\\(\textit{+60.678\%})\end{tabular}  \\
                           & HITS@5  & 26.490$\pm$0.783 & 26.498$\pm$0.515 & 19.112$\pm$0.029 & 21.360$\pm$0.037         & 1.679$\pm$0.000   & \uline{35.412}$\pm$1.459  & \begin{tabular}[c]{@{}c@{}}\textbf{52.710}$^*$$\pm$0.064\\(\textit{+48.848\%})\end{tabular}  \\
                           & HITS@10 & 45.700$\pm$0.890 & 50.922$\pm$0.368 & 35.664$\pm$0.044 & 36.984$\pm$0.013         & 3.398$\pm$0.000   & \uline{52.362}$\pm$2.036  & \begin{tabular}[c]{@{}c@{}}\textbf{71.060}$^*$$\pm$0.147\\(\textit{+35.709\%})\end{tabular}  \\ 
\arrayrulecolor{black}\cline{1-1}\arrayrulecolor{black}\cline{2-9}
\multirow{4}{*}{UCI}       & MAE     & $\times$             & $\times$             & $\times$             & $\times$                     & 10.9052$\pm$0.480 & \uline{3.661}$\pm$0.026   & \begin{tabular}[c]{@{}c@{}}\textbf{2.601}$^*$$\pm$0.000\\(\textit{+28.954\%})\end{tabular}   \\
                           & HITS@3  & 3.009$\pm$0.100  & 1.820$\pm$0.146  & 1.010$\pm$0.041  & 5.859$\pm$0.000          & 0.157$\pm$0.000   & \uline{27.940}$\pm$0.200  & \begin{tabular}[c]{@{}c@{}}\textbf{33.460}$^*$$\pm$0.000\\(\textit{+19.757\%}) \end{tabular}  \\
                           & HITS@5  & 3.725$\pm$0.092  & 2.574$\pm$0.203  & 1.212$\pm$0.049  & 5.859$\pm$0.000          & 0.229$\pm$0.000   & \uline{33.92}$\pm$0.212   & \begin{tabular}[c]{@{}c@{}}\textbf{39.111}$^*$$\pm$0.723\\(\textit{+15.304\%})\end{tabular}  \\
                           & HITS@10 & 6.372$\pm$0.201  & 4.810$\pm$0.242  & 1.812$\pm$0.019  & 7.227$\pm$0.000          & 0.377$\pm$0.000   & \uline{40.73}$\pm$0.217   & \begin{tabular}[c]{@{}c@{}}\textbf{45.646}$^*$$\pm$0.115\\(\textit{+12.070\%})\end{tabular}  \\ 
\arrayrulecolor{black}\cline{1-1}\arrayrulecolor{black}\cline{2-9}
\multirow{4}{*}{HYPERTEXT} & MAE     & $\times$             & $\times$             & $\times$             & $\times$                     & 47.496$\pm$0.002  & \uline{10.117}$\pm$0.367  & \begin{tabular}[c]{@{}c@{}}\textbf{1.085}$^*$$\pm$0.010\\(\textit{+89.275\%})\end{tabular}   \\
                           & HITS@3  & 10.310$\pm$0.151 & 8.415$\pm$0.276  & 7.810$\pm$0.000  & 6.463$\pm$0.014          & 1.717$\pm$0.000   & \uline{12.331}$\pm$0.384  & \begin{tabular}[c]{@{}c@{}}\textbf{15.150}$^*$$\pm$0.096\\(\textit{+22.861\%})\end{tabular}  \\
                           & HITS@5  & 14.698$\pm$0.113 & 14.248$\pm$0.237 & 12.224$\pm$0.000 & 12.002$\pm$0.000         & 2.258$\pm$0.000   & \uline{18.446}$\pm$0.480  & \begin{tabular}[c]{@{}c@{}}\textbf{19.762}$^*$$\pm$0.079\\(\textit{+7.134\%})\end{tabular}   \\
                           & HITS@10 & 21.846$\pm$0.157 & 22.726$\pm$0.354 & 22.998$\pm$0.000 & 22.952$\pm$0.000         & 3.877$\pm$0.000   & \uline{28.460}$\pm$1.044  & \begin{tabular}[c]{@{}c@{}}\textbf{28.660}$\pm$0.105\\(\textit{+0.703\%})\end{tabular}   \\ 
\arrayrulecolor{black}\cline{1-1}\arrayrulecolor{black}\cline{2-9}
\multirow{4}{*}{RT-POL}    & MAE     & $\times$             & $\times$             & $\times$             & $\times$                     & Inf            & \uline{50.096}$\pm$1.228  & \begin{tabular}[c]{@{}c@{}}\textbf{39.485}$^*$$\pm$0.030\\(\textit{+21.181\%})\end{tabular}  \\
                           & HITS@3  & 0.000$\pm$0.000  & 0.000$\pm$0.000  & 1.576$\pm$0.021  & 2.031$\pm$0.000          & 0.024$\pm$0.000   & \uline{2.564}$\pm$0.212   & \begin{tabular}[c]{@{}c@{}}\textbf{3.120}$^*$$\pm$0.021\\(\textit{+21.685\%})\end{tabular}   \\
                           & HITS@5  & 0.000$\pm$0.000  & 0.000$\pm$0.000  & 2.169$\pm$0.007  & 2.187$\pm$0.000          & 0.043$\pm$0.000   & \uline{3.945}$\pm$0.030   & \begin{tabular}[c]{@{}c@{}}\textbf{4.381}$^*$$\pm$0.001\\(\textit{+11.052\%})\end{tabular}   \\
                           & HITS@10 & 0.077$\pm$0.000  & 0.078$\pm$0.000  & 3.162$\pm$0.019  & 3.594$\pm$0.000          & 0.076$\pm$0.000   & \uline{5.624}$\pm$0.001   & \begin{tabular}[c]{@{}c@{}}\textbf{6.653}$^*$$\pm$0.003\\(\textit{+18.297\%})\end{tabular}   \\
\arrayrulecolor{black}\cline{1-1}\arrayrulecolor{black}\cline{2-9}
\end{tabular}
\arrayrulecolor{black}
\end{table*}

Table.~\ref{tab:overview} shows the performance comparison of \MODELNAME~ with state-of-the-art baselines. All methods are performed in 5 independent experiments on all datasets.


\textbf{Future link prediction task.} Compared with discrete temporal GNNs (DySAT and EvolveGCN), static GNNs (GCN and GAT) often obtain better link prediction performance on datasets with a lower percentage of inductive events (ENRON and HYPERTEXT). However, on datasets with high inductive event percentages, both discrete and continuous temporal GNNs (except KnowEvolve) show an obvious advantage. For example, RT-POL, a dataset with $83.1\%$ of inductive events in the test set, where GCN and GAT almost cannot work, but temporal GNNs still maintains some performance. This is due to the fact that GCN and GAT have remembered the structure of the cumulative graph, so they tend to predict the events that have been observed in history. Discrete temporal GNNs do capture some temporal information. However, fine-grained temporal information is lost during snapshots generation, which leads to underperforming continuous GNNs. KnowEvolve is a continuous GNN based on temporal point process, but the method simply treats the temporal graph as a sequence of events without considering evolving graph structure. Thus it does not work well in future link prediction task. Among all baselines, EvoKG achieves the best performance on four datasets, as it models event time and evolving graph structure jointly. Results show \MODELNAME~signigicantly outperforms all baselines, with up to $112.355\%$ more HITS@3 on LSED than the second-best method.

\textbf{Event time prediction task.} The MAE metric for static GNNs and discrete GNNs are marked with $\times$ in Table.~\ref{tab:overview} as they cannot estimate event time. The MAE metric for KnowEvolve on RT-POL is marked with INF as there is a numerical overflow. \MODELNAME~significantly outperforms the other two continuous GNNs, with up to $89.275\%$ less MAE on HYPERTEXT than the second best method.

\subsubsection{Detailed comparison to EvoKG}\label{sec:exp_evokg}
In this section, we illustrate the advantage of using uniform embeddings to learn the structural dynamics and temporal characteristics of nodes by comparing EvoKG in detail, which learns the structural embeddings and temporal embeddings separately.

\begin{figure*}[tbhp]
	\centering
	\includegraphics[width=0.195\linewidth]{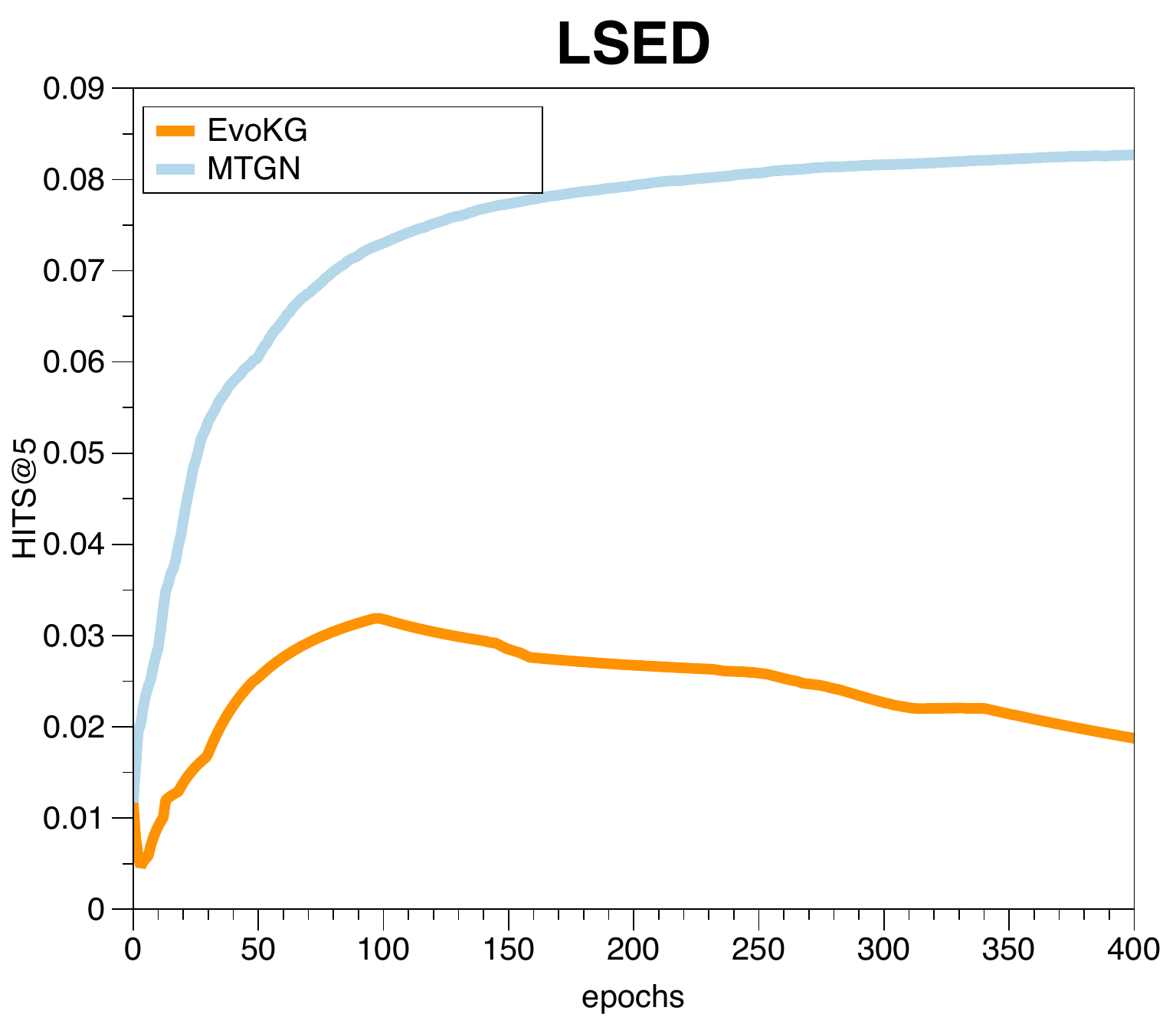}
	\includegraphics[width=0.195\linewidth]{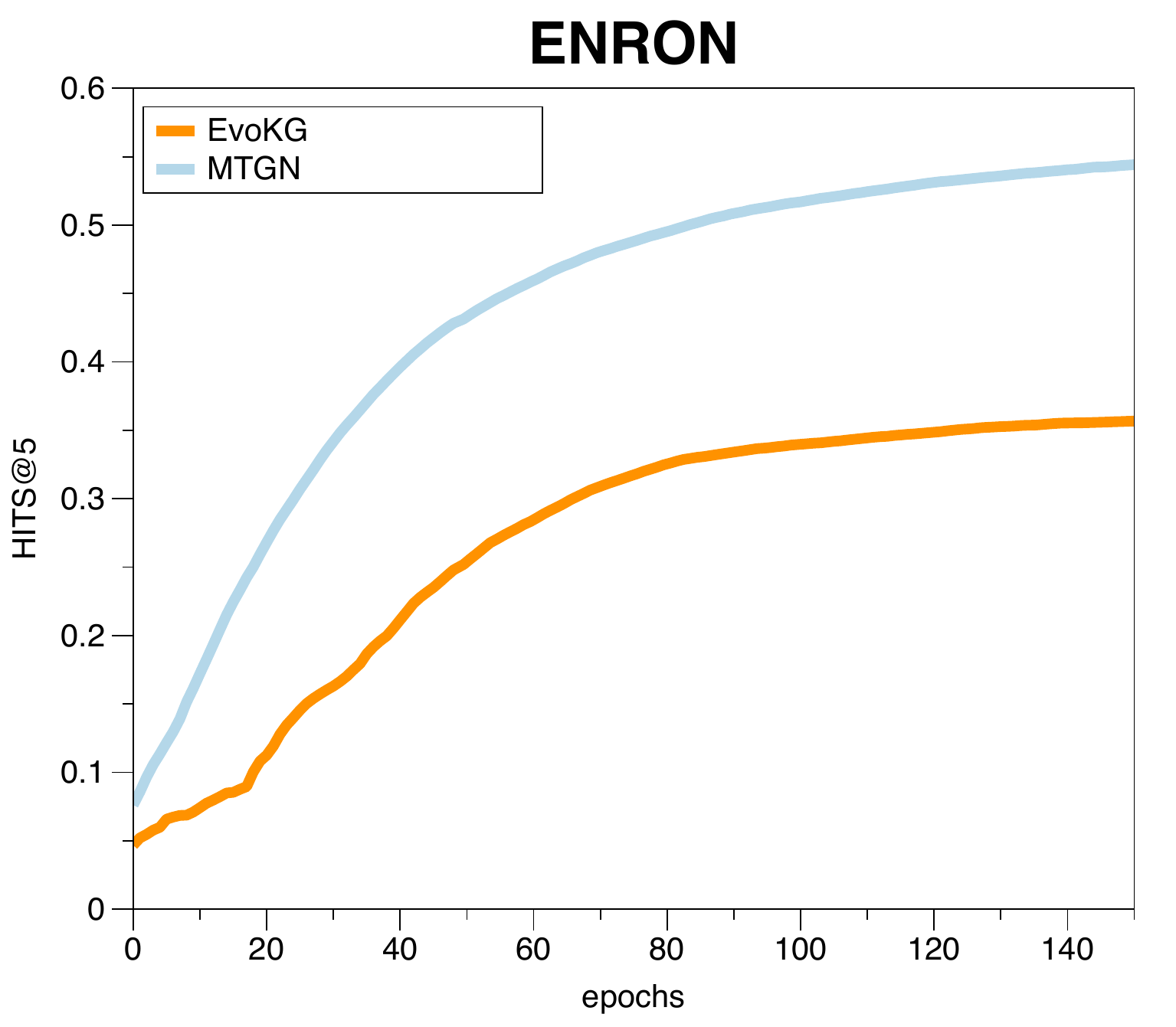}
	\includegraphics[width=0.195\linewidth]{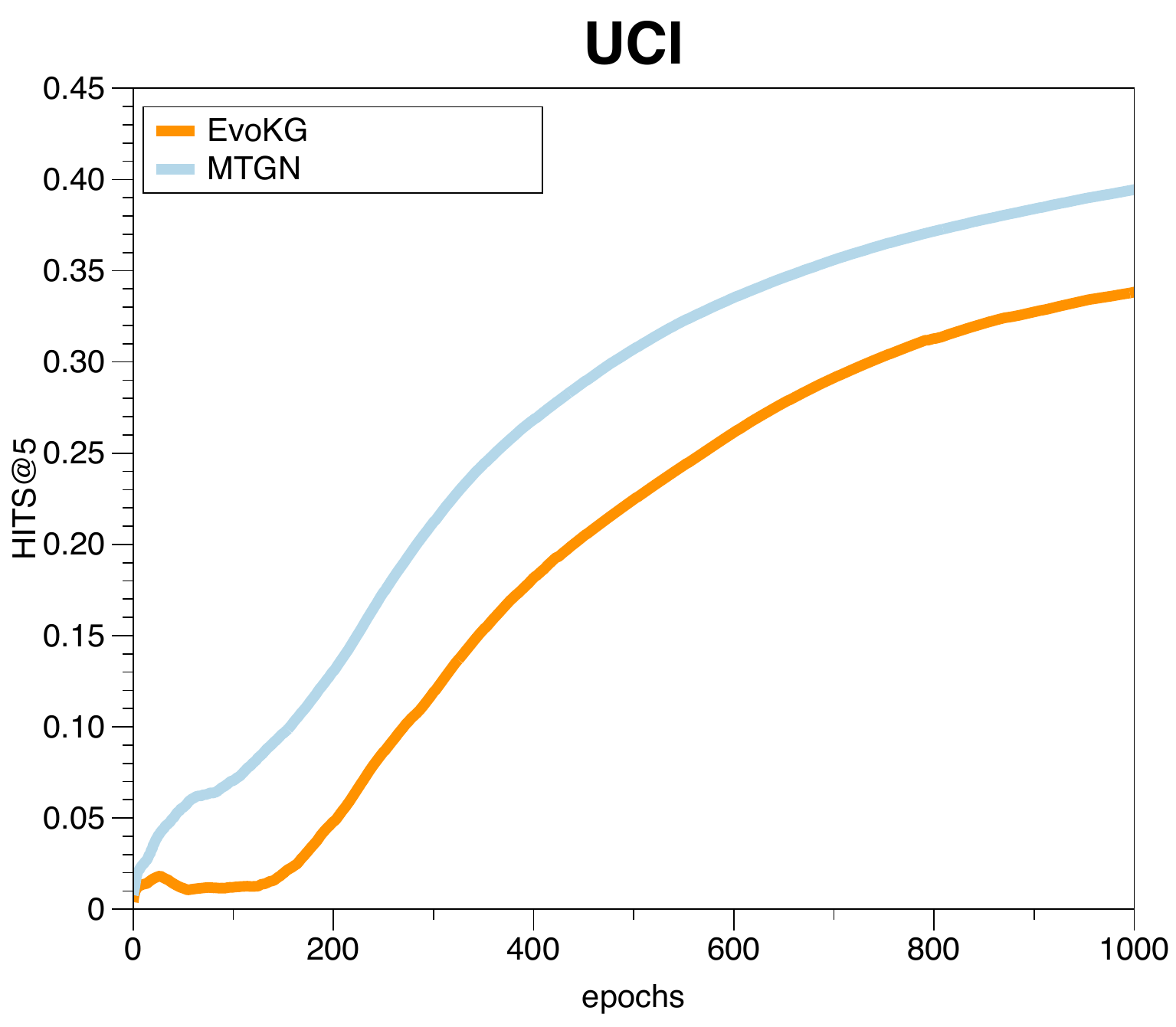}
	\includegraphics[width=0.195\linewidth]{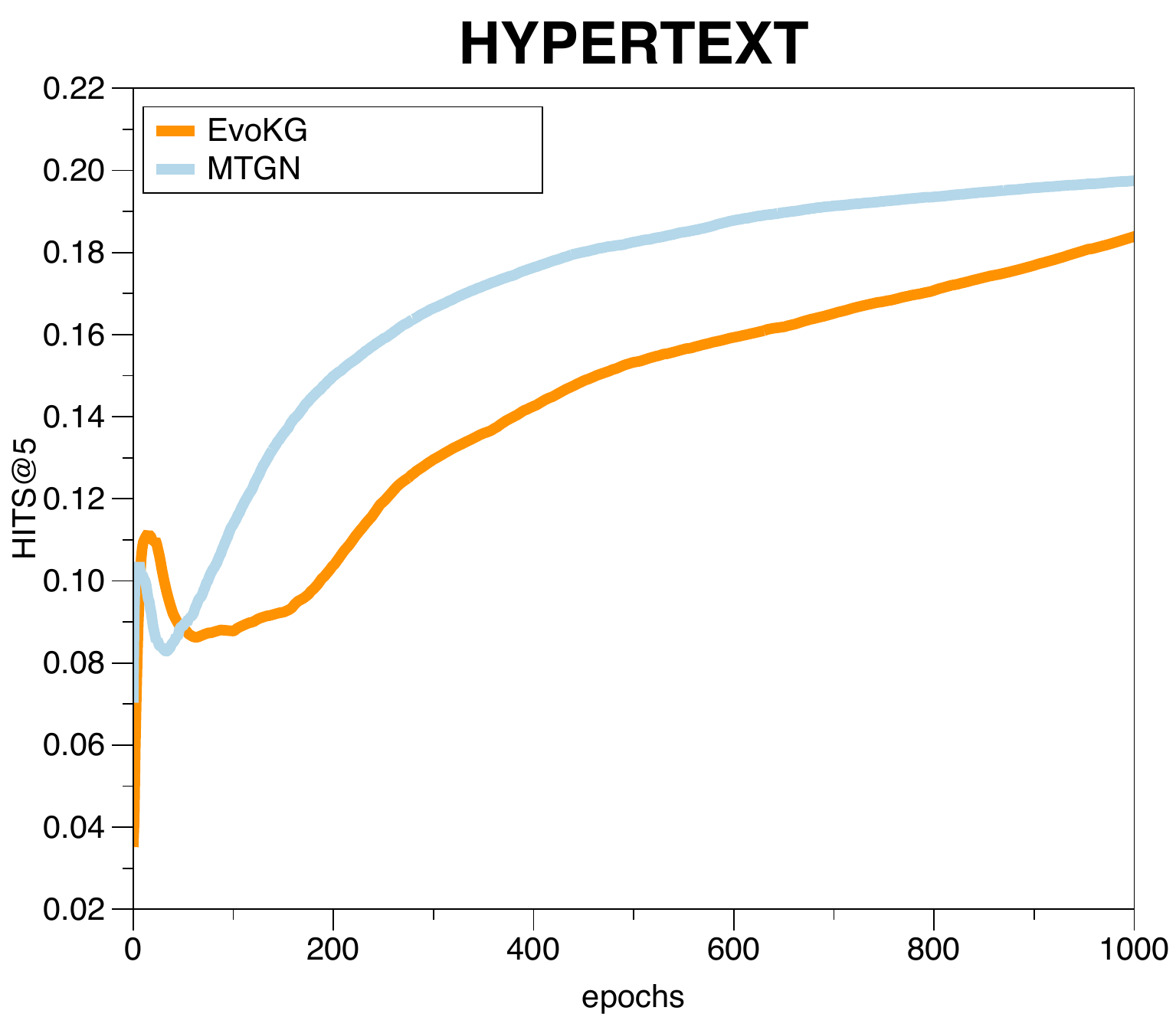}
	\includegraphics[width=0.195\linewidth]{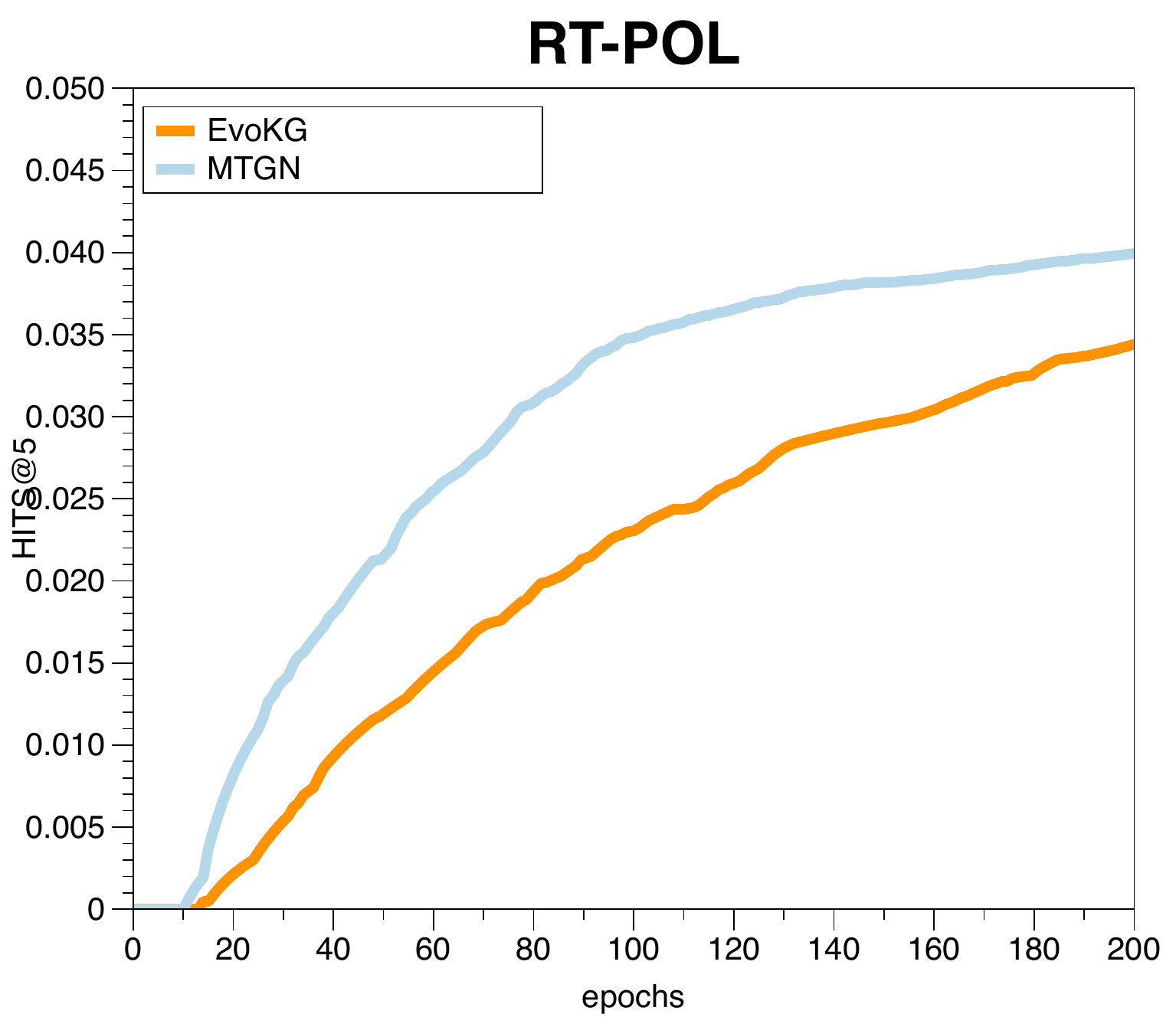}
	\includegraphics[width=0.195\linewidth]{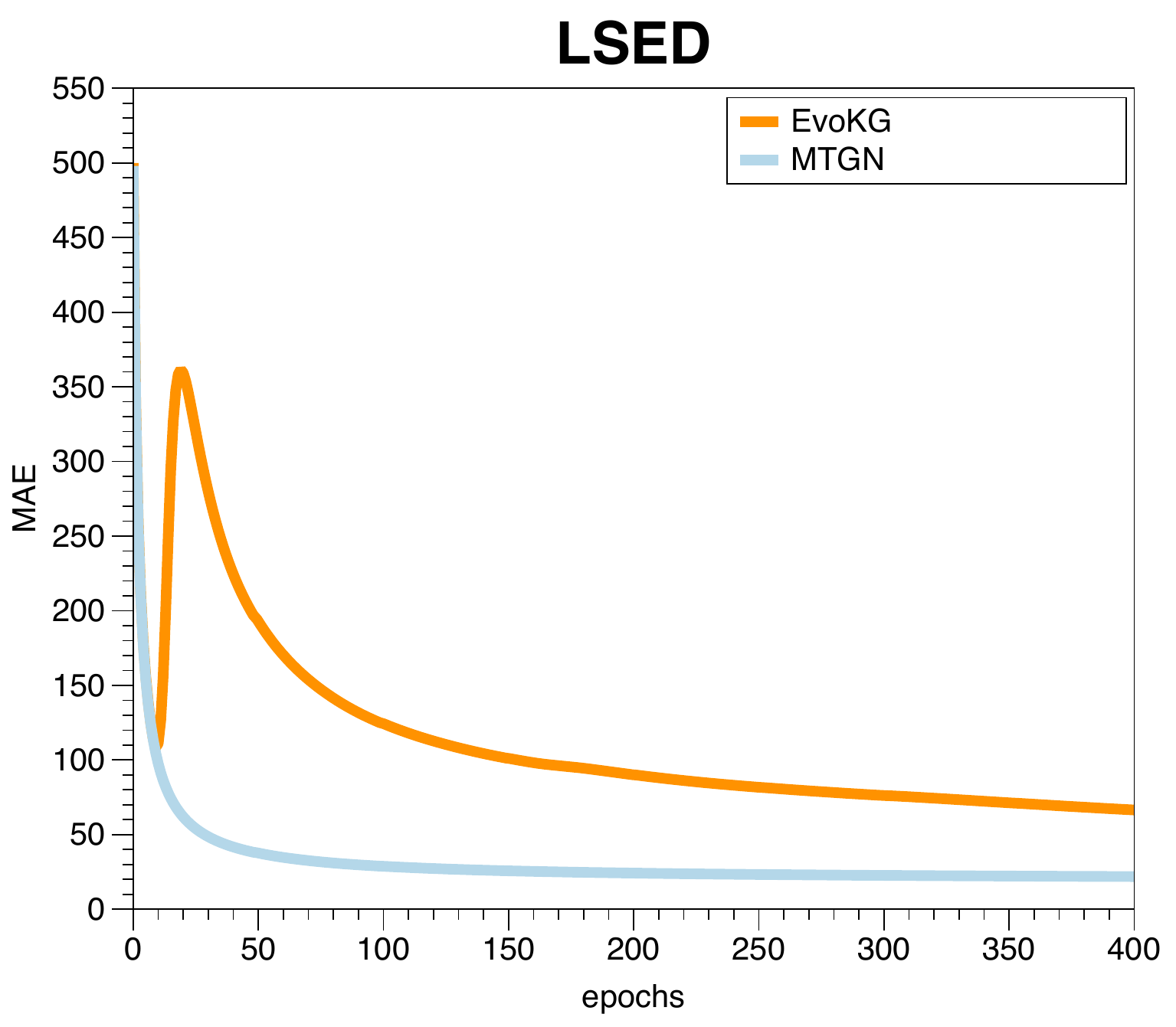}
	\includegraphics[width=0.195\linewidth]{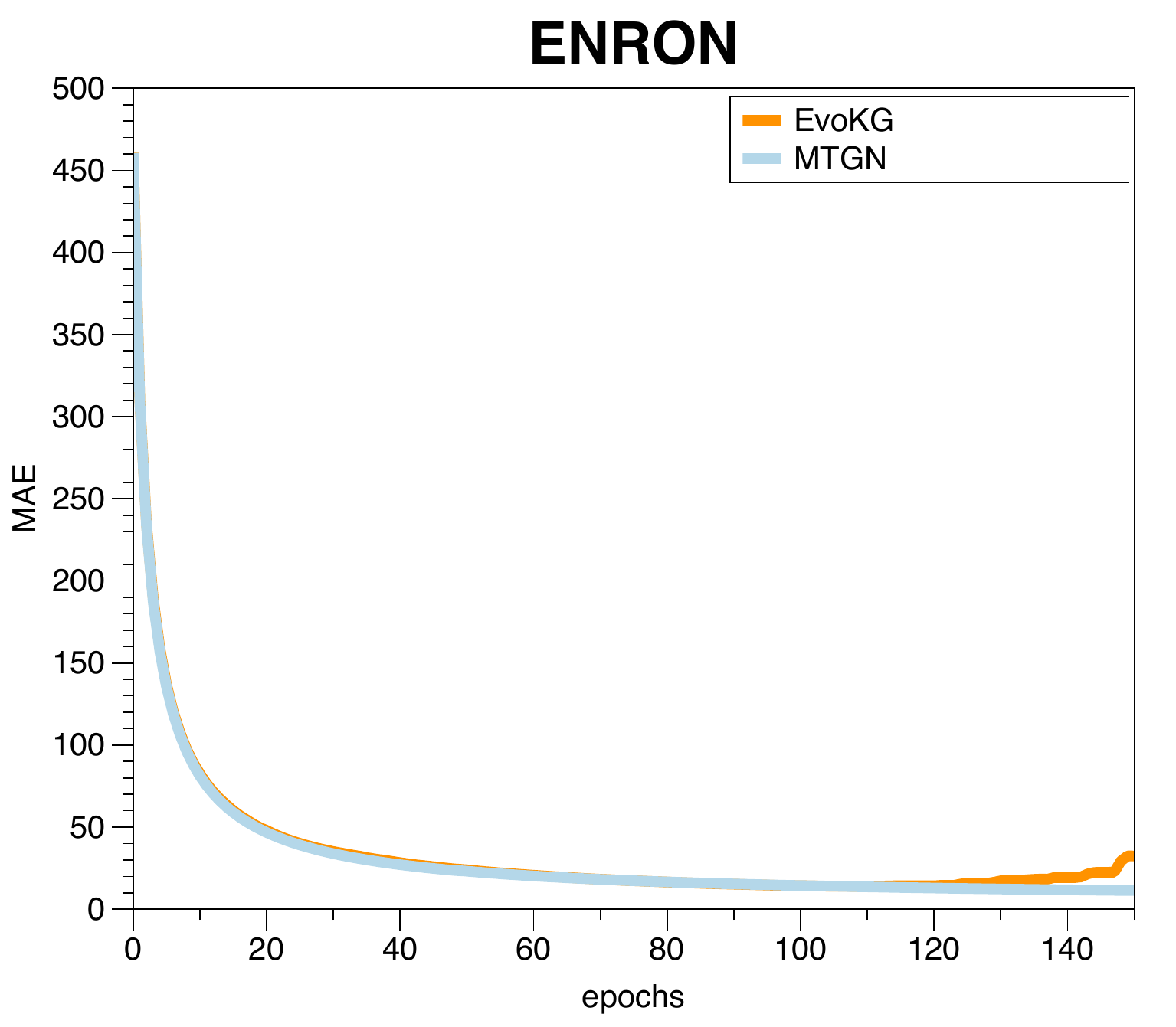}
	\includegraphics[width=0.195\linewidth]{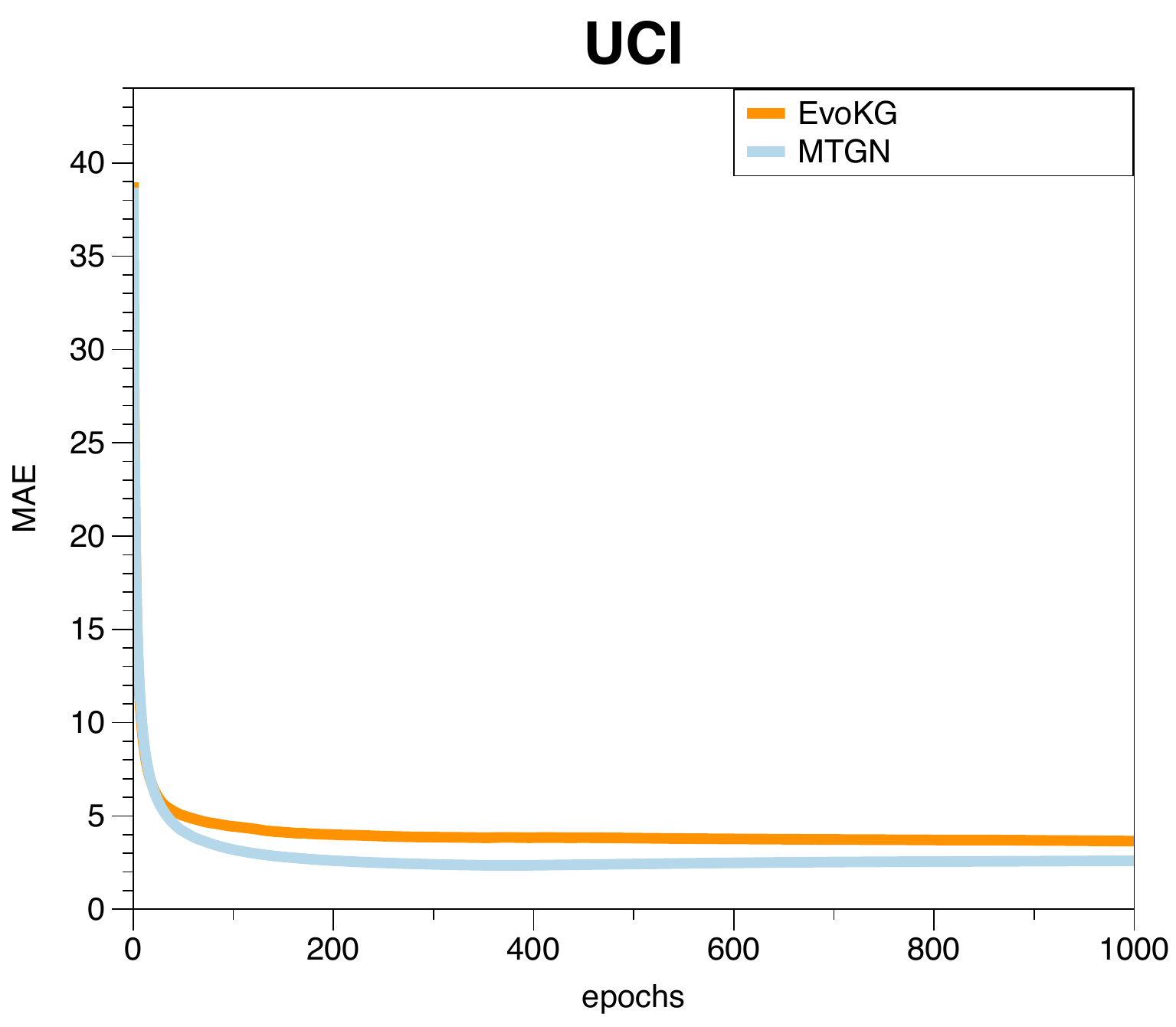}
	\includegraphics[width=0.195\linewidth]{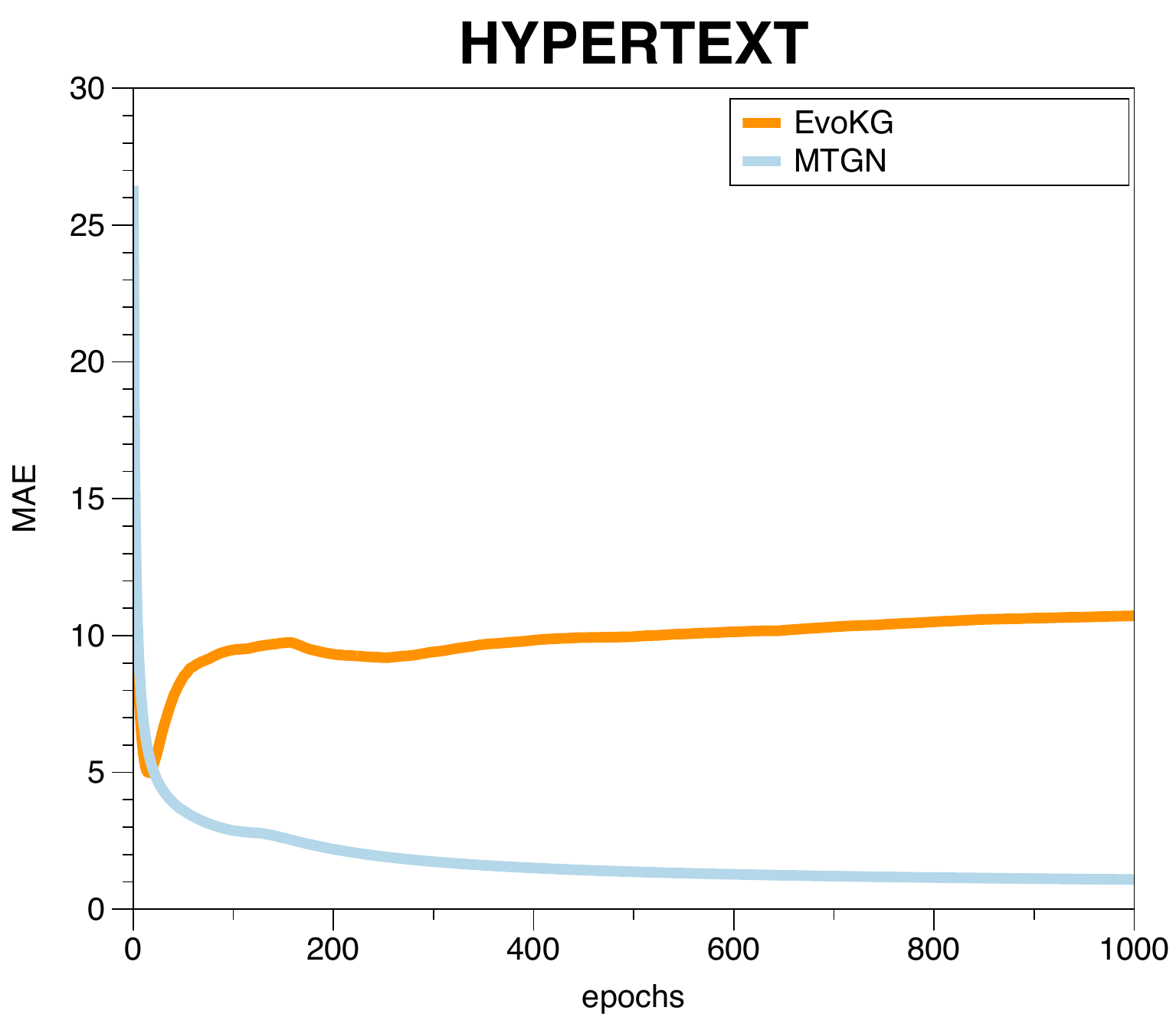}
	\includegraphics[width=0.195\linewidth]{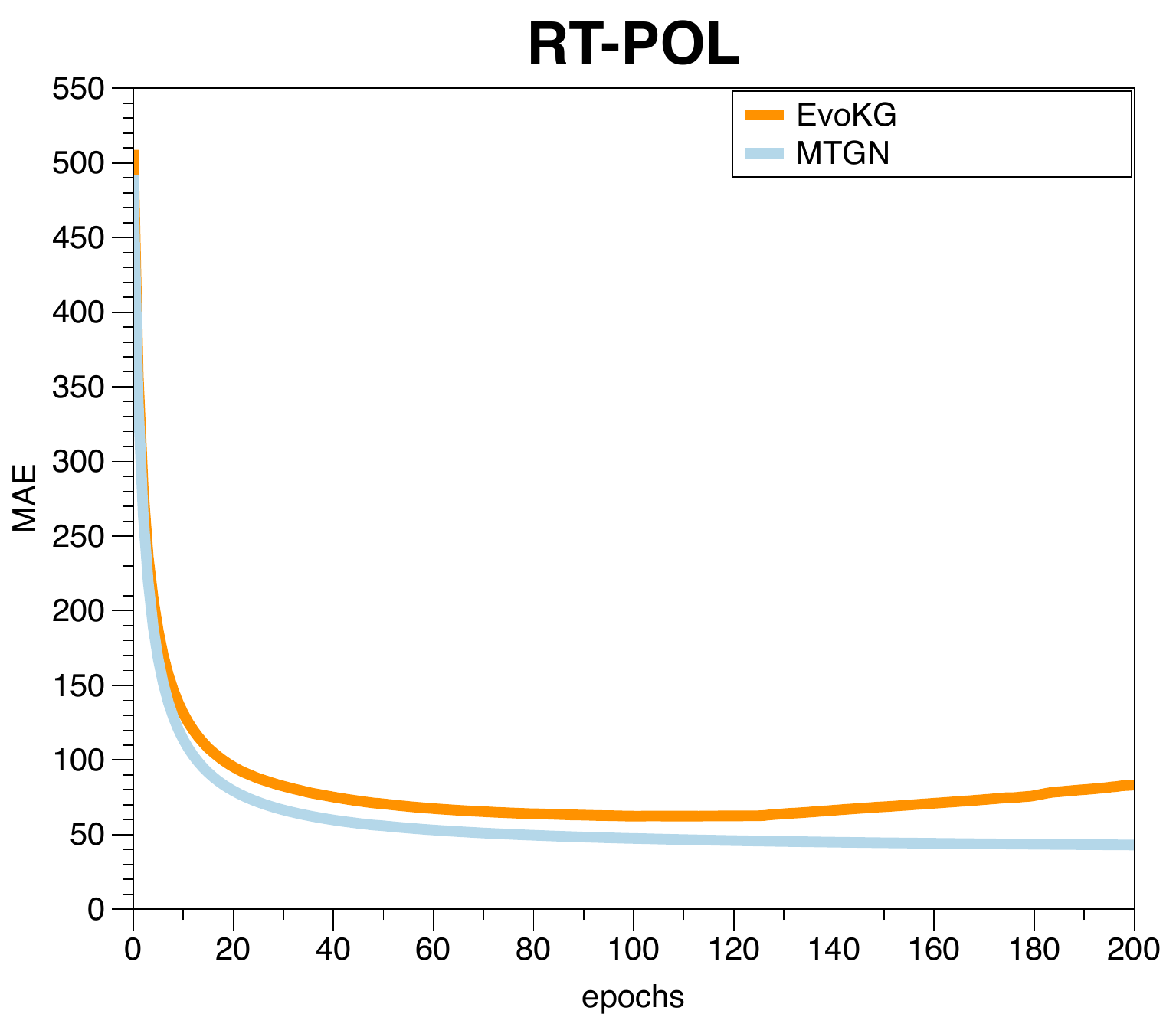}
	\caption{Changes in HITS@5 and MAE performance of EvoKG and \MODELNAME~on five datasets with the number of training epochs increases}\label{fig:compare_EvoKG}
\end{figure*}
Fig.~\ref{fig:compare_EvoKG} shows the changes in HITS@5 and MAE performance of EvoKG and \MODELNAME~on all datasets with the number of training epochs increasing. It can be seen from the results that EvoKG does not optimize the link prediction performance and event time prediction performance simultaneously due to learning structural embedding and temporal embeddings separately. On LSED, the link prediction performance is overfitted before the event time prediction performance is converged. However, on other datasets (except UCI), event time prediction performance is overfitted before the link prediction performance is converged.

\MODELNAME~address the above issues by learning structural dynamic and temporal characteristics of nodes in uniform embeddings. As shown in Fig.~\ref{fig:compare_EvoKG}, \MODELNAME~can obtain the best performance of link prediction and event time prediction at the same training epoch. And event time prediction in \MODELNAME~less likely to be overfitted.

\begin{figure}[h]
	\centering
	
	\subfigure[EvoKG-HYPERTEXT-structural]{
		\includegraphics[width=0.465\linewidth]{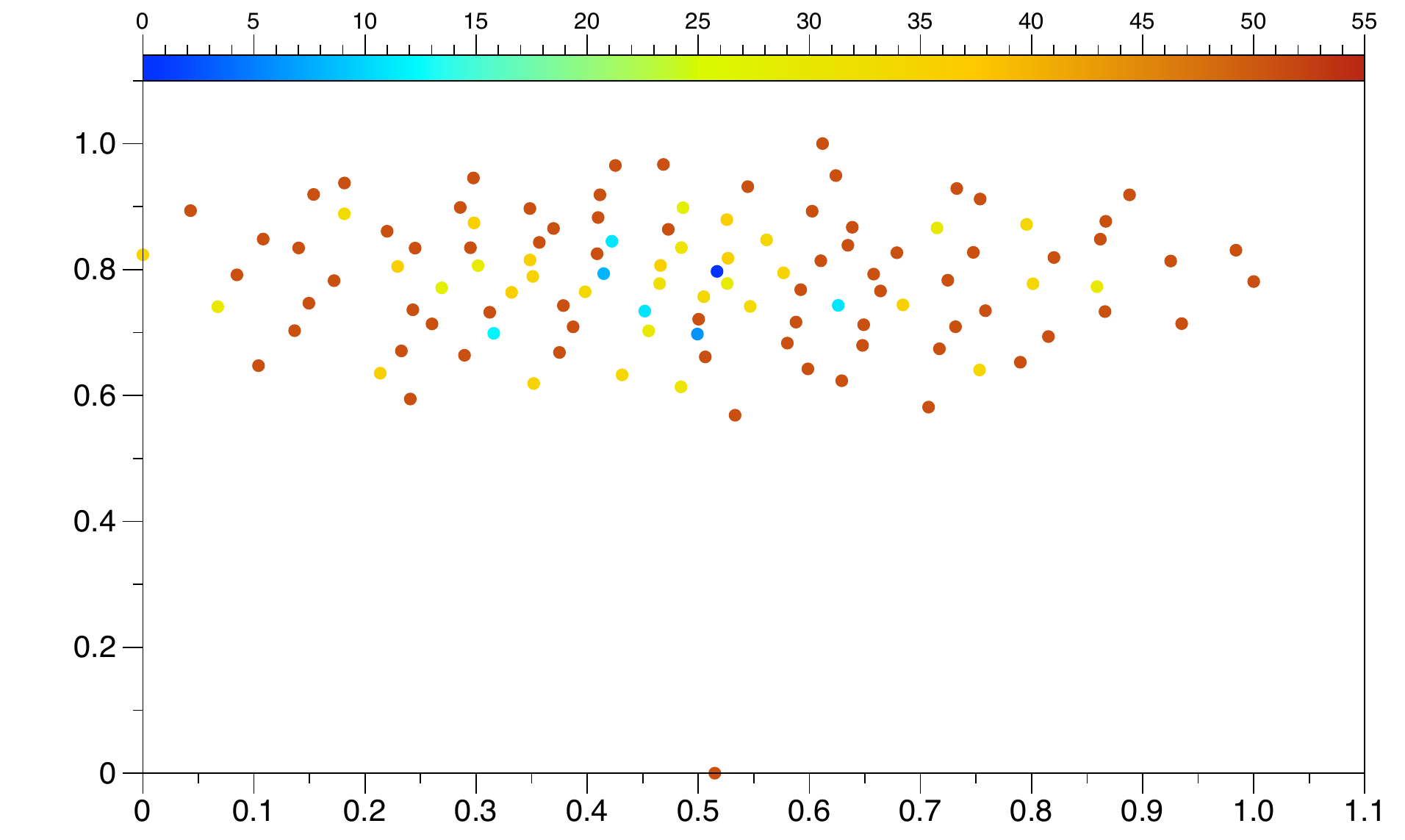}
	}
	\subfigure[EvoKG-HYPERTEXT-temporal] {
		\includegraphics[width=0.465\linewidth]{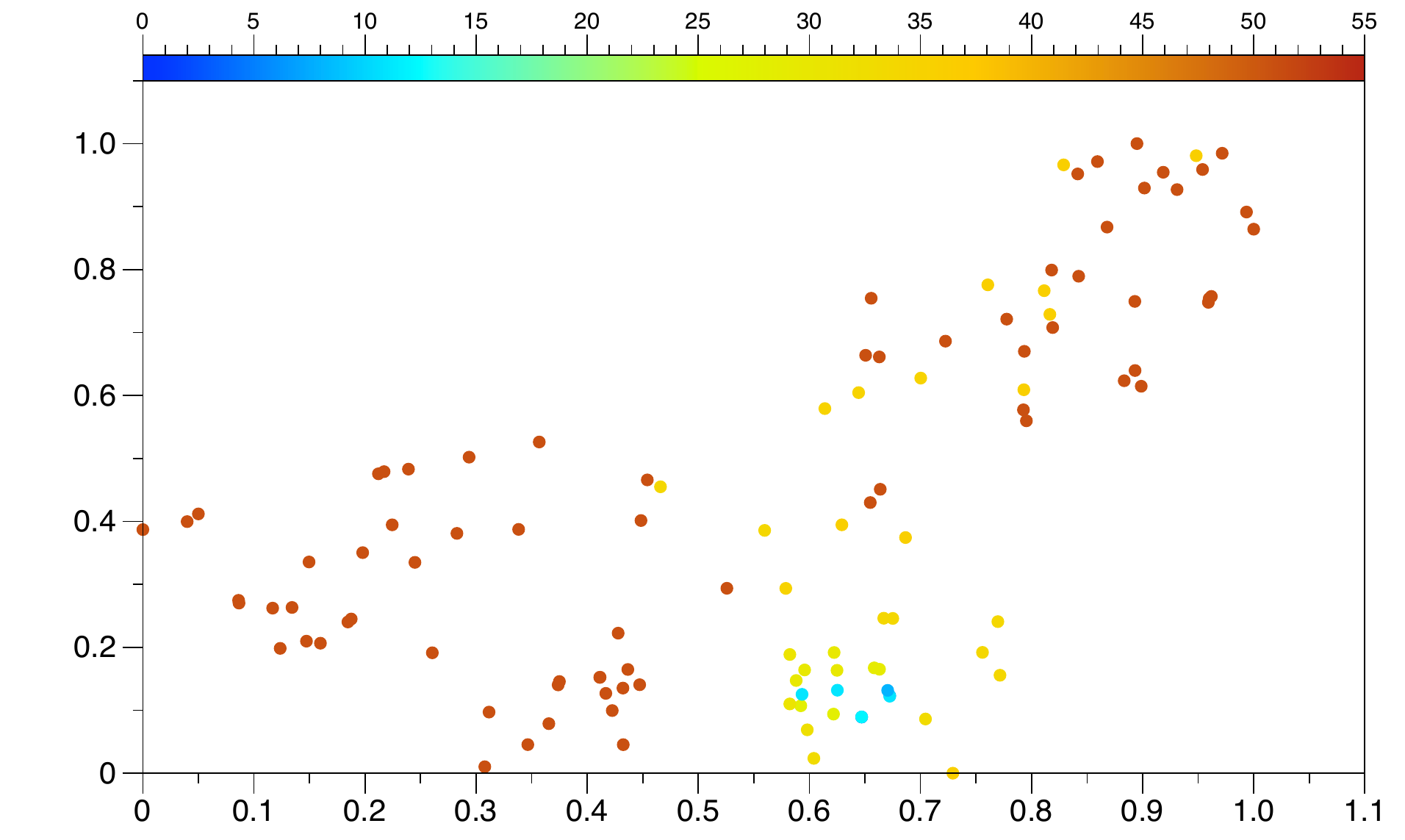}
	}
	\subfigure[MTGN-HYPERTEXT-structural]{
		\includegraphics[width=0.465\linewidth]{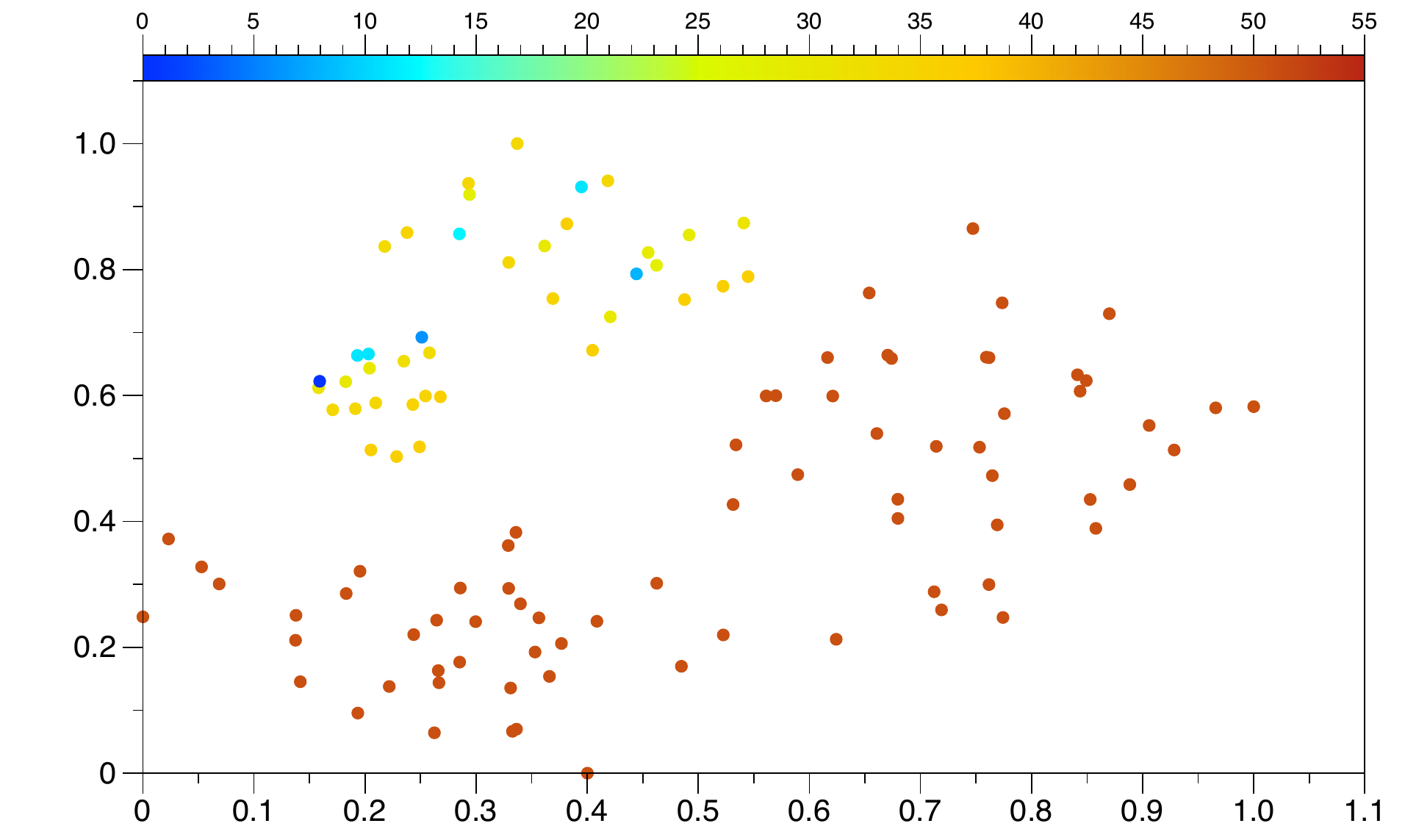}
	}
	\subfigure[MTGN--HYPERTEXT-temporal]{
		\includegraphics[width=0.465\linewidth]{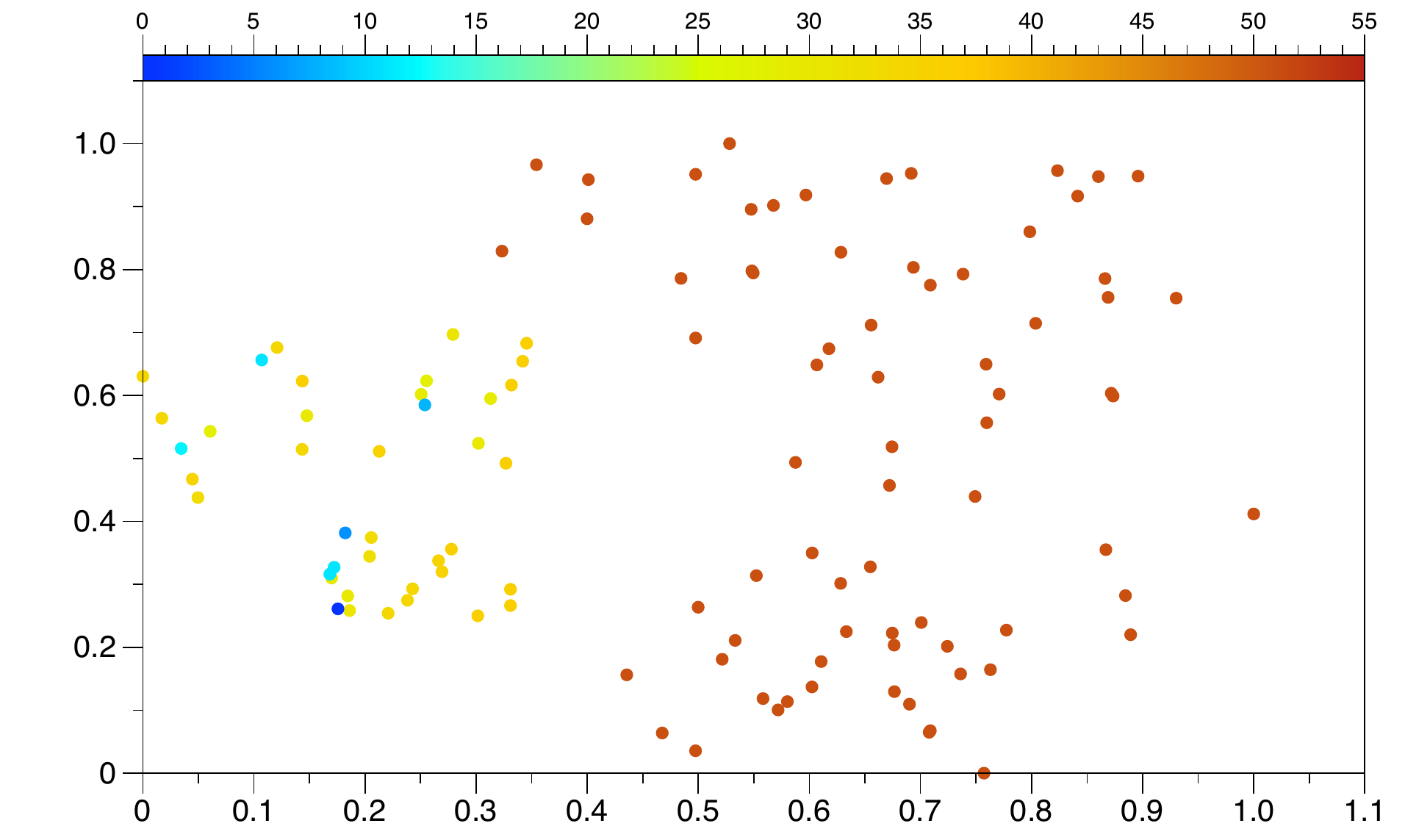}
	}
	\subfigure[EvoKG-LSED-structural]{
		\includegraphics[width=0.465\linewidth]{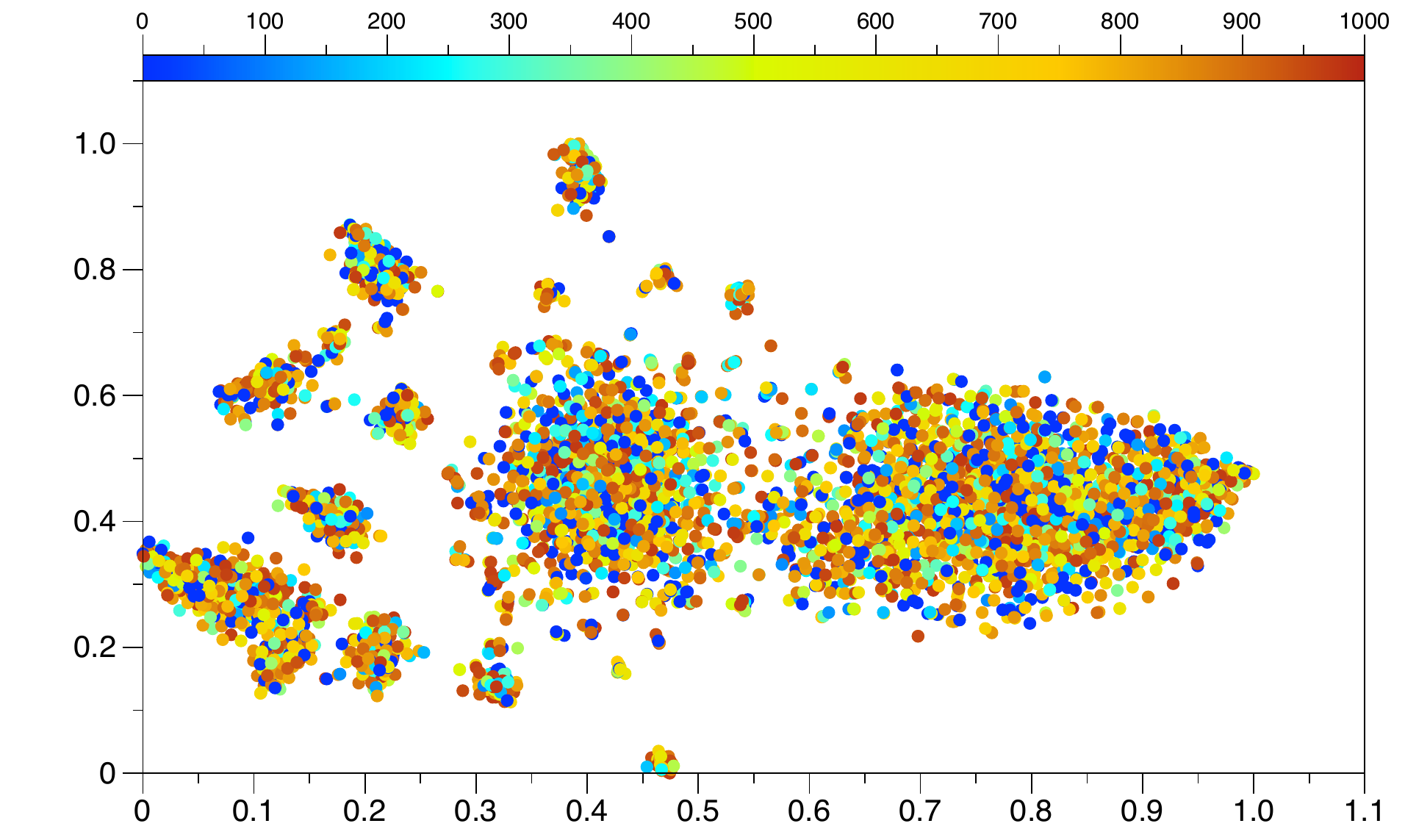}
	}
	\subfigure[EvoKG-LSED-temporal] {
		\includegraphics[width=0.465\linewidth]{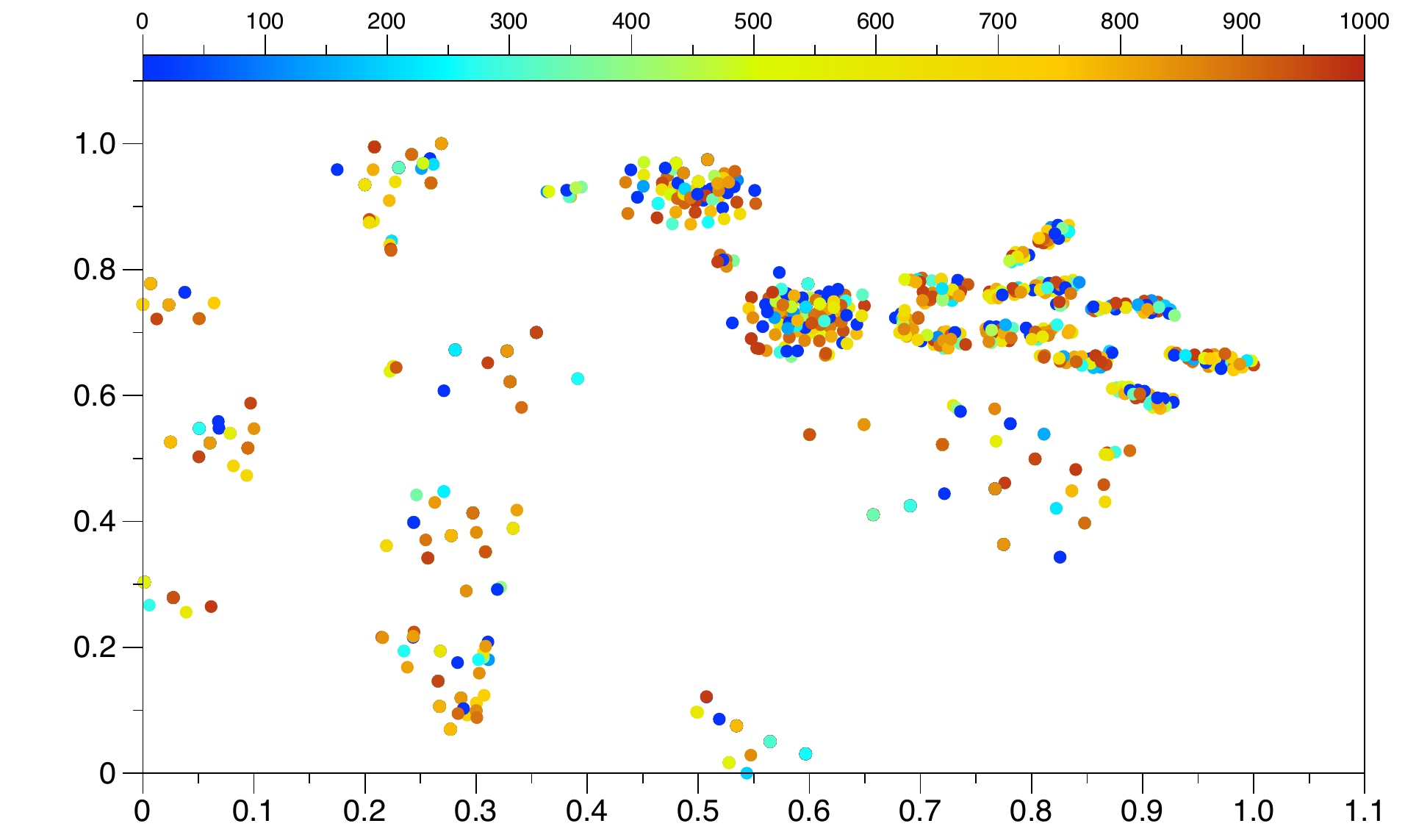}
	}
	\subfigure[MTGN-LSED-structural]{
		\includegraphics[width=0.465\linewidth]{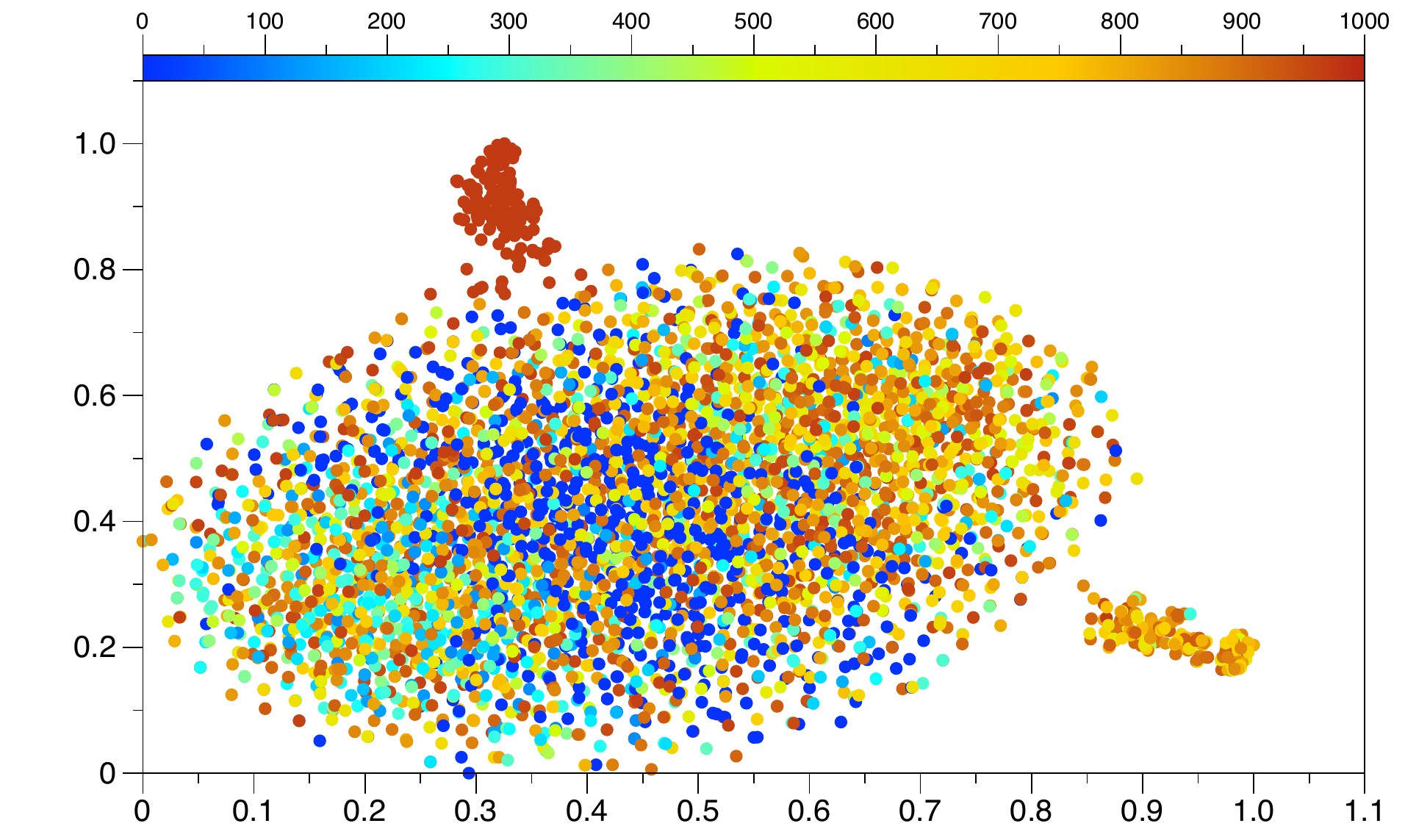}
	}
	\subfigure[MTGN-LSED-temporal]{
		\includegraphics[width=0.465\linewidth]{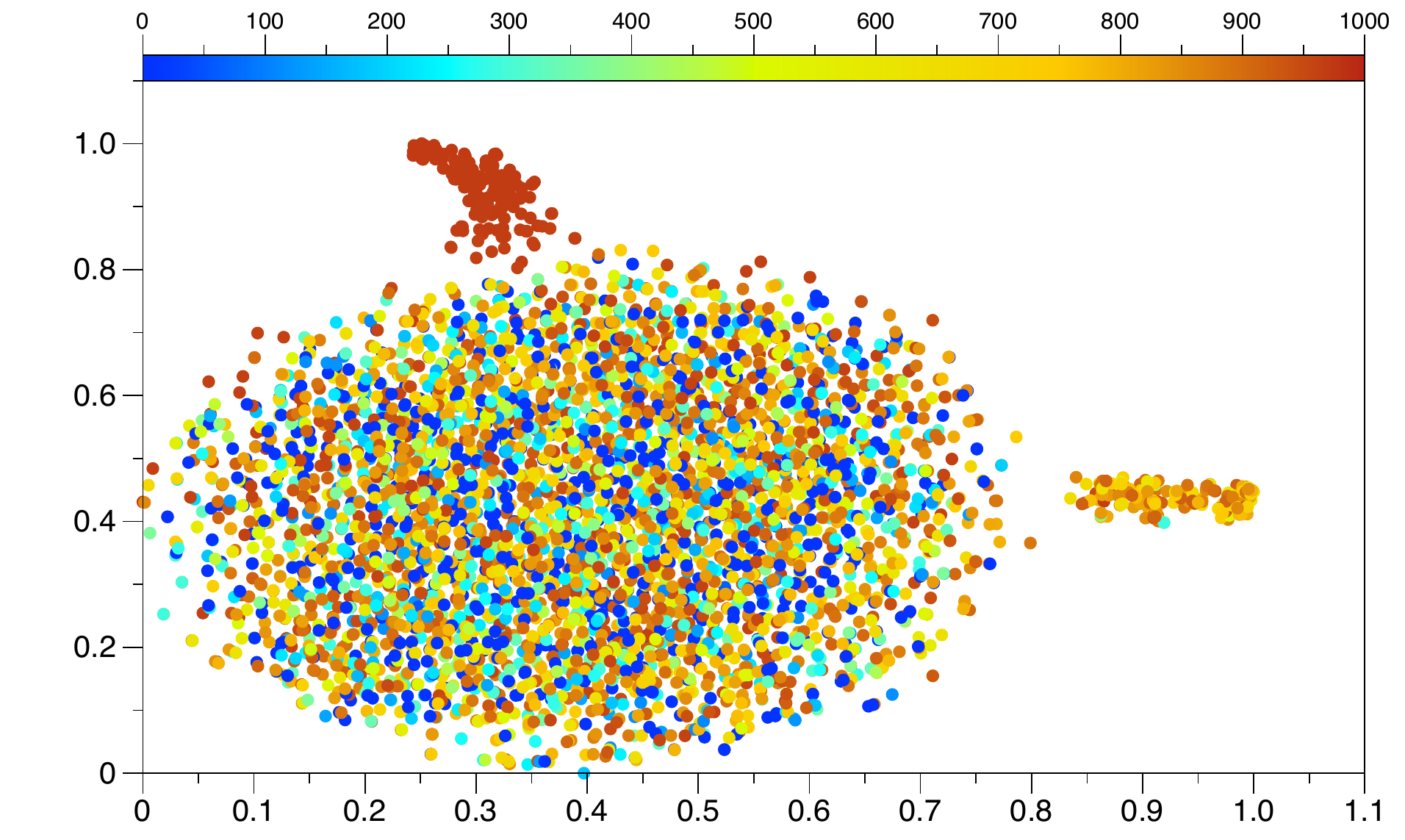}
	}
	\caption{tSNE visualization of node embeddings on HYPERTEXT and LSED. We use the color map to indicate the last time of a node is involved in an observed event.}\label{fig:embeddings_vis}
\end{figure}
To further explain the difference between EvoKG and \MODELNAME, we show the learned graph structural embeddings (used for link prediction task, specially Eq.\eqref{eq:g_s} in \MODELNAME) and temporal characteristics embeddings (used for event time prediction, specially Eq.\eqref{eq:g_star_t} in \MODELNAME) in Fig.~\ref{fig:embeddings_vis} by using tSNE~\cite{tSNE} to project node embeddings from high dimension into a 2D space. From the figure, we can find the following differences:
\begin{enumerate}
	\item The distributions of structural and temporal embeddings separately learned by EvoKG are significantly different. While \MODELNAME~uses uniform embedding embeddings to learn structural and temporal embeddings (the only difference is whether concatenate static embeddings.), so the embeddings used for link prediction and event time prediction are similarly distributed overall.
	\item EvoKG's structure embeddings do not distinguish well between nodes that are last observed at different times as they do not model temporal information. In contrast, both the structural and temporal embeddings learned by \MODELNAME~can clearly distinguish nodes that were recently involved in the observed event from other nodes that were not involved in the event for a long time.
\end{enumerate}

\subsection{Ablation Study}
\begin{figure*}
	\centering
	\includegraphics[width=0.325\linewidth]{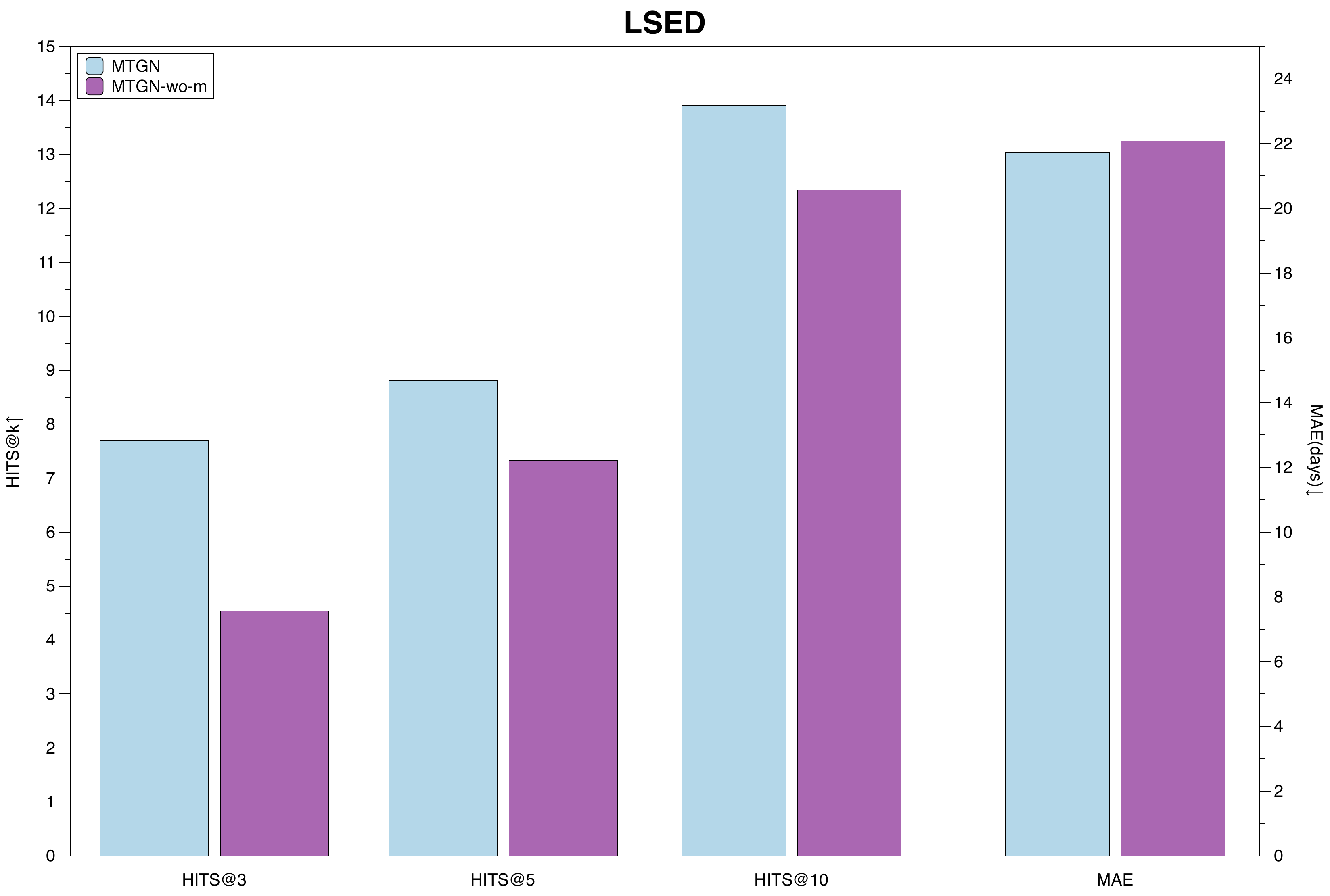}
	\includegraphics[width=0.325\linewidth]{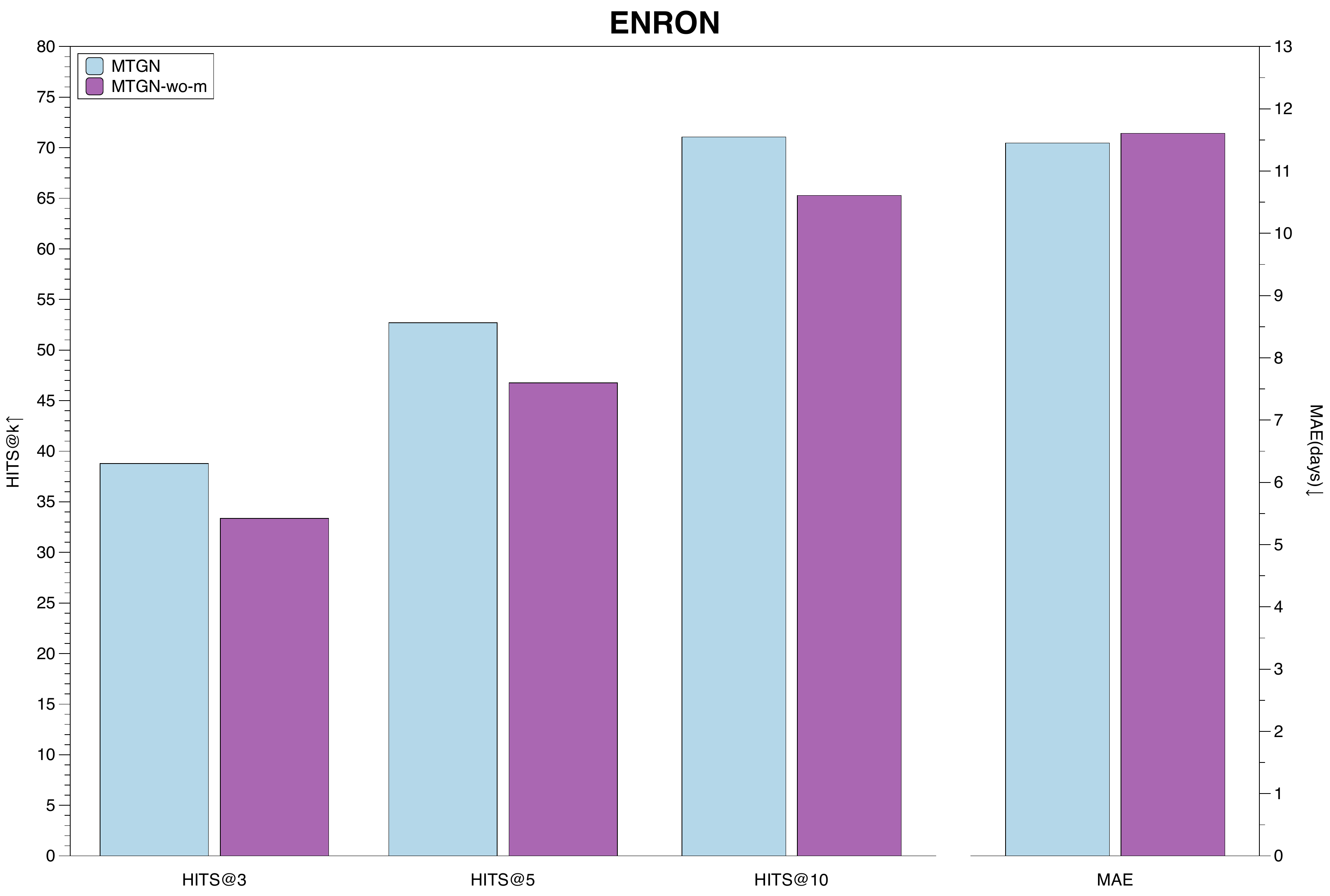}
	\includegraphics[width=0.325\linewidth]{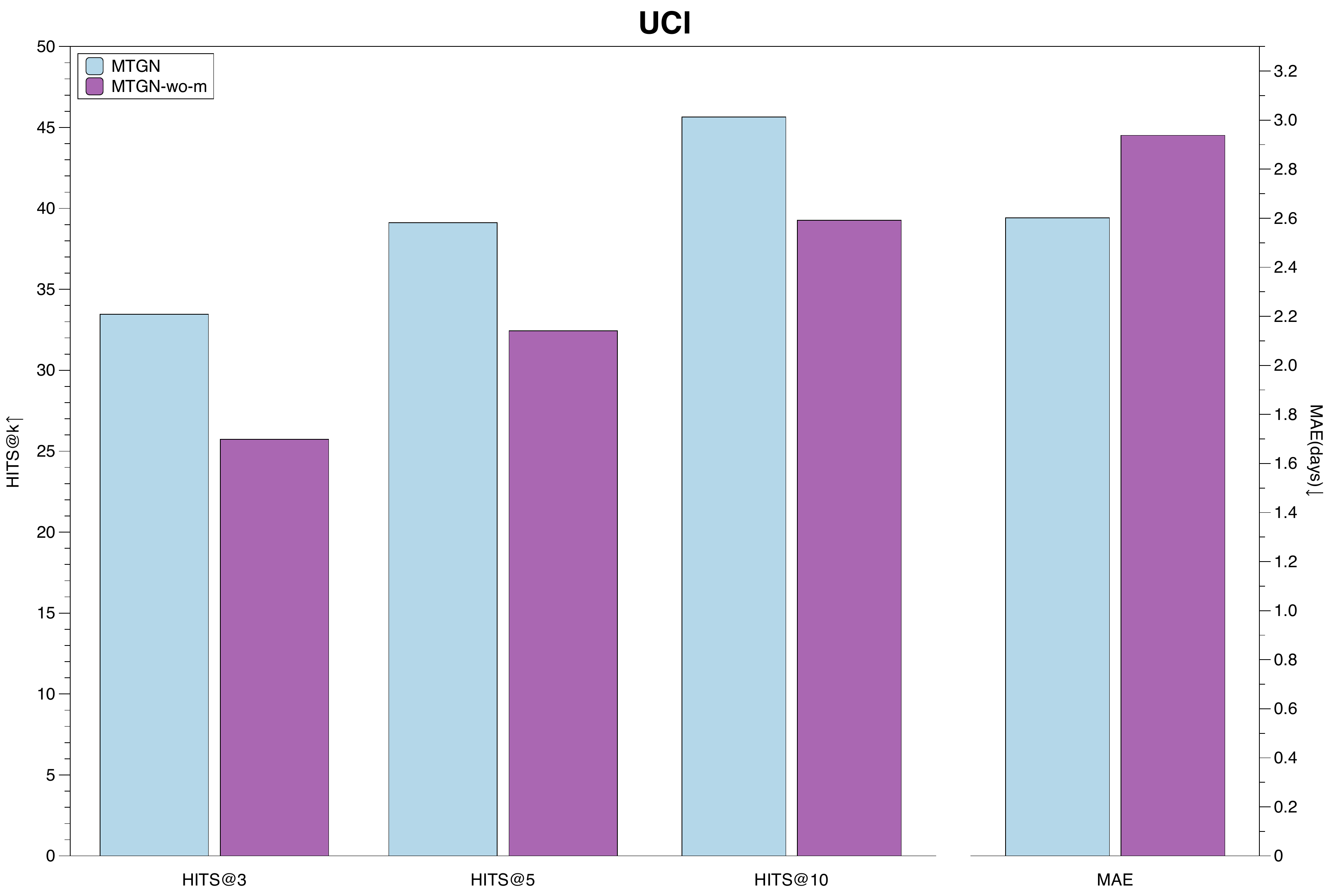}
	\includegraphics[width=0.325\linewidth]{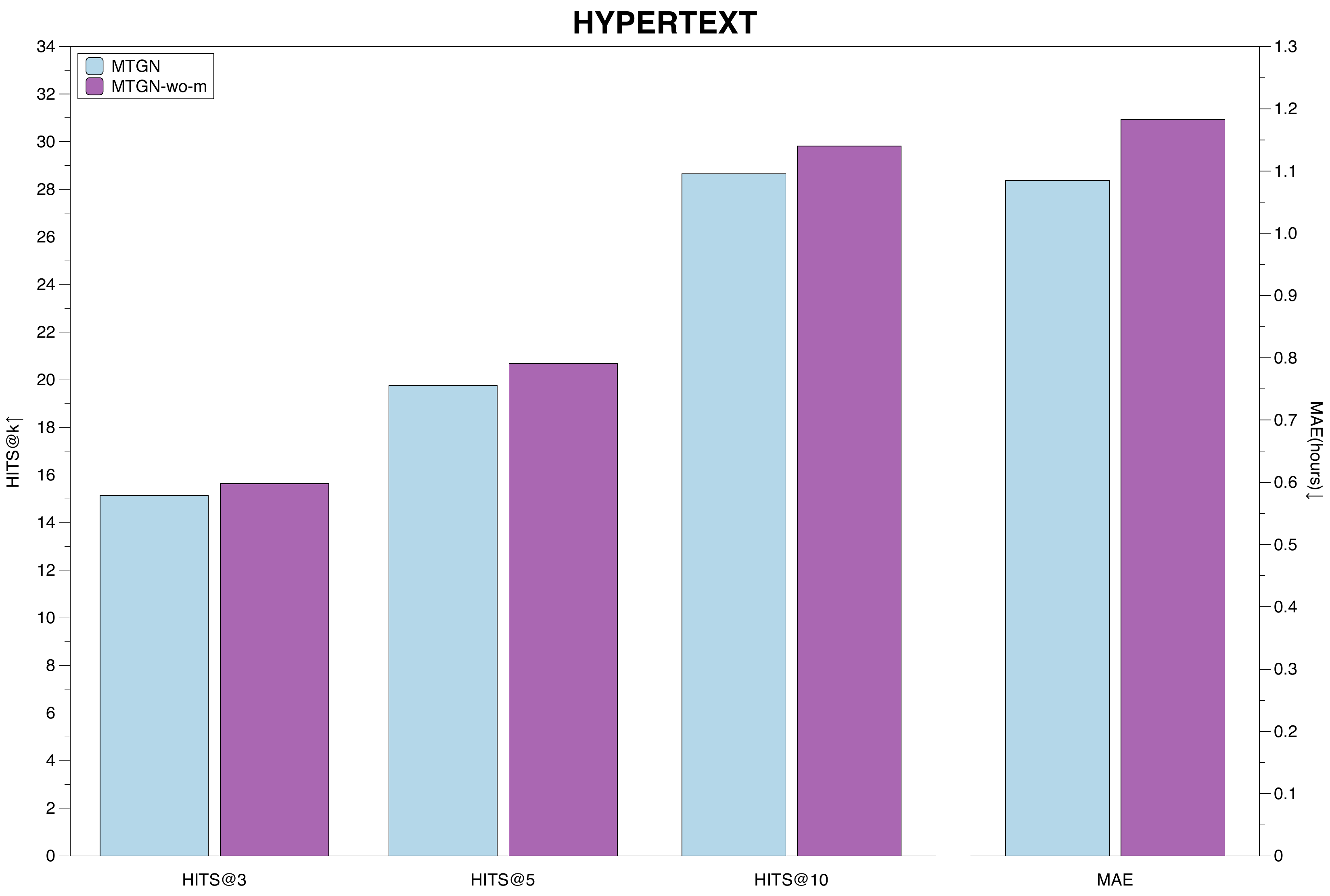}
	\includegraphics[width=0.325\linewidth]{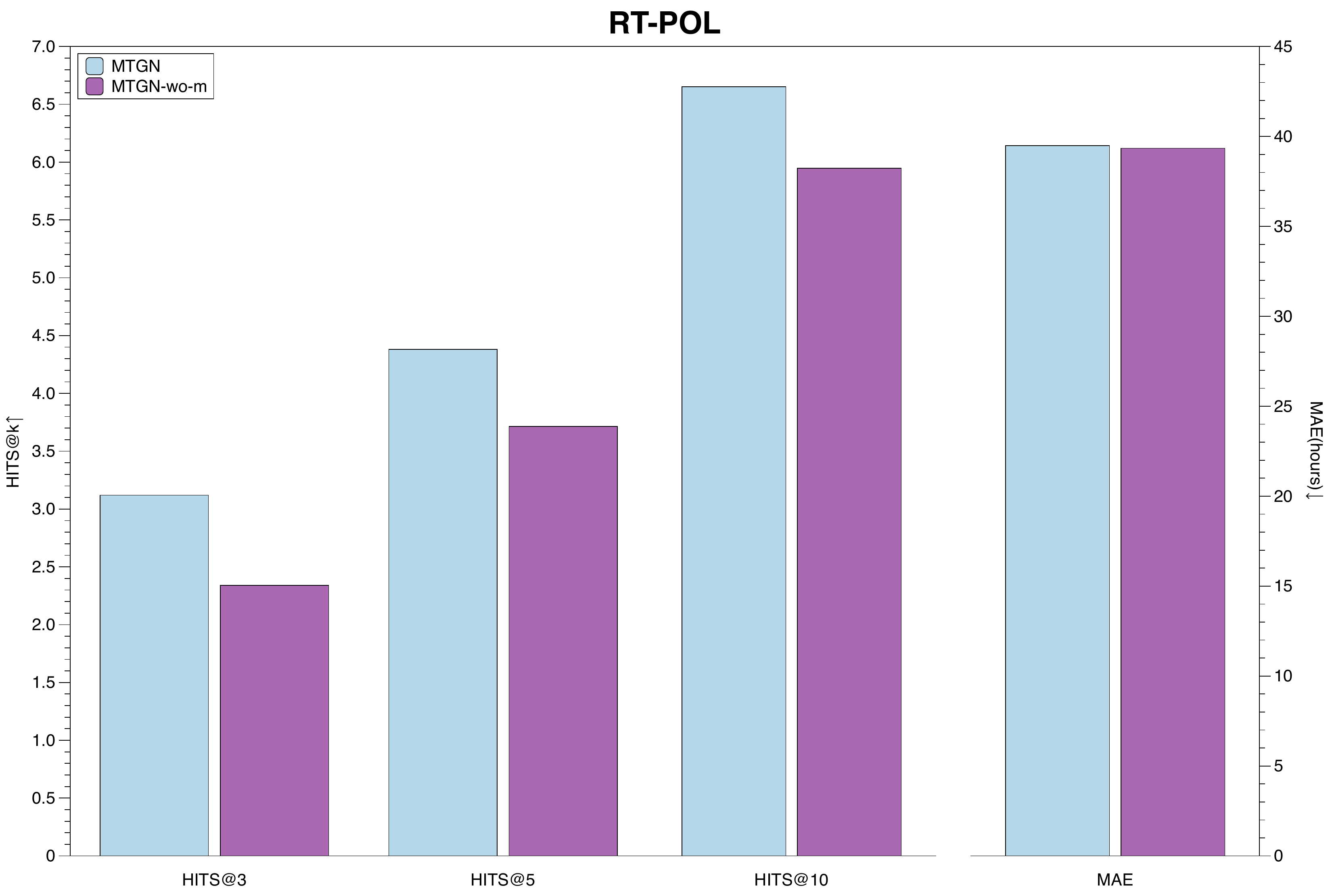}
	\caption{Performance comparison for ablation study.}\label{fig:no_missing}
\end{figure*}
By detailed comparison to EvoKG in previous section, we verify the superiority of using uniform embeddings to learn nodes' structural information and temporal characteristics. In this section, we demonstrate the superiority of modeling missing events by conducting an ablation study. We perform the following test:
\begin{itemize}
	\item \textbf{\MODELNAME-wo-m}. In this variant, we do not generate missing events (i.e. make $\mathcal{G}=\mathcal{O}$).
\end{itemize}

The results are shown in Fig.~\ref{fig:no_missing}. We can observe that \MODELNAME~consistently outperforms \MODELNAME-wo-m in all cases, except for the HITS@\{3,5,10\} metrics in the HYPERTEXT dataset, where \MODELNAME~is competitive with \MODELNAME-wo-m. Modeling missing events does not bring significant benefits to HYPERTEXT mainly because HYPERTEXT is an offline face-to-face dataset, where the proportion of missing events is relatively small. However, the experimental results also show that modeling missing events do not bring significant disadvantages even under the condition that the observed events are relatively well established. Additionally, we also find that modeling missing events improves the performance of link prediction more than it improves the performance of event time prediction. The link prediction performance improvement from modeling missing events for large, sparse graphs is more significant than that for small networks.

\subsection{Parameter Sensitivity}
The proposed \MODELNAME~involves a number of parameters that may affect its performance. In this section, we discuss two parameters that are related to the main idea of this paper: missing event ratio $Q$ and mixture component number $K$. The impacts of other parameters, such as the number of GNN layers $L$, are given in the Appendix~\ref{apd:params_sense}. To save computation time, we only test on LSED, UCI and HYPERTEXT with 5 independent experiments.

\subsubsection{Impact of missing events ratio $Q$}
\begin{figure}[h]
	\centering
	\includegraphics[width=0.49\linewidth]{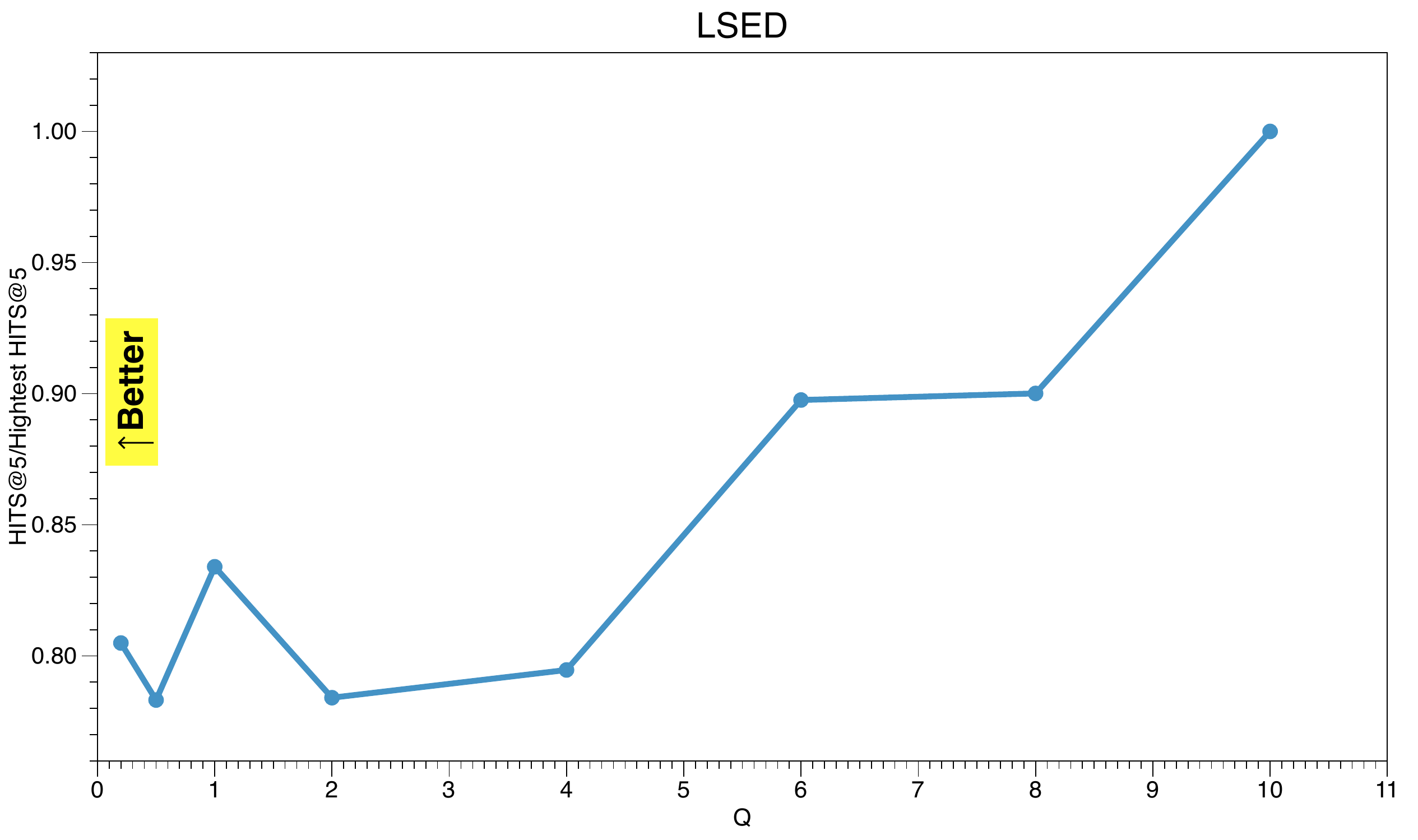}
	\includegraphics[width=0.49\linewidth]{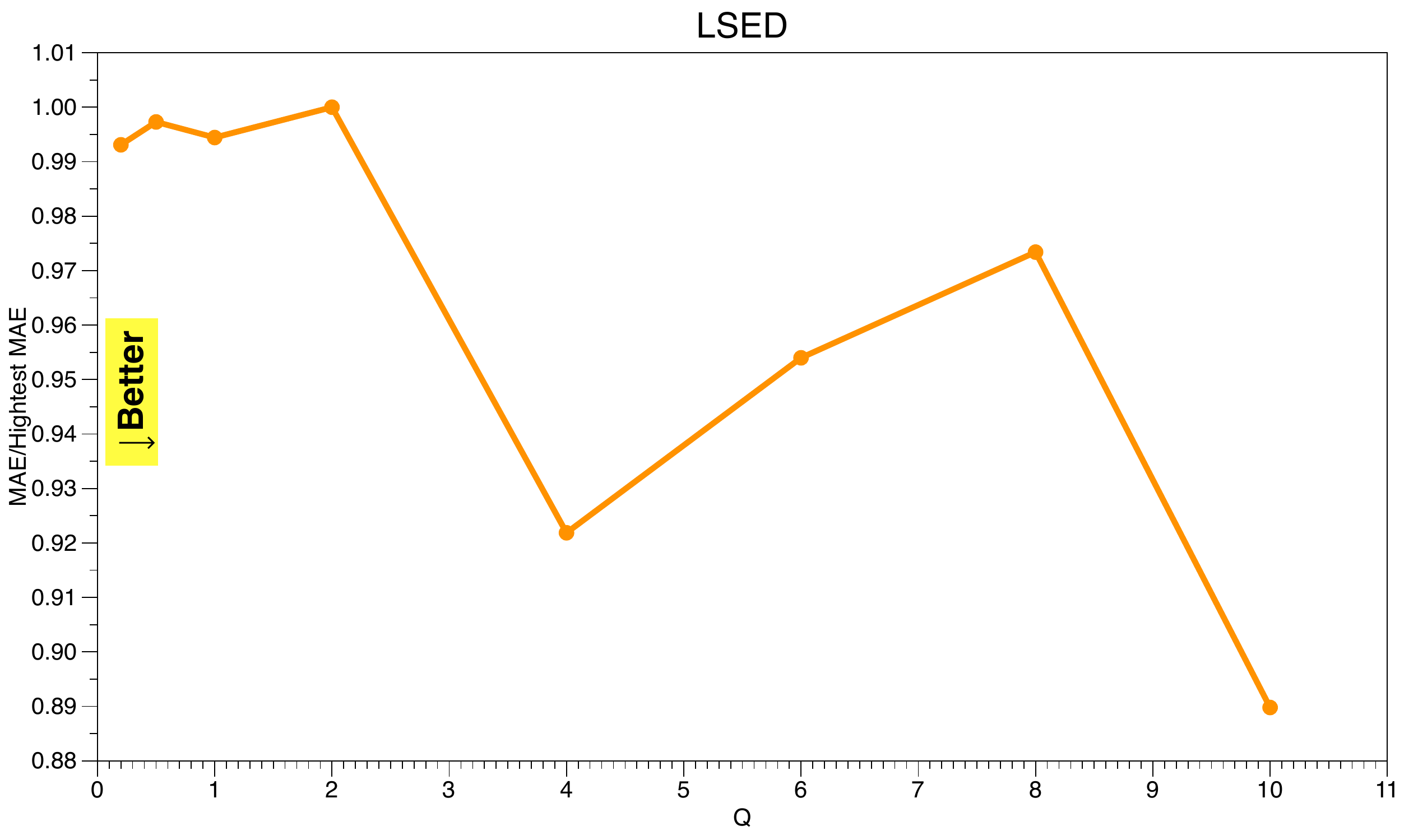}
	\includegraphics[width=0.49\linewidth]{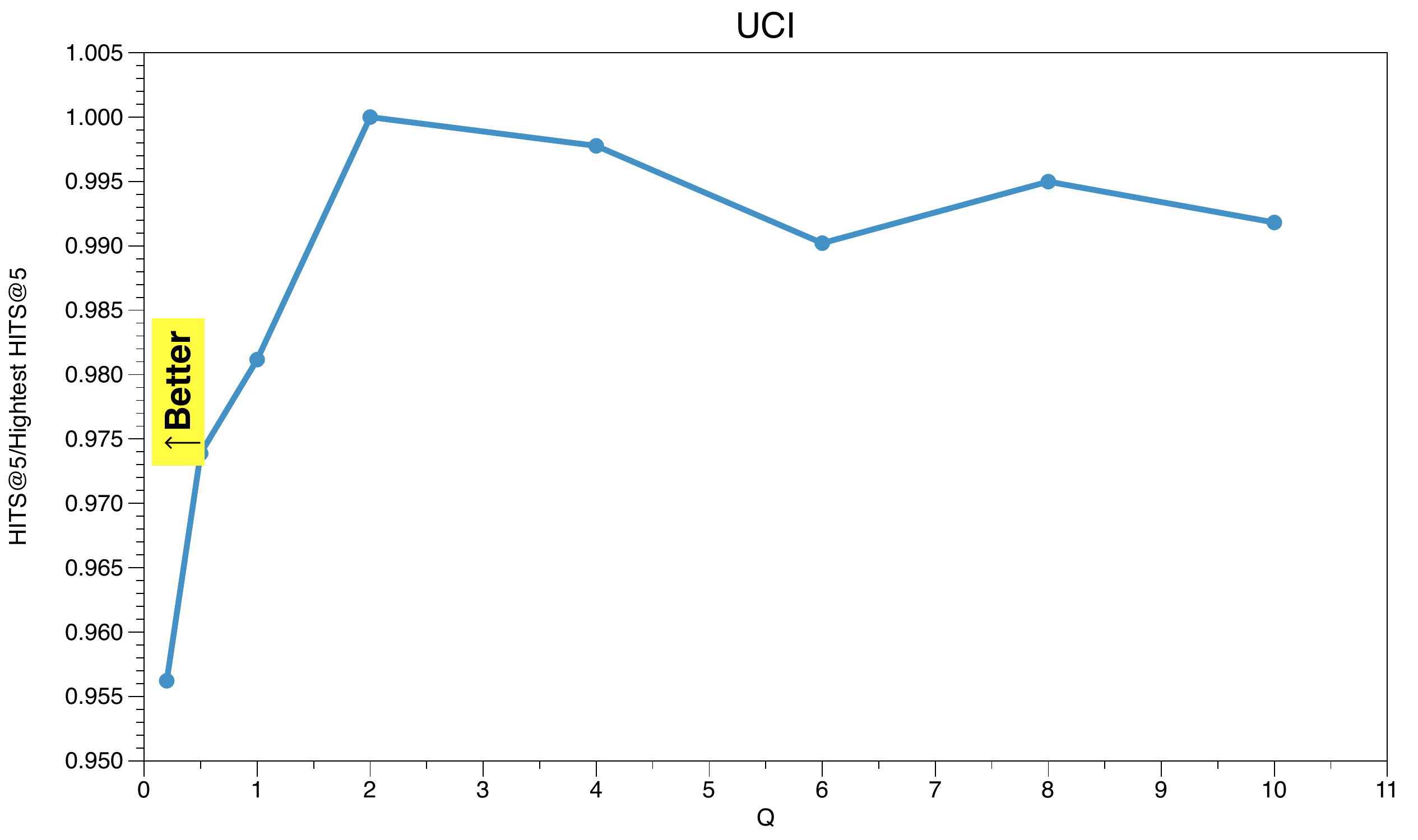}
	\includegraphics[width=0.49\linewidth]{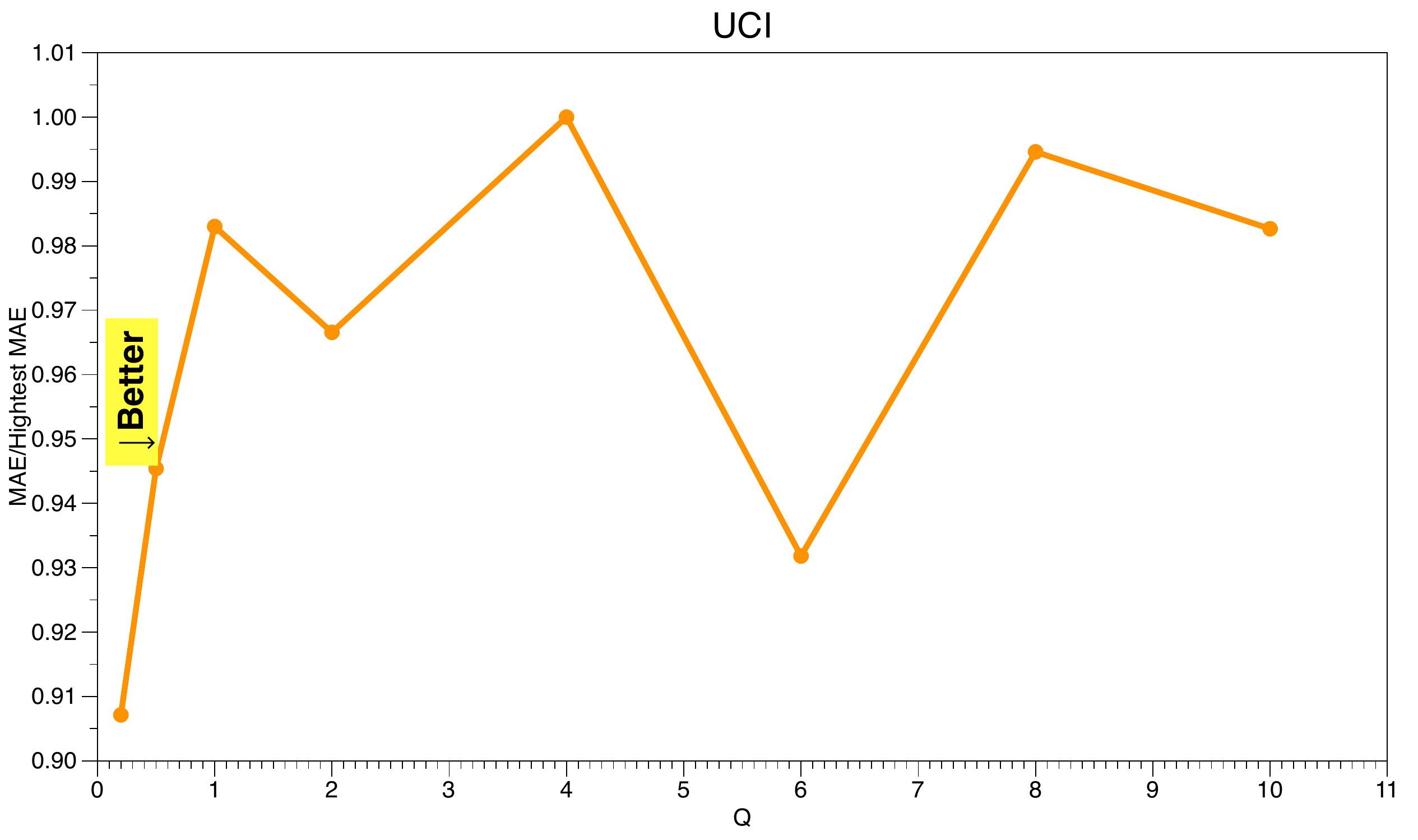}
	\includegraphics[width=0.49\linewidth]{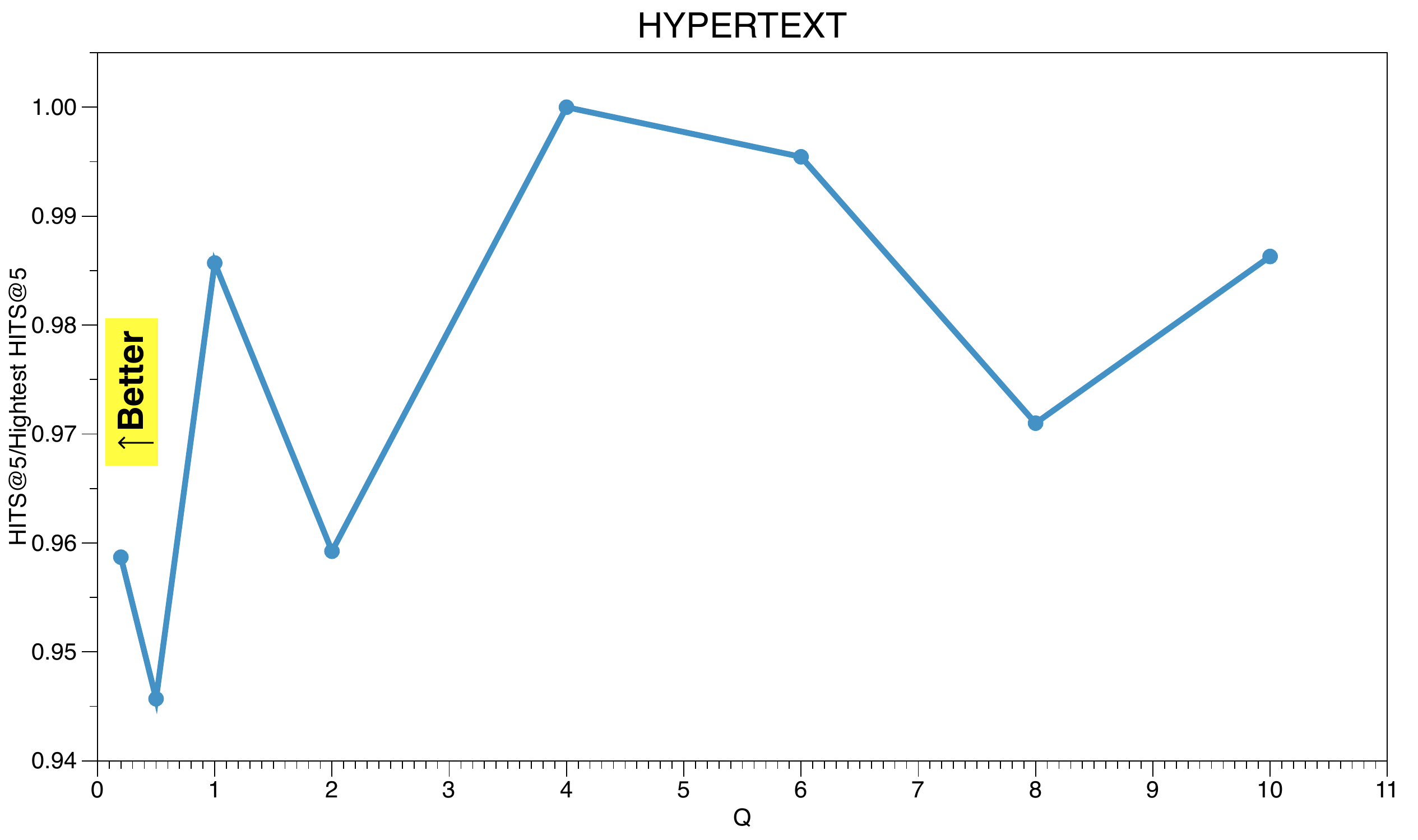}
	\includegraphics[width=0.49\linewidth]{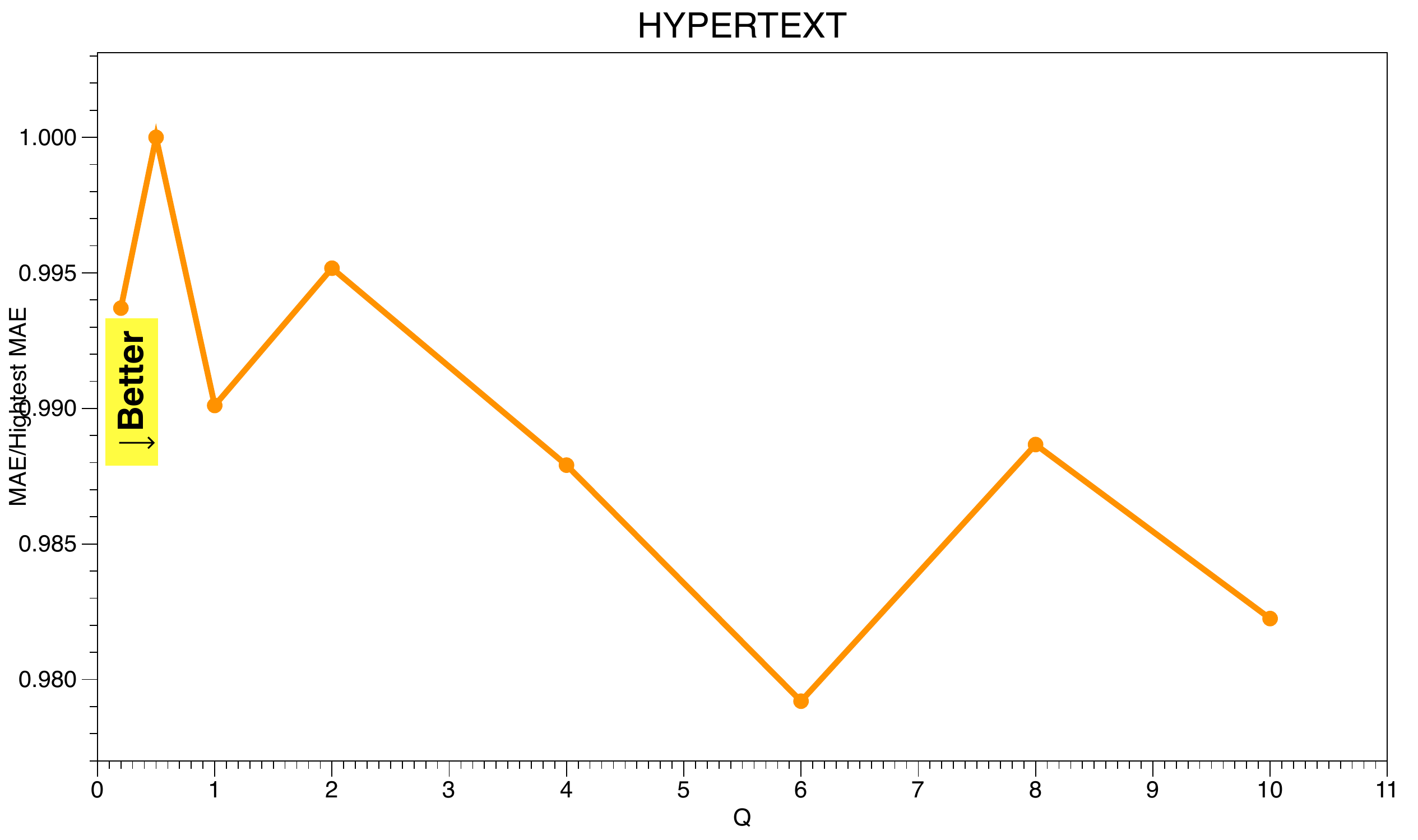}
	\caption{Relative HITS@5 and MAE score with respect to the missing events ratio $Q$ on the LSED, UCI and HYPERTEXT dataset.}\label{fig:Q}
\end{figure}
We test the cases where the missing events ratio $Q$ is \{0.2, 0.5, 1, 2, 4, 6, 8, 10\}. The results are shown in Fig.~\ref{fig:Q}. 

For the link prediction, the performance improves as $Q$ increases. However, when $Q$ increases beyond the appropriate value, the link prediction performance decrease. Link prediction performance is significantly affected by $Q$ on dataset with a large number of missing events, for example, the performance gap reaches more than 20\% on LSED, but only about 5\% on UCI and HYPERTEXT.

For the event time prediction, the performance improves with increasing $Q$ on the LSED dataset, decreases with increasing $Q$ on the UCI dataset, and remains relatively stable (about 2\% performance fluctuation) on the HYPERTEXT dataset.

In addition, increasing $Q$ will bring additional memory consumption and increase the training time, thus decreasing the efficiency of the model training. In real-world applications, we need to choose an appropriate $Q$ to trade-off the performance and efficiency.

\subsubsection{Impact of mixture component number $K$}
\begin{figure}
	\centering
	\includegraphics[width=0.48\linewidth]{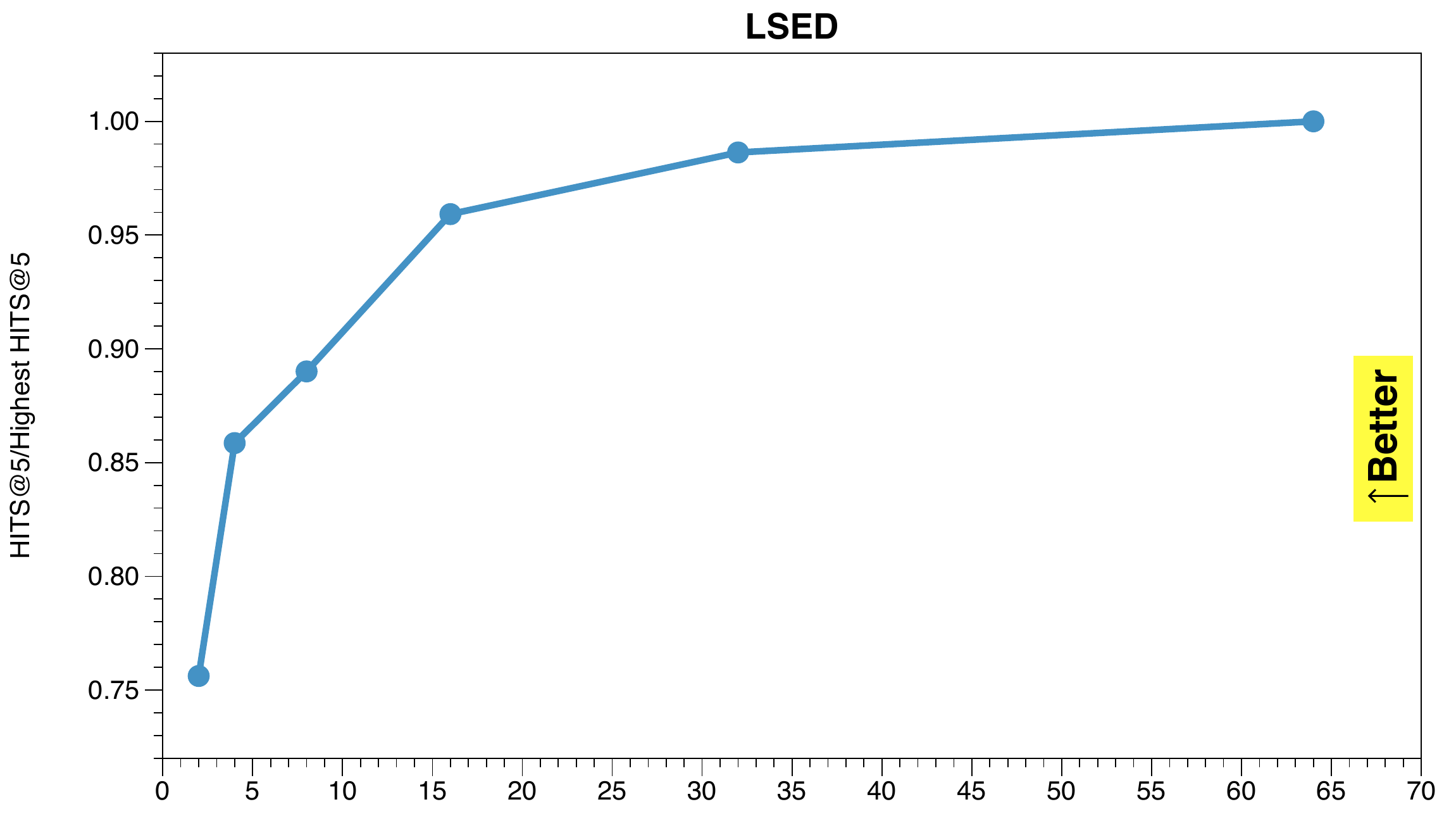}
	\includegraphics[width=0.48\linewidth]{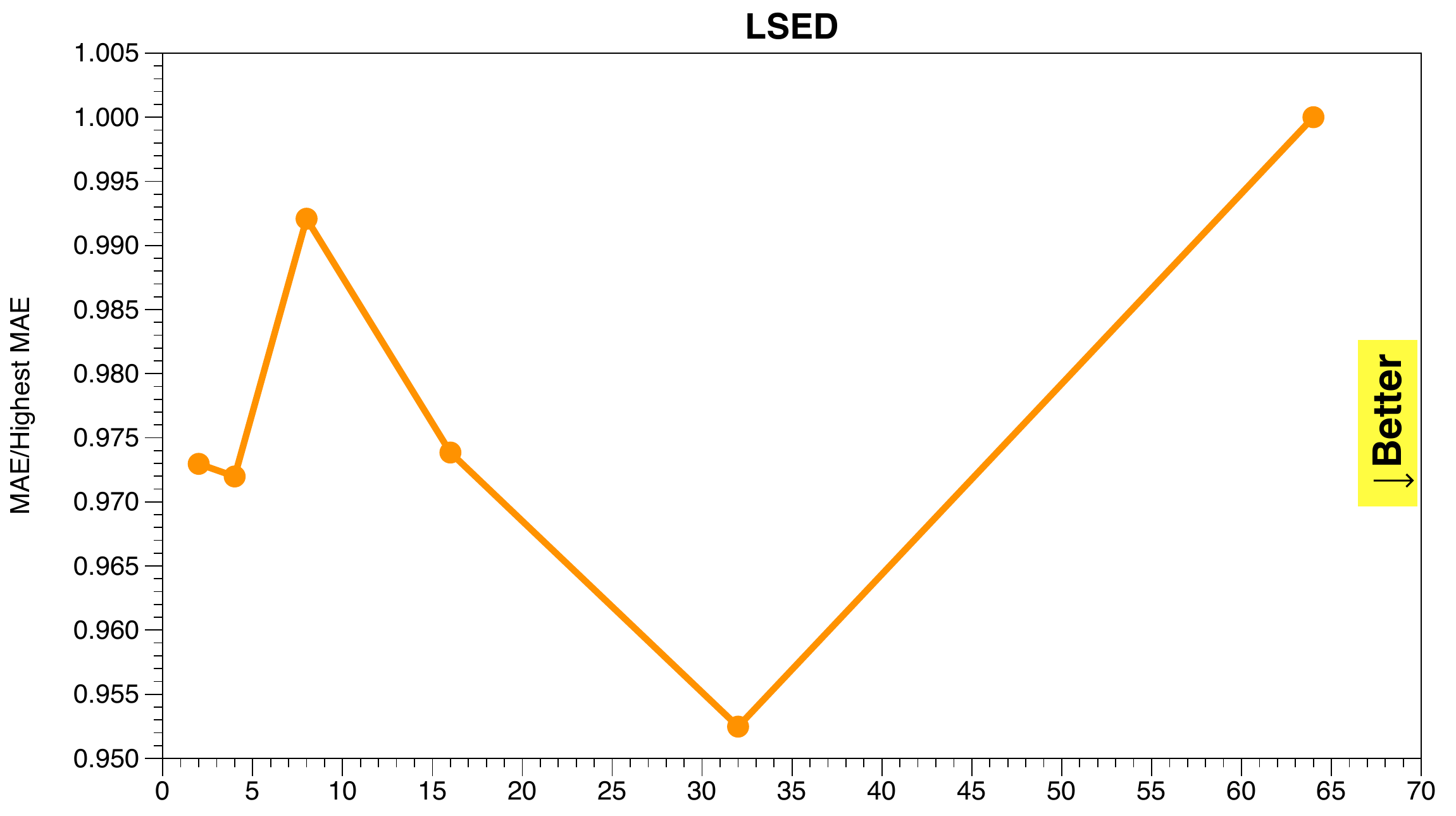}
	\includegraphics[width=0.48\linewidth]{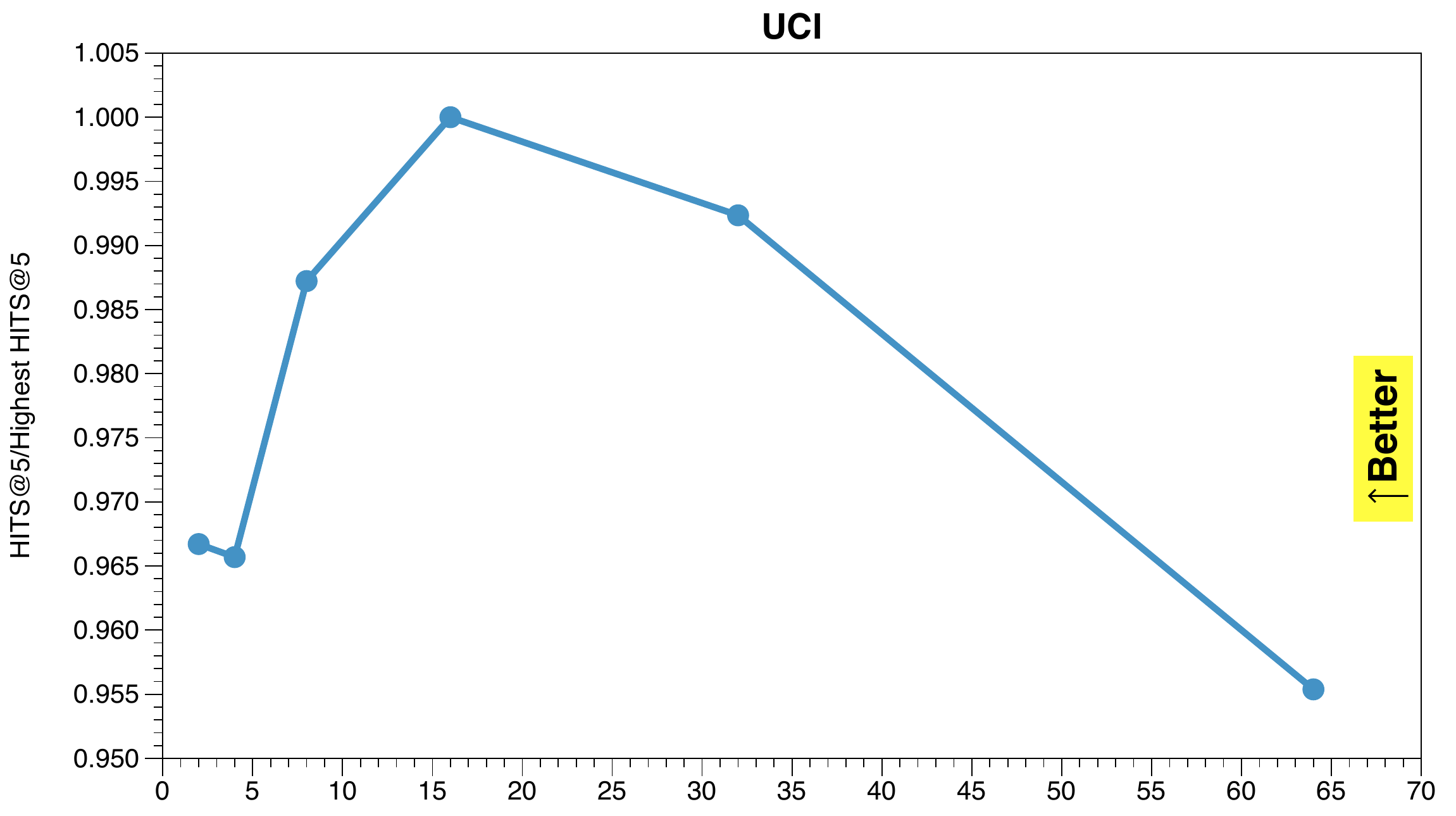}
	\includegraphics[width=0.48\linewidth]{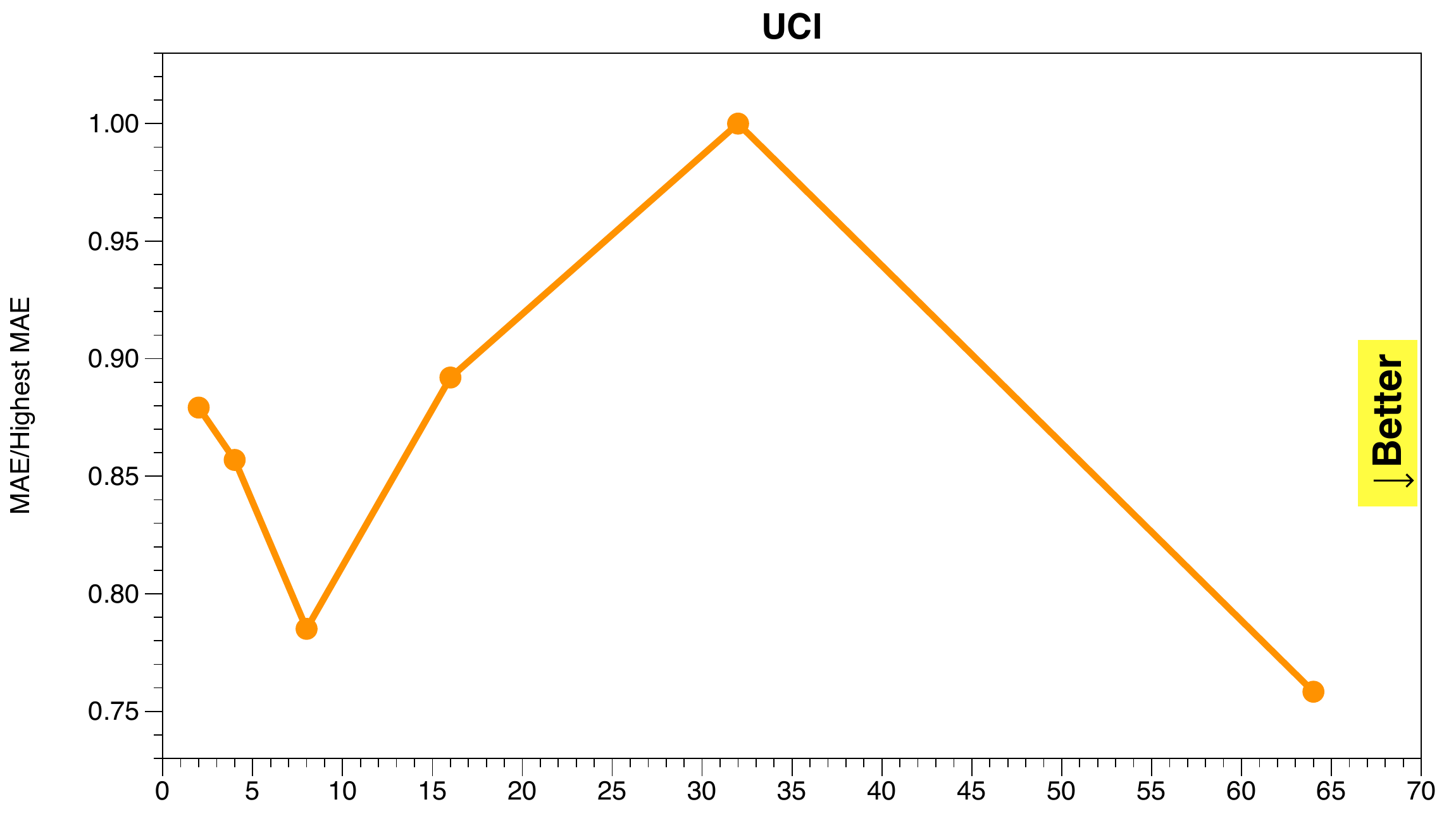}
	\includegraphics[width=0.48\linewidth]{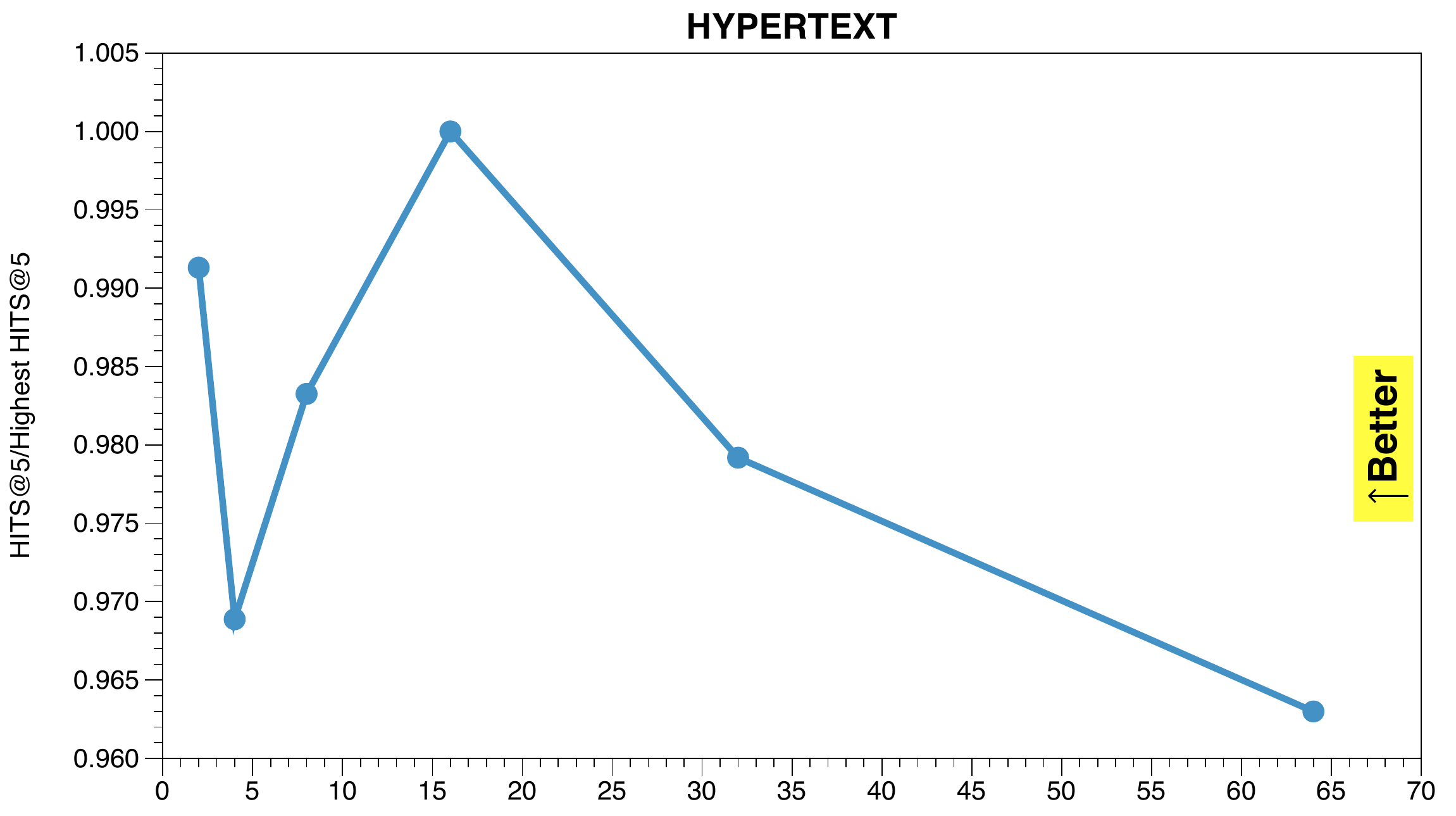}
	\includegraphics[width=0.48\linewidth]{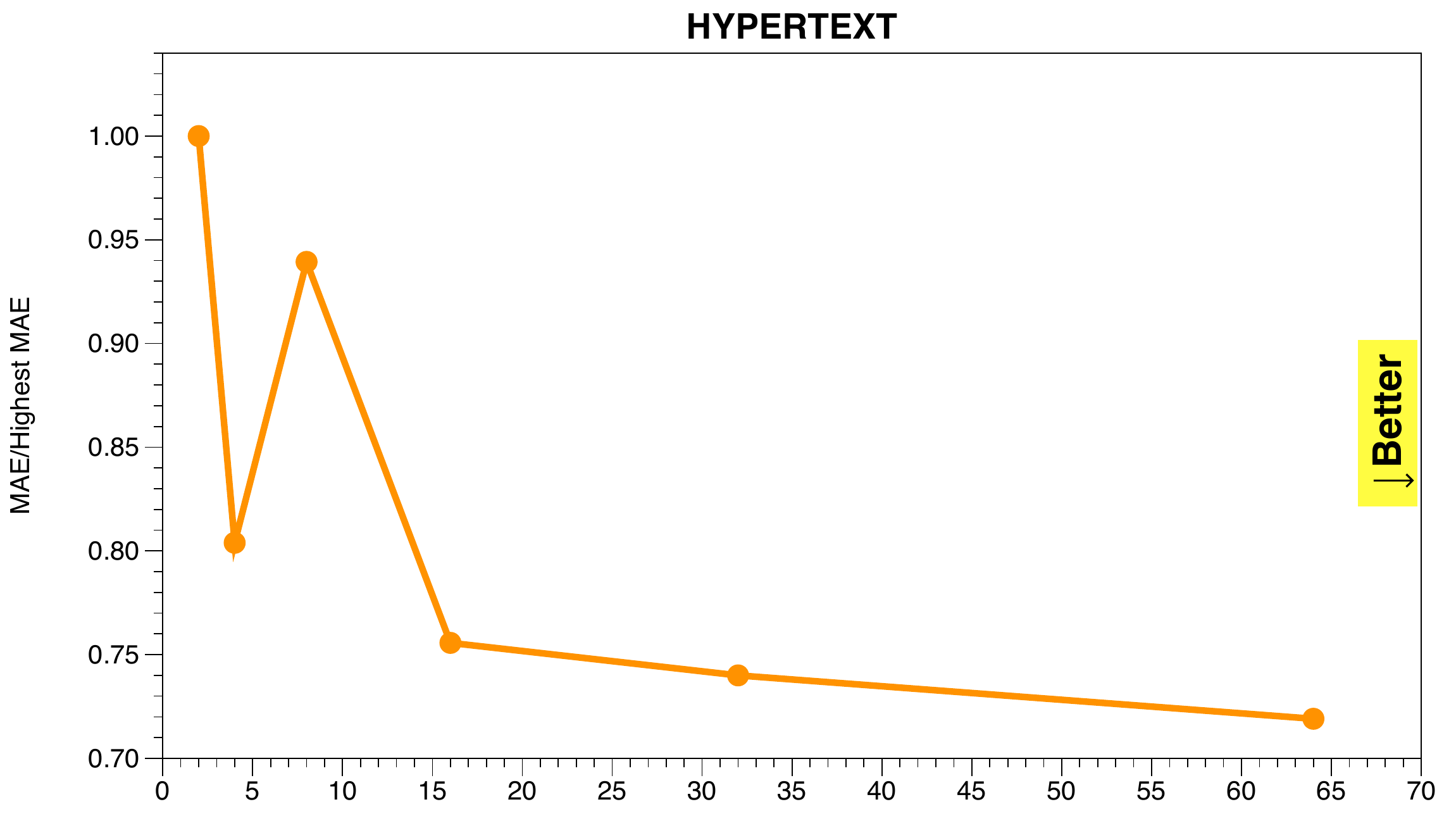}
	\caption{Relative HITS@5 and MAE score with respect to the mixture component number $K$ on the LSED, UCI and HYPERTEXT dataset.}\label{fig:K}
\end{figure}
We test the cases where the mixture component number $K$ is \{2, 4, 8, 16, 32, 64\}. The results are shown in Fig.~\ref{fig:K}.

For the link prediction, the performance on the dataset LSED is significantly affected by the value of $K$ (up to $25\%$ performance fluctuation), while on UCI and HYPERTEXT the impact is insignificant. On LSED, the link prediction performance first increases very rapidly and then slows down with increasing $K$.

In contrast to link prediction, for event time prediction, the mixture component number $K$ has a smaller effect on MAE score on LSED (about $5\%$ performance fluctuation) and a large effect on UCI and HYPERTEXT (up to $25\%$ performance fluctuation). Normally the higher is $K$, the better the event time prediction performance.

\section{Conclusion}\label{sec:conclusion}
Many complex systems in the real world can be naturally represented using graph data, and they are continuously evolving. How learning on temporal graphs is critical to leveraging these complex systems to support downstream applications.

In this paper, we propose a missing event aware temporal graph neural network called \MODELNAME. \MODELNAME~focuses on solving two challenges that are overlooked in existing temporal neural networks:  1) uniformly modeling event time and evolving graph structure; and 2) modeling missing events. For uniformly modeling event time and evolving graph structure, \MODELNAME~propose a variant message passing neural network that can learn node's structural and temporal information in one embedding. For modeling missing events, \MODELNAME~treats missing events as a latent variable and then connect it with variational autoencoders. \MODELNAME~denotes temporal graphs as a sequence of observed/missing events and use three temporal point process to model the sequence, which makes \MODELNAME~can estimate further event time. Experimental results on several real-world temporal graphs demonstrate the effectiveness of \MODELNAME. 

{In our future work, we will make every effort to improve MTGN in two aspects: 1) explore adaptive missing event ratio $Q$ instead of setting it manually; 2) improve the sampling efficiency of missing events.}

\section*{Acknowledgment}
The research in this paper is partially supported by the National Key Research and Development Program of China (No 2021YFB3300700) and the National Science Foundation of China (61832004, 61832014).

\bibliographystyle{IEEEtran}
\bibliography{main.bib}

\begin{IEEEbiography}[{\includegraphics[width=1in,height=1.25in,clip,keepaspectratio]{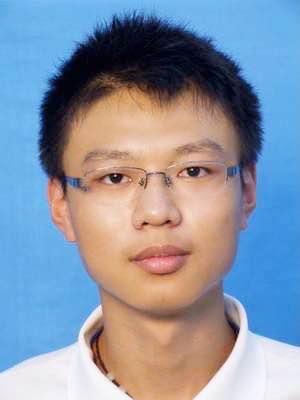}}]{Mingyi Liu}
received his B.S. degree from the School of Computer Science and Technology, Harbin Institute of Technology in 2018. He is currently pursuing the Ph.D. degree in software engineering at Harbin Institute of Technology (HIT), China. His research interests include service ecosystem model, service evolution analysis, data mining and graph neural networks.
\end{IEEEbiography}
\begin{IEEEbiography}[{\includegraphics[width=1in,height=1.25in,clip,keepaspectratio]{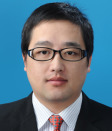}}]{Zhiying Tu}
 is an associate professor of the School of Computer Science and Technology at Harbin Institute of Technology (HIT). He holds a PhD degree in Computer Integrated Manufacturing (Productique) from the University of Bordeaux. Since 2013, he start to work at HIT. His research interest are Service Computing, Enterprise Interoperability, and Cognitive Computing. He has 20 publications as edited books and proceedings, refereed book chapters, and refereed technical papers in journals and conferences. He is the member of IEEE Computer Society, and CCF China.
\end{IEEEbiography}

\begin{IEEEbiography}[{\includegraphics[width=1in,height=1.25in,clip,keepaspectratio]{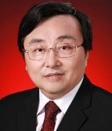}}]{Xiaofei Xu}
is a professor or at School of Computer Science and Technology, and Vice President of Harbin Institute of Technology. He received the Ph.D. degree in computer science from Harbin Institute of Technology in 1988. His research interests include enterprise intelligent computing, services computing, Internet of services, and data mining. He is the Associate Chair of IFIP TC5 WG5.8, chair of INTEROP-VLab China Pole, Fellow of China Computer Federation (CCF), and the vice director of the technical committee of service computing of CCF. He is the author of more than 300 publications. He is member of the IEEE and ACM.
\end{IEEEbiography}

\begin{IEEEbiography}[{\includegraphics[width=1in,height=1.25in,clip,keepaspectratio]{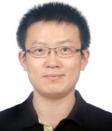}}]{Zhongjie Wang} is a professor at Faculty of Computing, Harbin Institute of Technology (HIT). He received the Ph.D. degree in computer science from Harbin Institute of Technology in 2006. His research interests include services computing, mobile and social networking services, and software architecture. He is the author of more than 80 publications. He is a member of the IEEE.\end{IEEEbiography}

\appendices
\section{Notations Table}\label{tab:notations}
\begin{table}[h]
\centering
\caption{Frequently-used  notations used in this paper}
\begin{tabular}{lp{5cm}}
\hline
Notion                                                                                                                                                                                                                                                    & Description                                                                                         \\ \hline
$u,v$                                                                                                                                                                                                                                               & nodes in temporal graph                                                                             \\
$t,t'$                                                                                                                                                                                                                                              & timestamp of an event                                                                               \\
$\bar{t}$                                                                                                                                                                                                                          & timestamp of last observed event                                                                    \\
$\bar{t}^o_{u,v},\bar{t}^m_{u,v}$                                                                                                                                         & the last timestamp of either node $u$ or node $v$ involved in an observed/missing event \\
$\mathcal{O}_t$,$\mathcal{M}_t$                                                                                                                                                                           & observed and generated missing events at time $t$                                             \\
$\mathcal{G}^*_t$                                                                                                                                                                                                                  & all observed and generated missing events until time $t$                                      \\
$\mathbf{W}^{\bullet}_{\bullet}$, $\mathbf{U}^{\bullet}_{\bullet}$                                  & weights in message passing frameworks                                                               \\
$\mathbf{o}^{\bullet}_u$, $\mathbf{m}^{\bullet}_u$,$\mathbf{g}^{\bullet}_u$ & different observed/missing/overall node embeddings of node $u$                                \\
$\mathbf{\bar{o}}^t$, $\mathbf{\bar{m}}^{t}$,$\mathbf{\bar{g}}^t$            & different observed/missing/overall graph-level embeddings of the temporal graph at time $t$   \\ \hline
\end{tabular}
\end{table}

\section{Derivations}\label{apd:derivations}
\subsection{Derivation of Eq.~\eqref{eq:pg}}
$$
\begin{aligned}
p_{\theta}(\mathcal{O}_T) & = \prod_{t=1}^T \int_{\mathcal{M}_{t}} p_{\theta}(\mathcal{O}_t|\mathcal{G}^*_{\bar{t}}, \mathcal{M}_{t})p_{\theta}( \mathcal{M}_{t} )\mathrm{d} \mathcal{M}_{t}\\
&= \prod_{t=1}^T \int_{\mathcal{M}_{t}} p_{\theta}(\mathcal{O}_t|\mathcal{G}^*_{\bar{t}}, \mathcal{M}_{t})p_{\theta}( \mathcal{M}_{t} )\frac{q_{\phi}(\mathcal{M}_{t} | \mathcal{O}_t, \mathcal{G}^*_{\bar{t}})}{q_{\phi}(\mathcal{M}_{t} | \mathcal{O}_t, \mathcal{G}^*_{\bar{t}})} \mathrm{d} \mathcal{M}_{t}\\
&= \prod_{t=1}^T \int_{\mathcal{M}_{t}} \frac{p_{\theta}(\mathcal{O}_t|\mathcal{G}^*_{\bar{t}}, \mathcal{M}_{t})p_{\theta}( \mathcal{M}_{t} )}{q_{\phi}(\mathcal{M}_{t} | \mathcal{O}_t, \mathcal{G}^*_{\bar{t}})}q_{\phi}(\mathcal{M}_{t} | \mathcal{O}_t, \mathcal{G}^*_{\bar{t}}) \mathrm{d} \mathcal{M}_{t} \\
&= \mathbb{E}_{q_{\phi}} \prod_{t=1}^T \frac{ p_{\theta}(\mathcal{O}_t|\mathcal{G}^*_{\bar{t}}, \mathcal{M}_{t})p_{\theta}(\mathcal{M}_{t} |\mathcal{G}^*_{\bar{t}})}{q_{\phi}(\mathcal{M}_{t} | \mathcal{O}_t, \mathcal{G}^*_{\bar{t}})} \\
\end{aligned}
$$

\subsection{Derivation of Eq.~\eqref{eq:elbo}}
$$
\begin{aligned}
\log p(&\mathcal{O}_T) = \log \mathbb{E}_{q_{\phi}} \prod_{t=1}^T \frac{ p_{\theta}(\mathcal{O}_t|\mathcal{G}^*_{\bar{t}}, \mathcal{M}_{t})p_{\theta}(\mathcal{M}_{t} |\mathcal{G}^*_{\bar{t}})}{q_{\phi}(\mathcal{M}_{t} | \mathcal{O}_t, \mathcal{G}^*_{\bar{t}})} \\
&\ge \mathbb{E}_{q_\phi} \log \prod_{t=1}^T \frac{ p_{\theta}(\mathcal{O}_t|\mathcal{G}^*_{\bar{t}}, \mathcal{M}_{t})p_{\theta}(\mathcal{M}_{t} |\mathcal{G}^*_{\bar{t}})}{q_{\phi}(\mathcal{M}_{t} | \mathcal{O}_t, \mathcal{G}^*_{\bar{t}})} \\
&= \mathbb{E}_{q_\phi}  \sum_{t=1}^T  \Big(\log p_{\theta}(\mathcal{O}_t|\mathcal{G}^*_{\bar{t}}) +\log \frac{ p_{\theta}(\mathcal{M}_{t} |\mathcal{G}^*_{\bar{t}})}{q_{\phi}(\mathcal{M}_{t} | \mathcal{O}_t, \mathcal{G}^*_{\bar{t}})} \Big) \\
&= \mathbb{E}_{q_\phi}  \sum_{t=1}^T \log p_{\theta}(\mathcal{O}_t|\mathcal{G}^*_{\bar{t}}) {-}  \sum_{t=1}^T\text{KL}[q_{\phi}(\mathcal{M}_{t} | \mathcal{O}_t, \mathcal{G}^*_{\bar{t}})||p_{\theta}(\mathcal{M}_{t} |\mathcal{G}^*_{\bar{t}})]
\end{aligned}
$$
The seconde line holds as Jensen inequality.

\section{Implement details}\label{apd:implement}
For all baselines and \MODELNAME, we uniformly set the the max training epochs to $1,000$ and the node embeddings dimension to $64$ on all datasets. We use the AdamW optimizer for methods. The $weight\_decay$ is set to $0.00005$ and the learning rate is searched from $\{0.01, 0.001, 0.0001, 0.00002, 0.00001\}$.
The best result within 1000 epochs in each independent experiment was taken as the result of that experiment. All experiments are running on a Ubuntu server with NVIDIA GeForce RTX 3080Ti, Intel(R) Xeon(R) CPU E5-2630 v4 @ 2.20GHz and 32GB RAM. The softwares used in this paper include: PyTorch-1.10.0+cu111, PyTorchLightning-1.5.4 and PyG-2.0.2.
The implementation details and other detailed parameters for all baselines and \MODELNAME~are listed as follows:
\begin{itemize}
	\item \textbf{GCN}. We use the code provided by PyG, we stacked 2 GCN layers.
	\item \textbf{GAT}. We use the code provided by PyG, we stacked 2 GAT layers.
	\item \textbf{EvoveGCN}.  We adopted the code provide by PyGT\cite{rozemberczki2021pytorch}.
	\item \textbf{DySAT}. We have reproduced it using PyG based on the official Tensorflow code. Based on discussion in \cite[Fig.~5]{DySAT} and their official code, we stack $2$ structural self-attention layer and set all self-attention layers (both structural and temporal) with $8$ heads.
	\item \textbf{KnowEvolve}. We have reproduced it using PyG based on the official C++ implement\footnote{https://github.com/rstriv/Know-Evolve}.
	\item \textbf{EvoKG}. We have reproduced it using PyG based on the official DGL implement\footnote{https://github.com/NamyongPark/EvoKG}. We stack 2 GNN layers, the number of time steps for BPTT is set to $5$, and set the mixture component number to 16.
	\item \textbf{\MODELNAME}.  We implement \MODELNAME~using PyG. We stack $2$ GNN layers, set the mixture component number $K$ to 16, the number of time steps for BPTT is set to $5$, and set the missing events ratio $Q$ to 1.0 on all datasets.
\end{itemize}

\section{Other Parameters Sensitivity}\label{apd:params_sense}

\begin{figure}[h]
	\centering
	\includegraphics[width=0.48\linewidth]{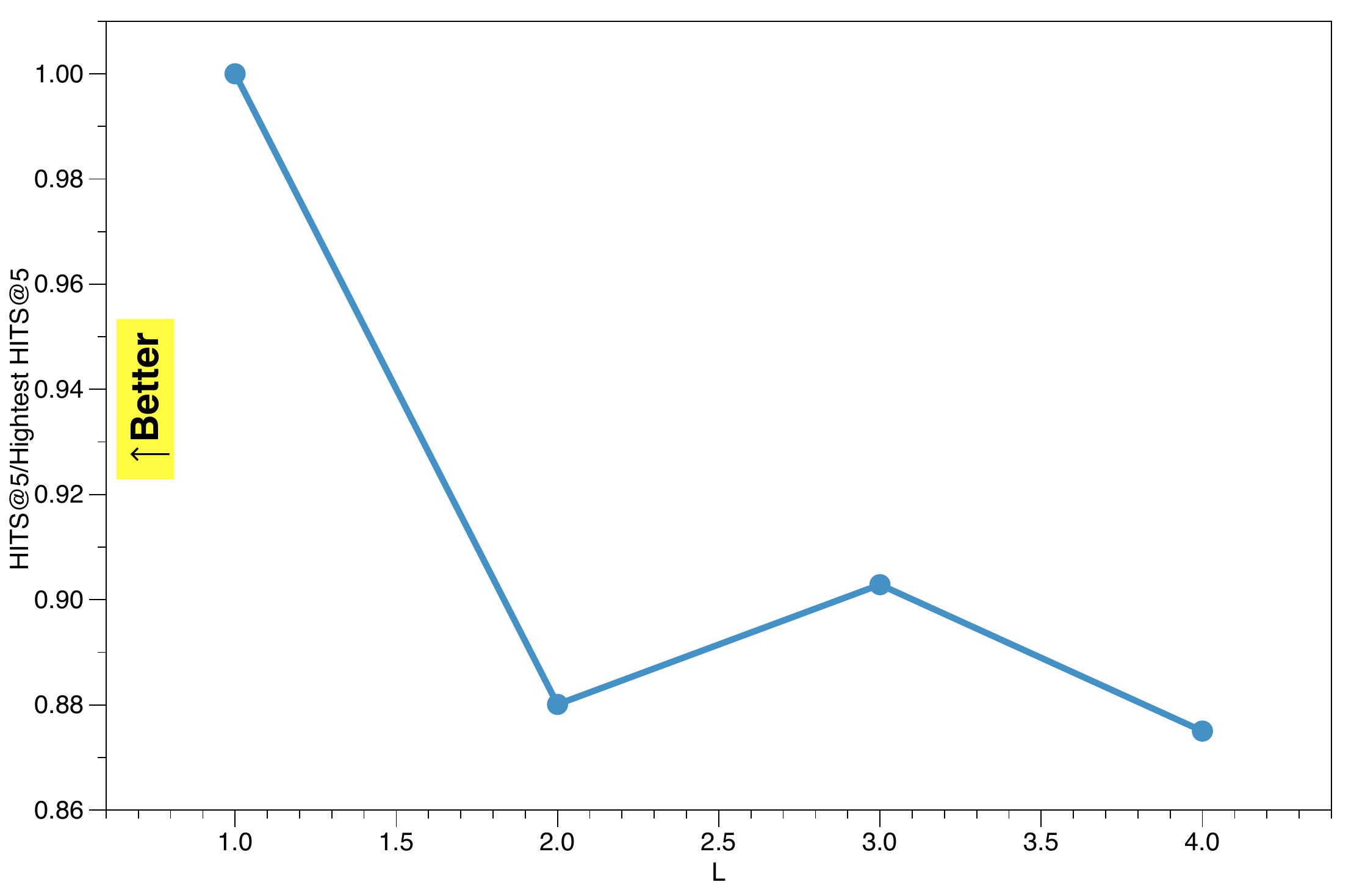}
	\includegraphics[width=0.48\linewidth]{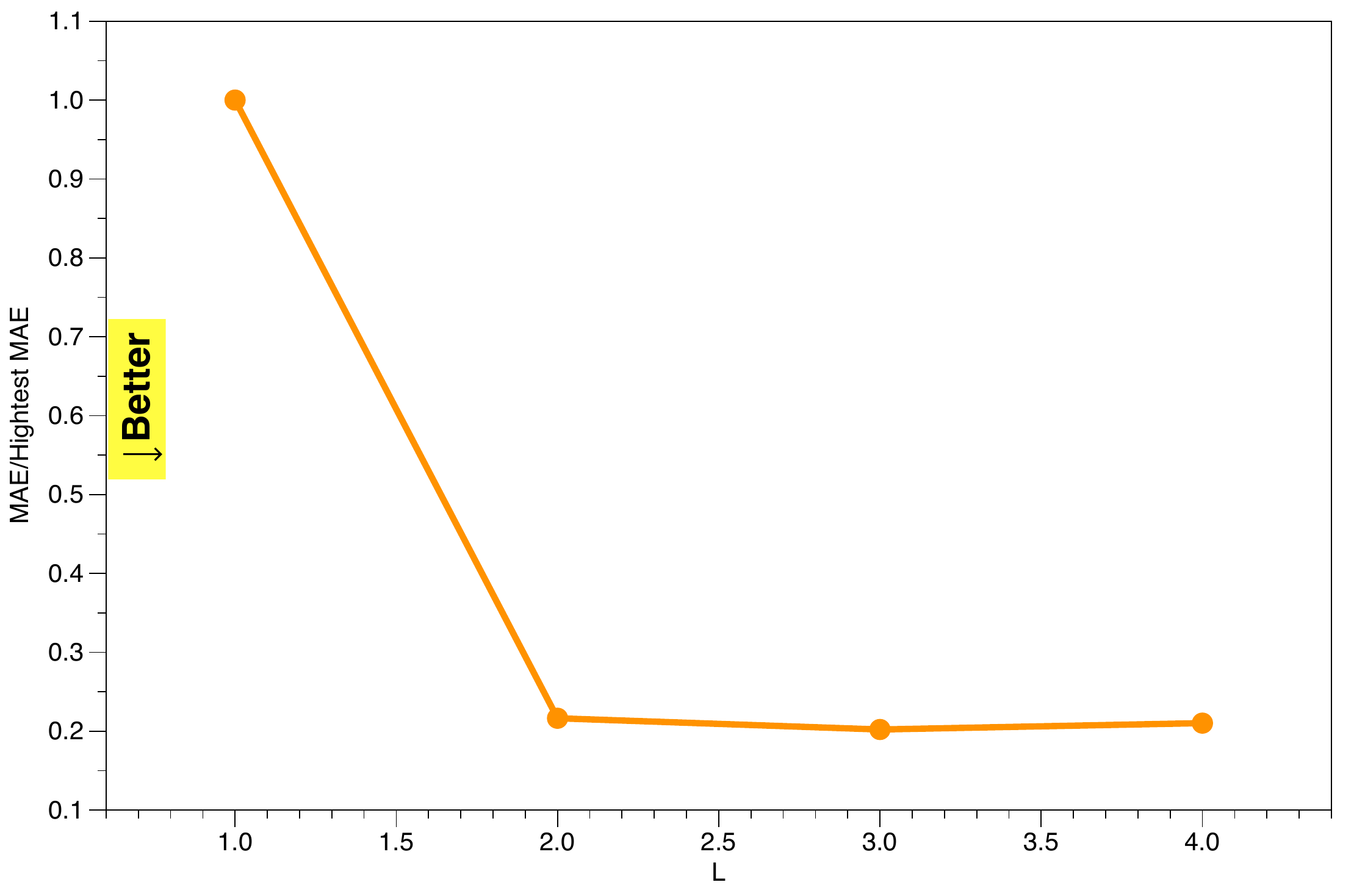}
	\caption{Relative HITS@5 and MAE score with respect to the number of GNN layers $L$ on the HYPERTEXT dataset.}\label{fig:params_L}
\end{figure}

\textbf{Number of GNN Layers $L$}. We test the cases where the number of GNN layers $K$ is $\{1, 2, 3, 4\}$. The results are shown in Fig.~\ref{fig:params_L}. On HYPERTEXT dataset, \MODELNAME~achieves best link prediction performance when using only a single GNN layer, but the event time prediction performance is poor at this setting. To trade-off the link prediction performance, event time prediction performance and model efficiency, it is appropriate to stack 2 GNN layers in practice.

\begin{figure}[h]
	\centering
	\includegraphics[width=0.48\linewidth]{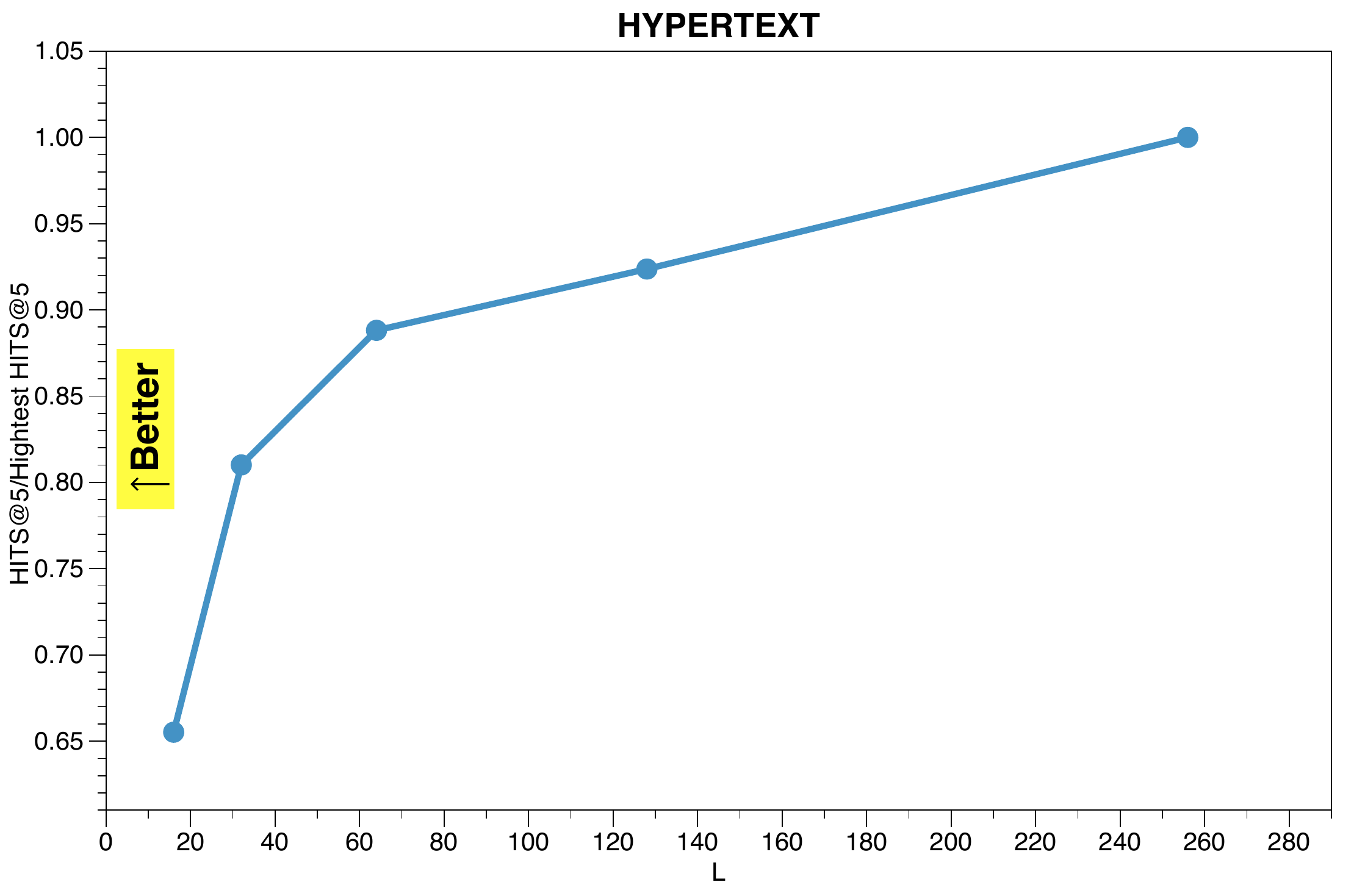}
	\includegraphics[width=0.48\linewidth]{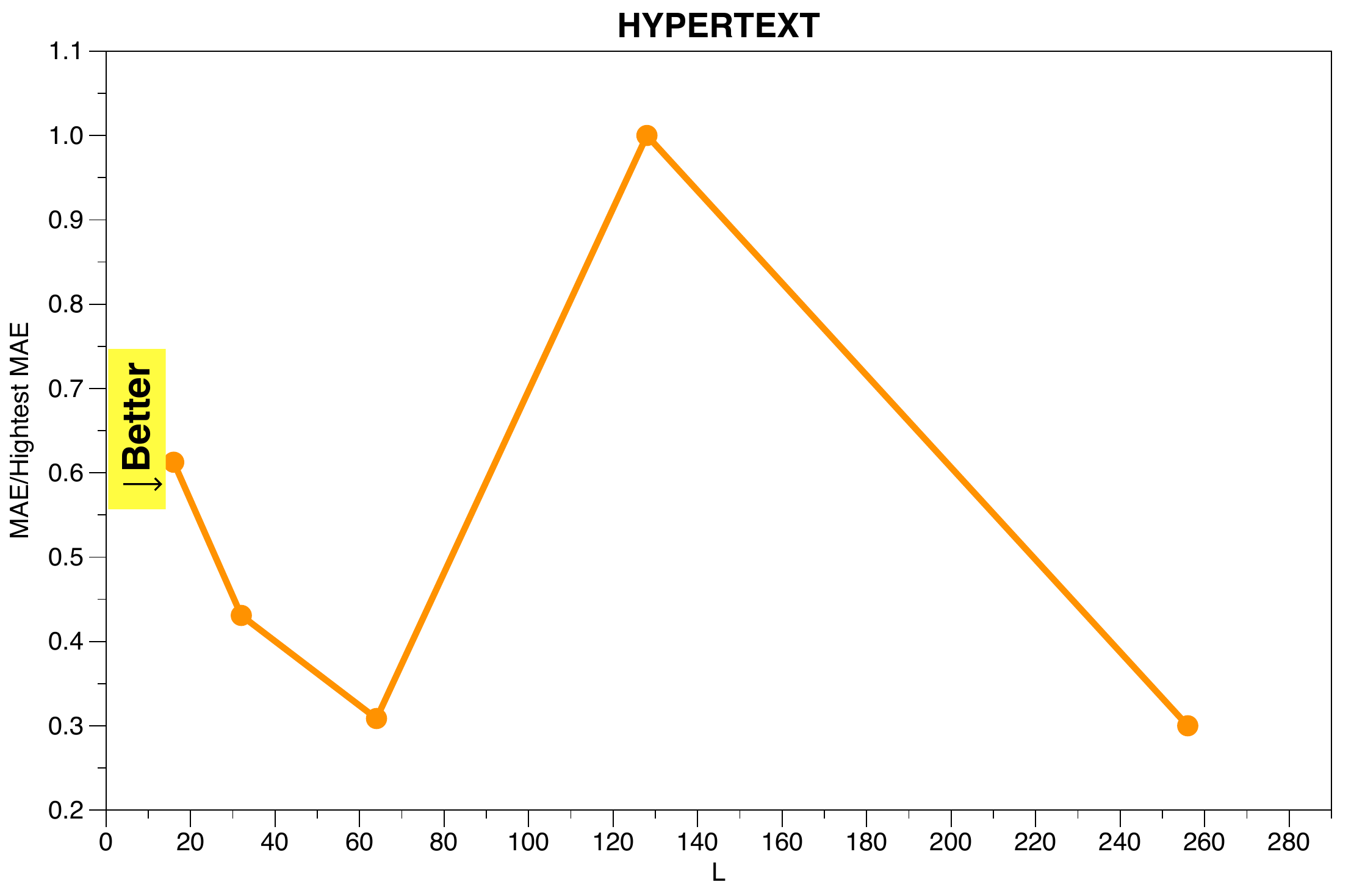}
	\includegraphics[width=0.48\linewidth]{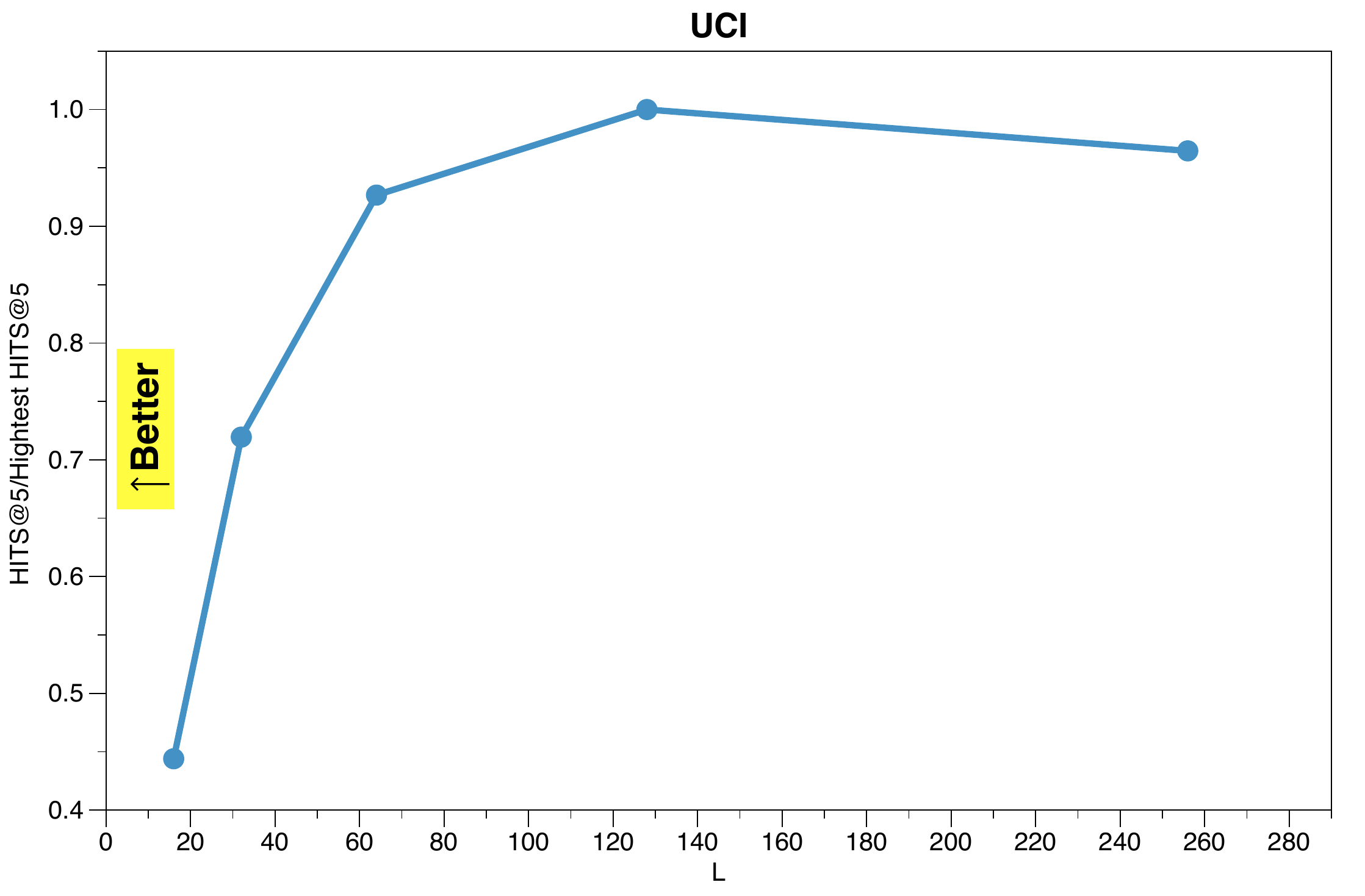}
	\includegraphics[width=0.48\linewidth]{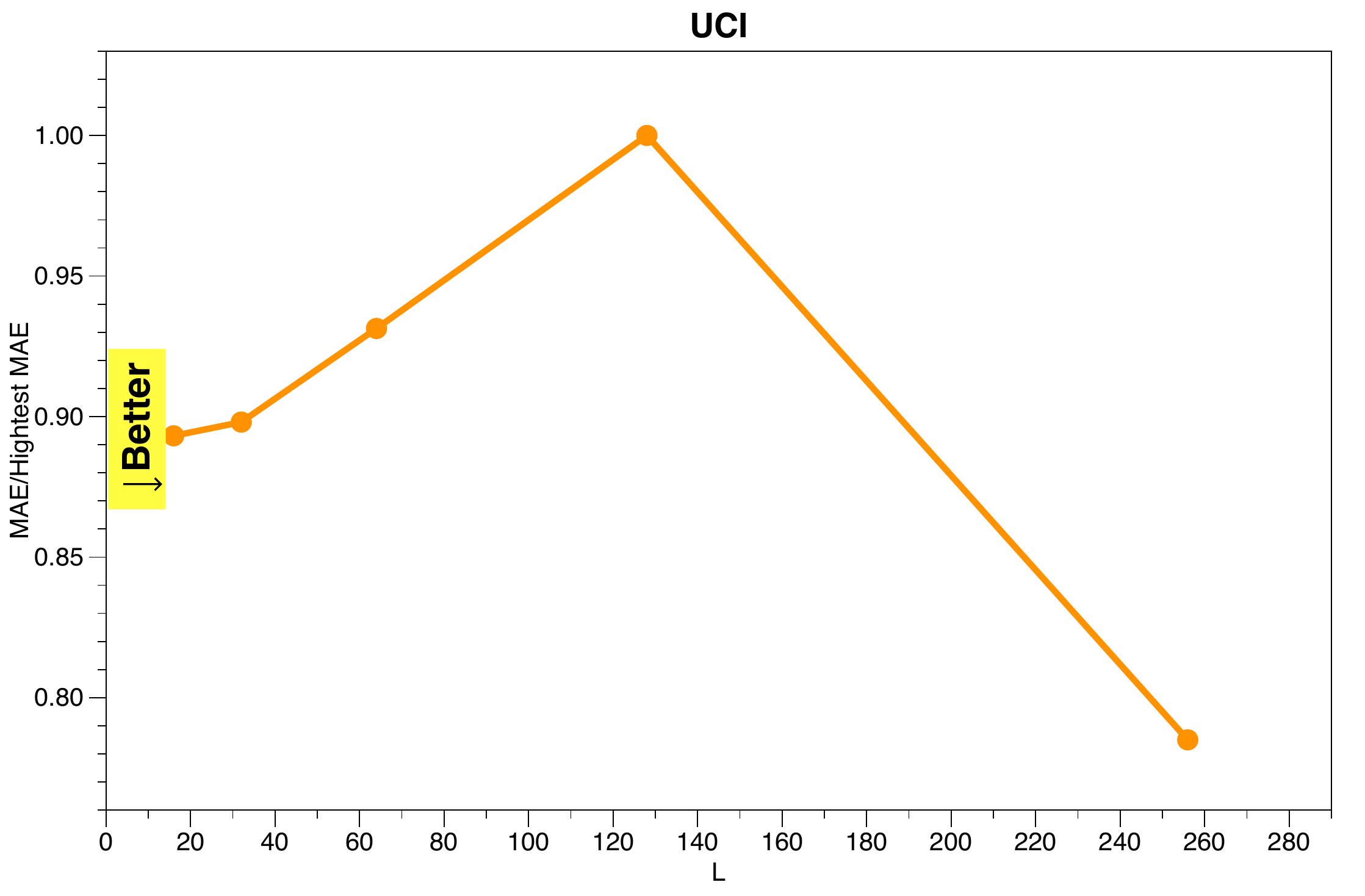}
	\caption{Relative HITS@5 and MAE score with respect to the node embedding dimension on the UCI and HYPERTEXT dataset.}\label{fig:params_D}
\end{figure}

\textbf{Node Embedding Dimension}. We test the cases where the node embedding dimension is $\{8, 16,32, 64, 128, 256\}$. As shown in Fig.~\ref{fig:params_D}, large embedding dimension can benefit both link prediction and event time prediction.

\section{Impact of model time interval in Eq.(11) and Eq.(12)}
To explore how MTGN is affected if the time interval $t-\bar{t}^o_{u,v}$ and $t'-\bar{t}^m_{u,v}$ are not encoded into Eq.(11) and Eq.(12), we perform the following test:
\begin{itemize}
    \item \textbf{MTGN-w-t}: In this variant, we remove the time interval from Eq.(11) and Eq.(12).
\end{itemize}

The results are shown in Fig.9. For the future link prediction task, whether to encode time interval in Eq.(11) and Eq.(12) performs differently on different datasets, the performance of MTGN and MTGN-w-t is close on ENRON, MTGN-w-t outperforms MTGN on HYPERTEXT and LSED, and MTGN outperforms MTGN-w-t on RT-POL and UCI. For the event time prediction task, MTGN has a competitive performance over MTGN-w-t on the UCI dataset and significantly outperforms MTGN-w-t on the other four datasets.
    \begin{figure}[h]
        \centering
        \subfigure[HITS@10] {
            \includegraphics[width=0.46\linewidth]{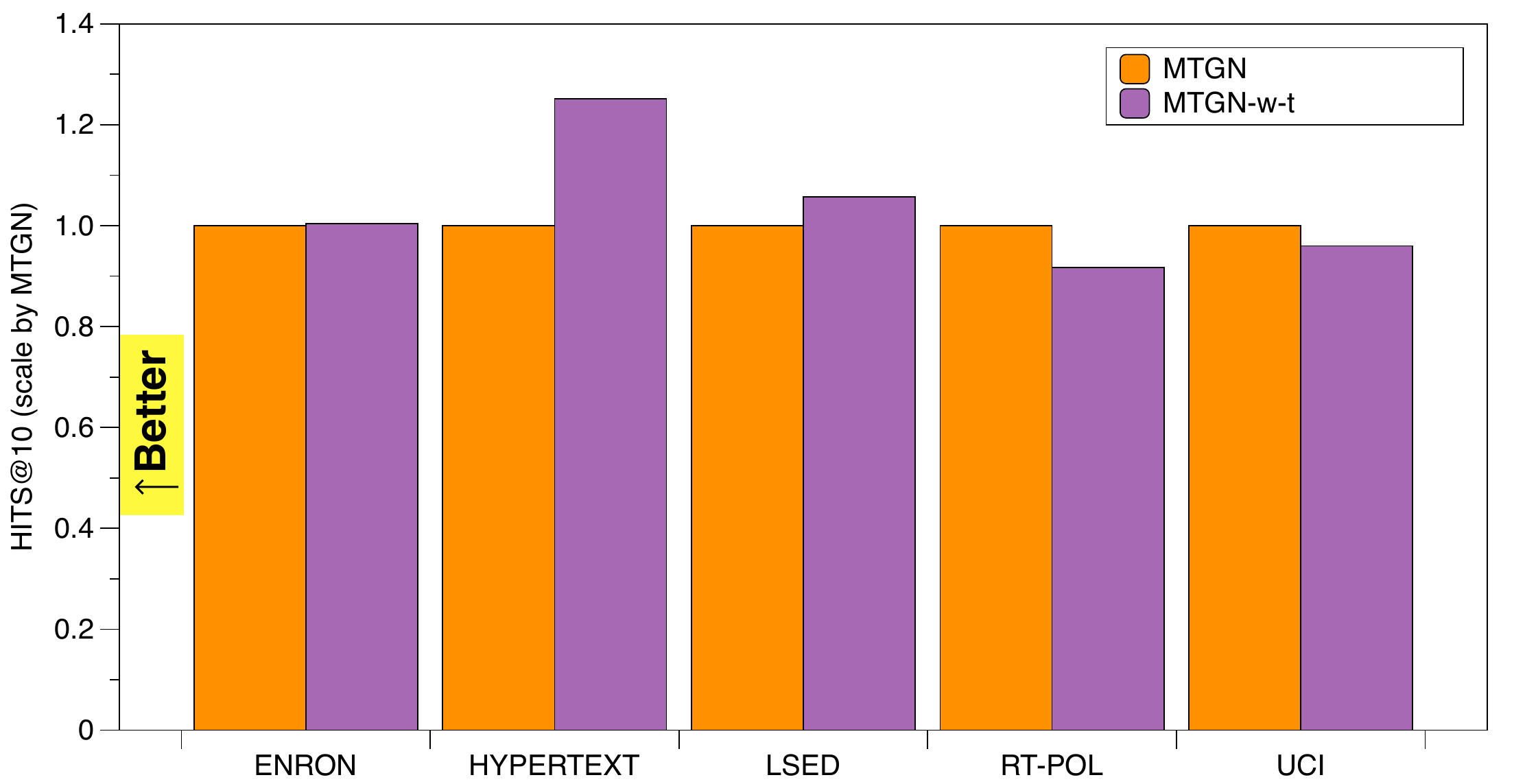}
        }
        \subfigure[MAE] {
            \includegraphics[width=0.46\linewidth]{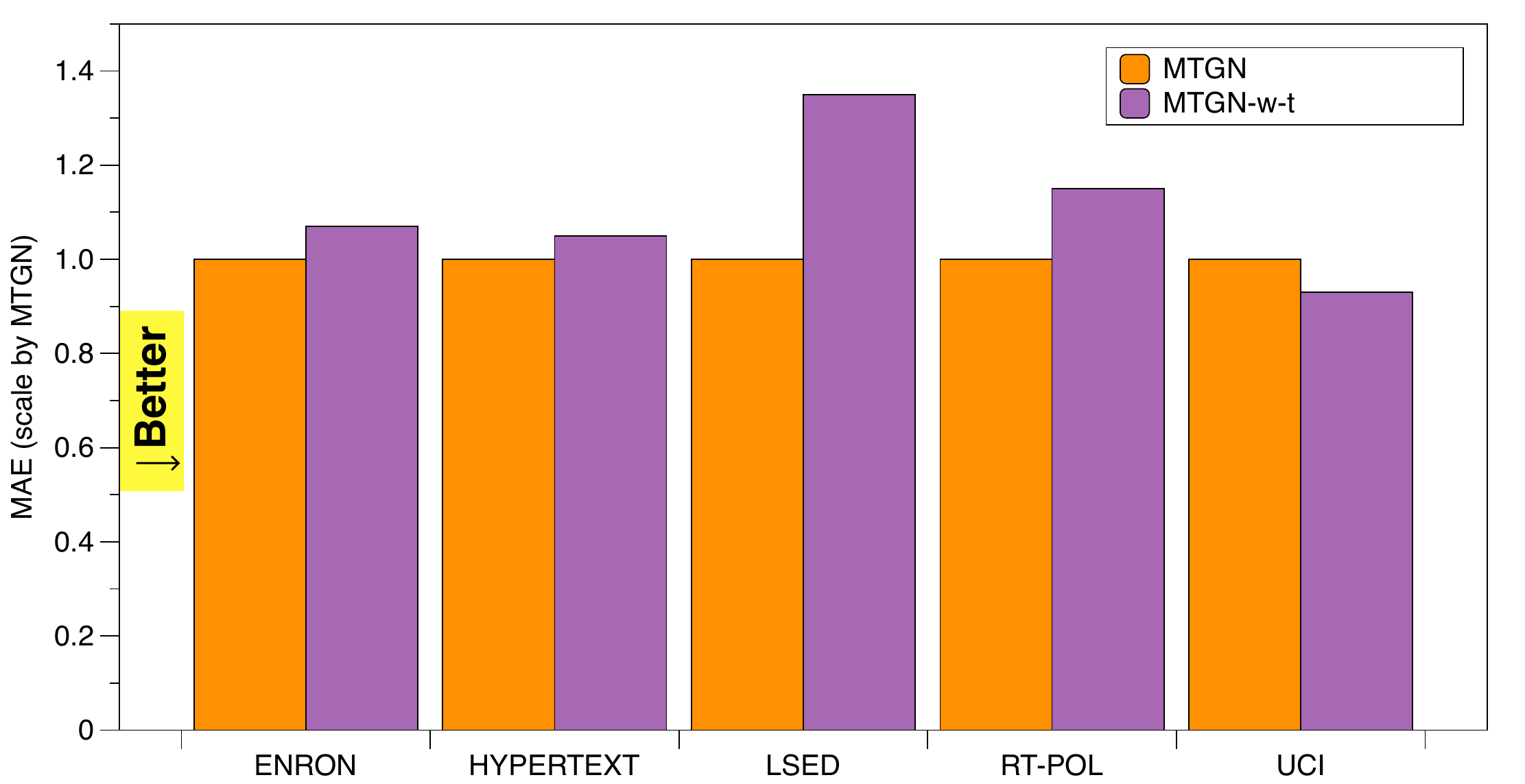}
        }
        \caption{Performance comparison for MTGN and its variant MTGN-w-t. All metrics are scale by the results on MTGN.}
    \end{figure}

Overall, encoding time intervals into Eq.(11) and Eq.(12) may lead to a slight decrease in future link prediction performance in some cases but can significantly improve the event time prediction performance.

\section{More experiment on $Q$ and masking observed events}
We test the cases where the percentage $z$ of masked observed events is \{0.1, 0.2, 0.3, 0.4, 0.5,0.6,0.7, 0.8\}, respectively, and choose the following three strategies to set missing events ratio $Q$:
\begin{enumerate}
    \item \textbf{MTGN-fixed-Q}: In this strategy, we always set $Q$ to a default value $1$. This simulates the situation when we know there are missing events but no more details.
    \item \textbf{MTGN-adaptive-Q1}: In this strategy, we suppose the observed events (before masked) are all events in a temporal graph. Then, under different mask percentage $z$, the missing events ratio $Q$ is set to $\frac{z}{1-z}$. This makes sure that in each training epoch, the model uses the same number of events with MTGN-w-m (MTGN without modeling missing event, without masking observed events).
    \item \textbf{MTGN-adaptive-Q2}: In this strategy, under different mask percentage $z$, the missing events ratio is set to $\frac{1+z}{1-z}$. This makes sure that in each training epoch, the model uses the same number of events with default MTGN (without masking observed events).
\end{enumerate}
\begin{figure}[h]
	\centering
	\includegraphics[width=0.49\linewidth]{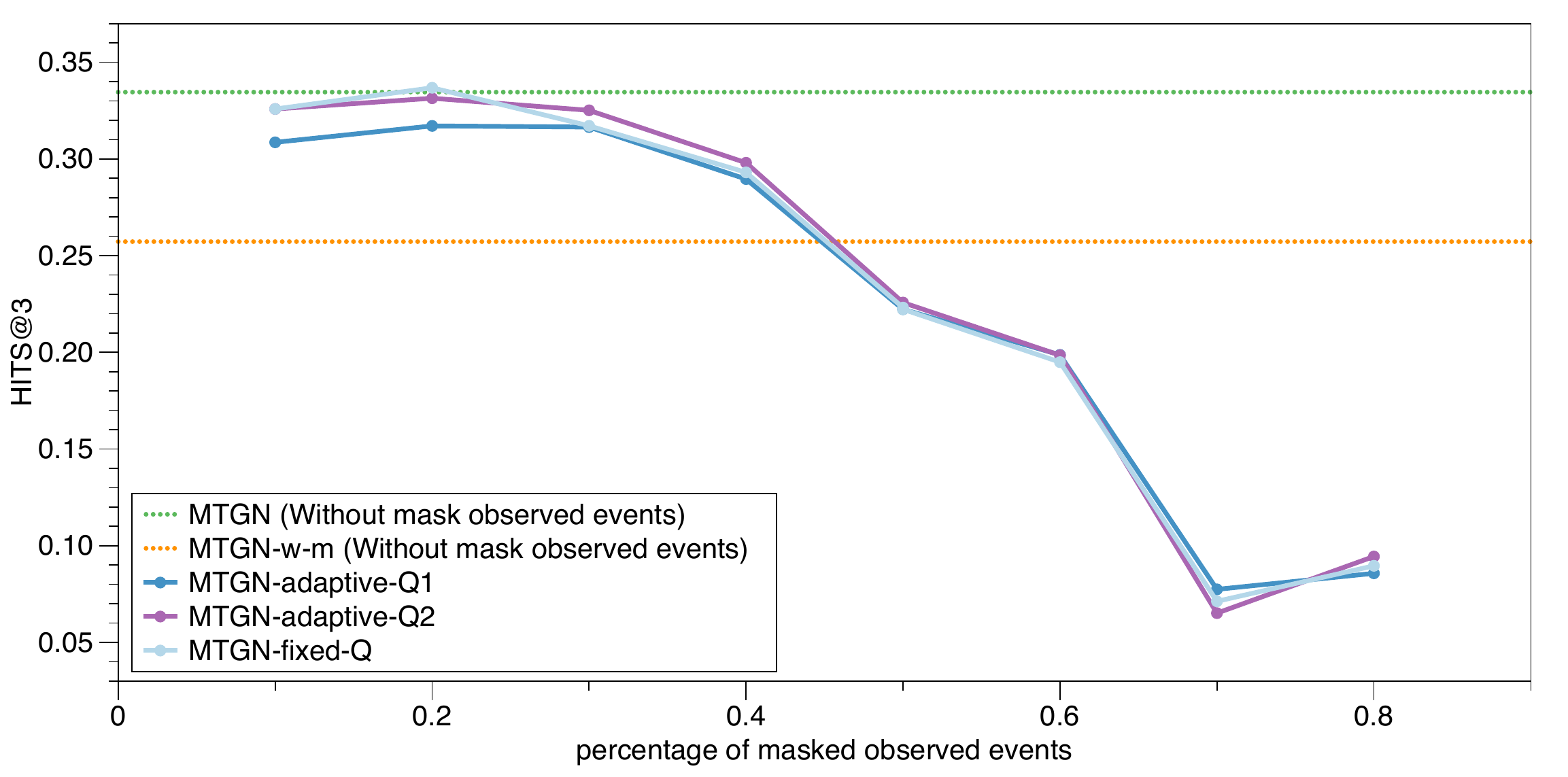}
	\includegraphics[width=0.49\linewidth]{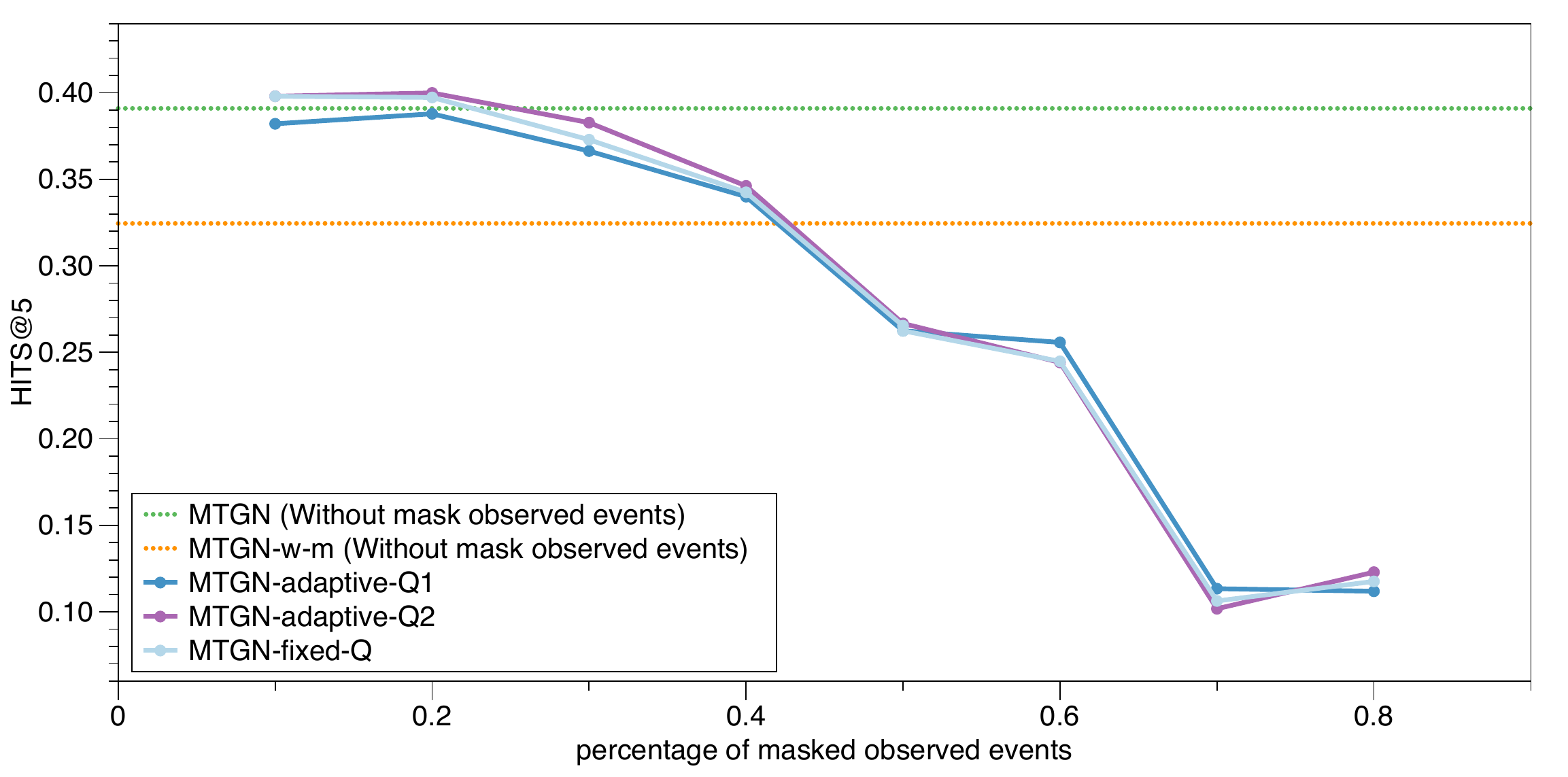}
	\includegraphics[width=0.49\linewidth]{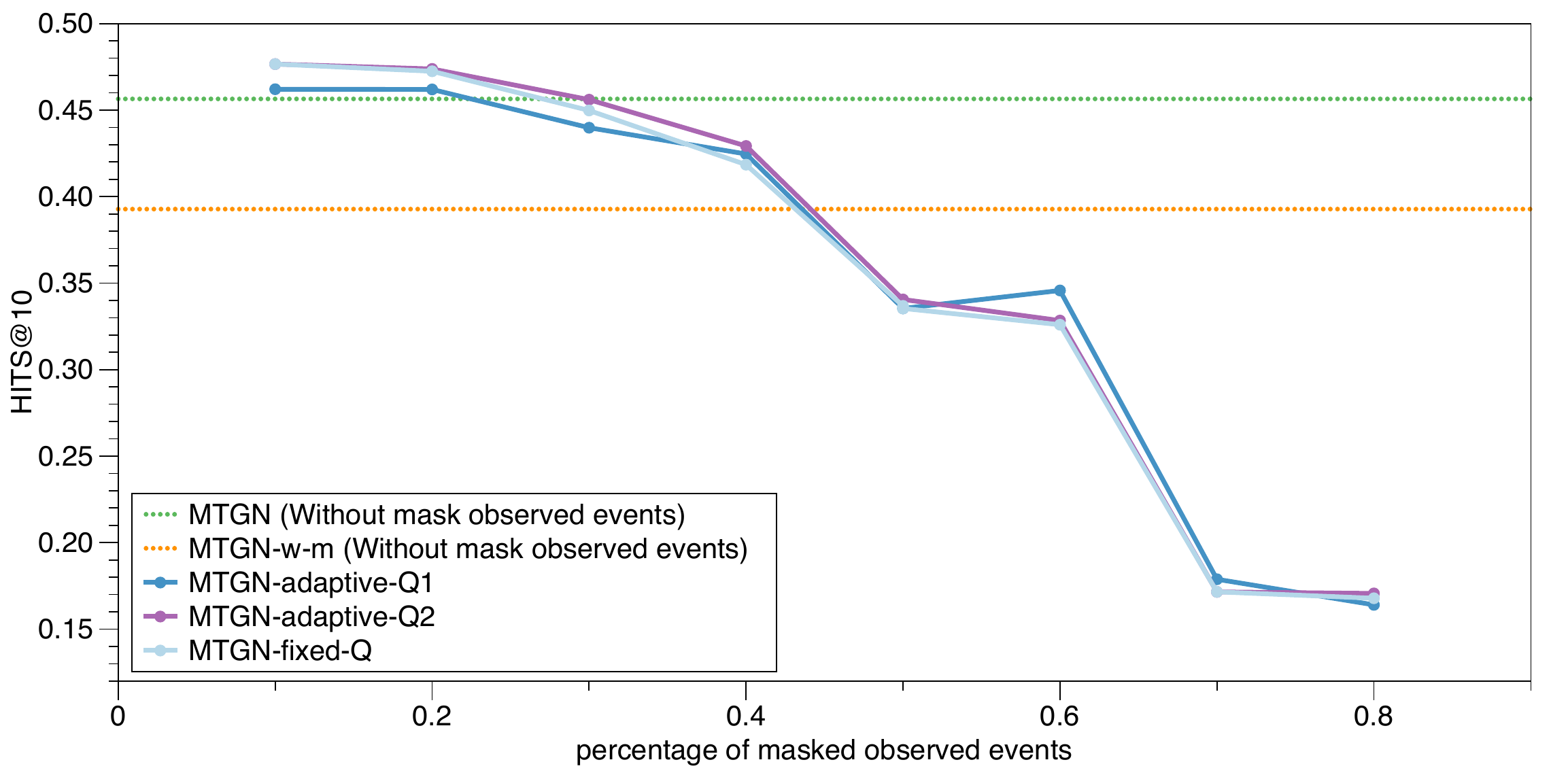}
	\includegraphics[width=0.49\linewidth]{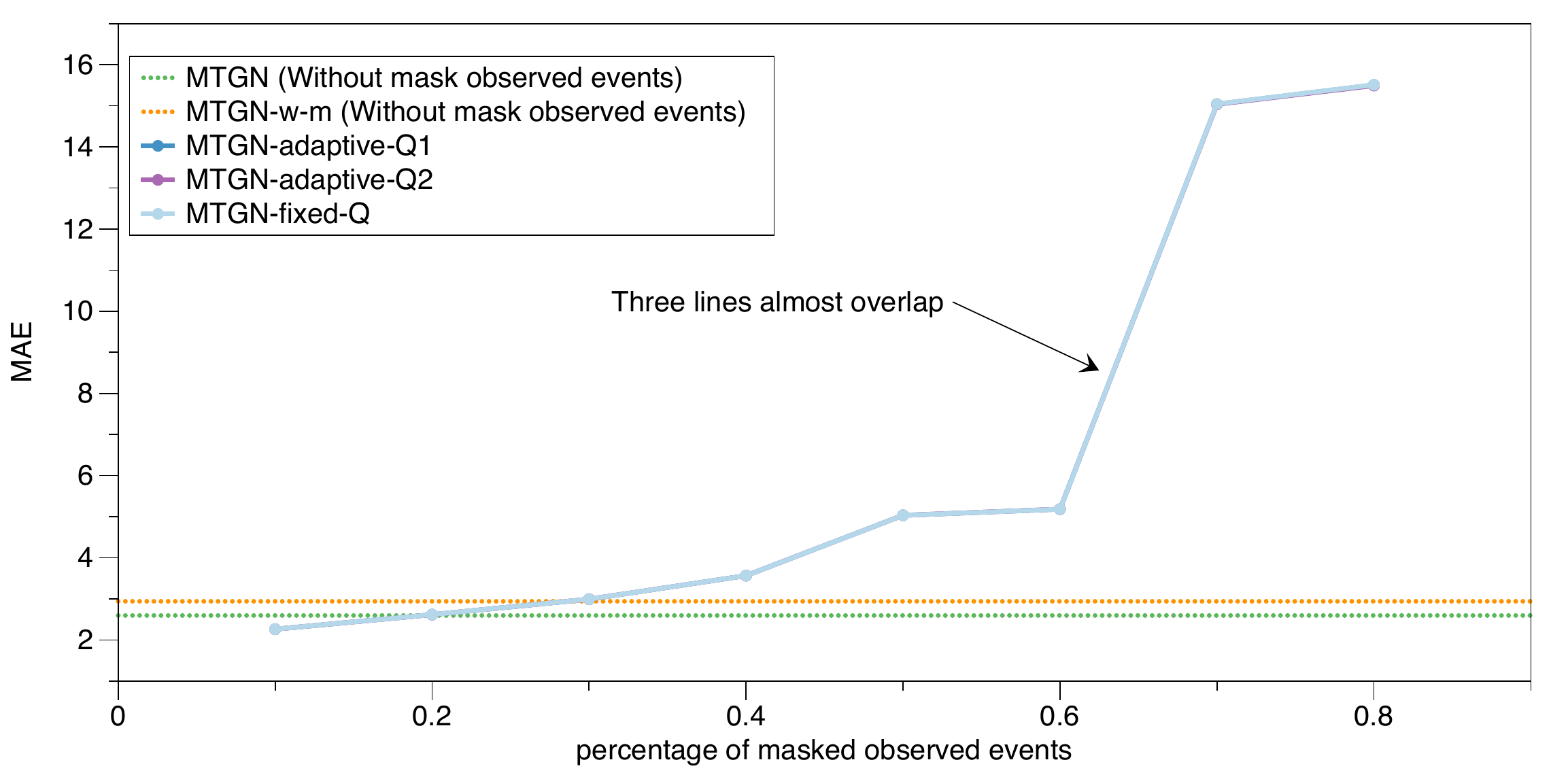}
	\caption{HITS@\{3,5,10\} and MAE score with respect to the the percentage of masked observed events and different missing events ratio $Q$ setting strategies on UCI dataset.}\label{fig:miss_r}
\end{figure}
Fig.~\mbox{\ref{fig:miss_r}} shows the experiment results, where the two dot lines denote the performance of MTGN (green) and MTGN-w-m (orange) on the whole observed events, respectively. From the results we can see that the performance of the model on both the link prediction and event time prediction tasks tends to decrease as the percentage of masked observed events increases. This is logical since we have less labeled/observed data. However, by modeling the missing events, we can greatly improve the robustness of the model. For example, the performance of the linked prediction remains stable when the observed event masking ratio $z \le 0.3$, and can even get better results than the original MTGN by adjusting $Q$ (this can be interpreted as some missing events contain more information than the observed events are sampled out). By comparing MTGN-adaptive-Q1 and MTGN-wo-m, we can illustrate that the improvement in model performance does not come from more event data in the training process due to the additional missing events obtained by sampling during the training process. The performance of the model degrades dramatically if the number of observed events is too small (e.g., the percentage of masked observed events exceeds 50\%). This is a validity threat for MTGN, i.e., MTGN may fail when the number of observed events is too small. As we claimed in the ablation study, modeling missing events will not affect event time prediction task too much.

\end{document}